\documentclass{article}
\usepackage[margin=2.8cm]{geometry}
\usepackage[]{inputenc} %

\usepackage{graphicx}
\usepackage{picinpar}
\usepackage{subcaption}
\usepackage{xcolor}
\usepackage{colortbl}

\usepackage{microtype} %

\usepackage{booktabs}
\usepackage{multirow}
\usepackage{soul}
\usepackage{enumerate}
\usepackage{color}
\usepackage{hyperref}
\usepackage{amsmath}
\usepackage{amssymb}
\usepackage{longtable}

\usepackage{algorithm}
\usepackage{algpseudocode}
\usepackage{kpfonts}

\usepackage[sort]{natbib}
\setcitestyle{round, elide, numbers}

\usepackage[titletoc]{appendix}
\usepackage{cleveref}
\usepackage{twemojis}

\usepackage{enumitem}
\usepackage{listings}
\usepackage{authblk}

\usepackage{fancyhdr}
\fancypagestyle{firstpage}{\fancyhead[L]{Preprint}\fancyhead[R]{}}

\newcommand{\ch}[1]{\multicolumn{1}{c}{\textbf{#1}}}

\newcommand{\furl}[1]{\footnote{\url{http://#1}}}

\title{EMMA-500: Enhancing Massively Multilingual Adaptation of Large Language Models}

\author[$^1$]{Shaoxiong Ji\thanks{Corresponding author}}
\author[$^1$]{Zihao Li}
\author[$^1$]{Jaakko Paavola}
\author[$^{2,4}$]{Peiqin Lin}
\author[$^3$]{Pinzhen Chen}
\author[$^3$]{\authorcr Dayyán O'Brien}
\author[$^1$]{Hengyu Luo}
\author[$^{2,4}$]{Hinrich Schütze}
\author[$^1$]{Jörg Tiedemann}
\author[$^3$]{Barry Haddow}

\affil[ ]{$^1$University of Helsinki \quad $^2$University of Munich}
\affil[ ]{$^3$University of Edinburgh \quad $^4$Munich Center for Machine Learning}

\affil[ ]{\textit {\{shaoxiong.ji;~zihao.li;~jaakko.paavola;~hengyu.luo;~jorg.tiedemann\}@helsinki.fi \authorcr~linpq@cis.lmu.de;~\{pinzhen.chen;~dobrien4;~bhaddow\}@ed.ac.uk}}

\date{}

\begin{document}
\maketitle

\thispagestyle{firstpage}

\begin{abstract}
\noindent In this work, we introduce \textbf{EMMA-500}, a large-scale multilingual language model continue-trained on texts across 546 languages designed for enhanced multilingual performance, with a focus on improving language coverage for low-resource languages. 
To facilitate continual pre-training, we compile \textbf{the MaLA corpus}, a comprehensive multilingual dataset and enrich it with curated datasets across diverse domains. 
Leveraging this corpus, we conduct extensive continual pre-training of the Llama 2 7B model, resulting in EMMA-500, which demonstrates robust performance across a wide collection of benchmarks, including a comprehensive set of multilingual tasks. 
Our results highlight the effectiveness of continual pre-training in expanding large language models' language capacity, particularly for underrepresented languages, demonstrating significant gains in cross-lingual transfer, task generalization, and language adaptability. We release the MaLA corpus, EMMA-500 model weights, scripts, and model generations.

\vspace{2ex}
\noindent \twemoji{robot} Model: \href{https://huggingface.co/collections/MaLA-LM/emma-500-66eaa9acf1f512c8915b7166}{huggingface.co/collections/MaLA-LM $\rightarrow$~EMMA-500}\\
\noindent \twemoji{floppy disk} Data: \href{https://huggingface.co/collections/MaLA-LM/mala-corpus-66e05127641a51de34d39529}{huggingface.co/collections/MaLA-LM $\rightarrow$~MaLA corpus}\\
\noindent \twemoji{writing hand} PolyWrite: \href{https://huggingface.co/datasets/MaLA-LM/PolyWrite}{huggingface.co/datasets/MaLA-LM/PolyWrite}\\
\noindent \twemoji{bar chart} Evaluation: \href{https://github.com/MaLA-LM/emma-500}{github.com/MaLA-LM/emma-500}
\end{abstract}

\section{Introduction}
\label{sec:introduction}

Multilingual large language models (LLMs) are designed to process and generate text in multiple languages. 
These models have evolved rapidly over the past decade, fueled by advances in deep learning, e.g., Transformer networks~\citep{vaswani2017attention}, pre-training techniques, and the availability of large-scale multilingual corpora such as mC4~\citep{raffel2020exploring_t5} and ROOTS~\citep{laurenccon2022bigscience}. 
The development of models like BERT~\citep{devlin2019bert}, GPT, and T5~\citep{raffel2020exploring_t5} opened the door for multilingual counterparts such as mBERT, XLM-R~\citep{conneau2020unsupervised}, mGPT~\citep{shliazhko2022mgpt}, and mT5~\citep{xue2021mt5}. These models were trained on massive multilingual corpora, allowing text in dozens of languages to be processed with the same set of model weights. 
By increasing the scale of both the data and the model, multilingual language models have demonstrated impressive performance on a wide array of tasks, such as text classification, machine translation, and question answering across different languages.
A key advantage of multilingual models is their ability to leverage cross-lingual transfer, where knowledge learned from high-resource languages (like English or Chinese) can be applied to other languages. However, despite the remarkable progress in multilingual models, many low-resource languages, which have limited available data, remain underserved.
While massive and extensive corpora are readily available for high-resource languages like English, French, and Spanish, languages such as Xhosa and Inuktitut often have only scarce or fragmented data.
This disparity between languages tends to make training datasets imbalanced, and multilingual models trained on such imbalanced datasets tend to prioritize high-resource languages, leaving low-resource languages underrepresented.

Recent studies adopt continual pre-training to enhance the language coverage of large language models on low-resource languages. 
For example, Glot500~\citep{imanigooghari2023glot500} and MaLA-500~\citep{lin2024mala} use continual pre-training and vocabulary extension using XLM-R and LLaMA, respectively, on the Glot500-c corpus covering 534 languages.
xLLMs-100~\citep{lai-etal-2024-llms} proceeds to multilingual instruction fine-tuning to improve the multilingual performance of LLaMA and BLOOM models on 100 languages, and Aya model \citep{ustun2024aya} applies continual training to the mT5 model \citep{xue2021mt5} using their constructed instruction dataset.  
LLaMAX~\citep{lu2024llamax} pushes the envelope by focusing on translation tasks via continual pre-training of LLaMA in over 100 languages. 

\begin{figure*}[ht]
    \centering
    \includegraphics[width=\linewidth]{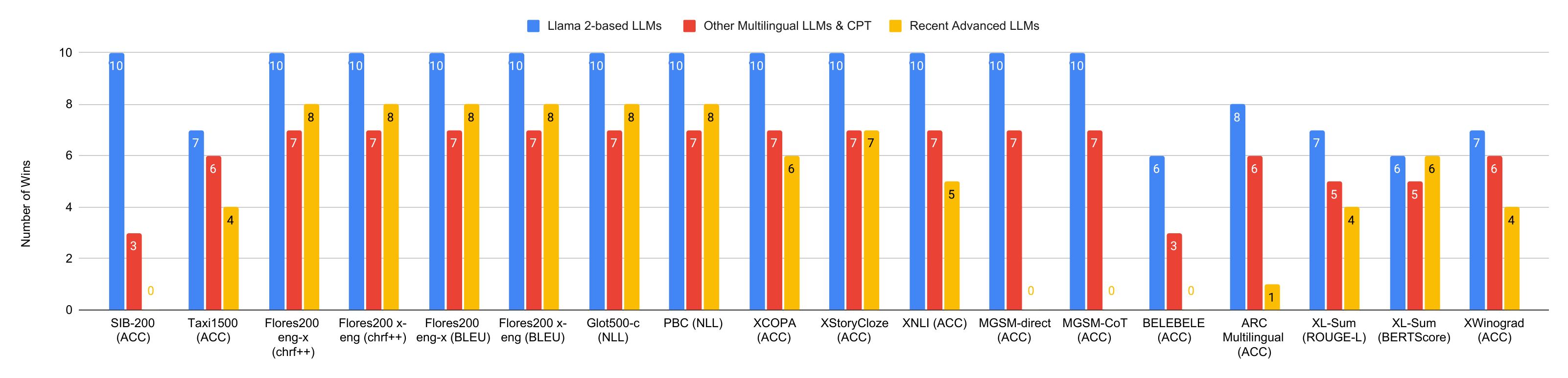}
    \caption{The number of wins, i.e., the number of times EMMA-500 achieves the best or superior performance compared to other models in the same category across various evaluation tasks and benchmarks.  We compare our EMMA-500 Llama 2 7B model to decoder-only LLMs of similar parameter size, including (i) 10 Llama 2-based LLMs, (ii) 7 multilingual LLMs and CPT models, and (iii) 8 recent advanced LLMs (see \Cref{sec:baselines}) on tasks and benchmarks in \Cref{tab:downstream_tasks}. If EMMA-500 scores higher than all compared models on a specific benchmark, it is considered a winning case for that particular evaluation. Our EMMA-500 Llama 2 model outperforms most Llama 2-based, multilingual LLMs and CPT models. Remarkably, our model achieves the best performance on Flores200, Glot500-c, and PBC among all the compared baselines.}
    \label{fig:overall_results}
\end{figure*}

As the field of multilingual LLMs evolves, the role of data becomes increasingly critical in enhancing the performance of the models, particularly when it comes to low-resource languages. 
To address this need for more and better data, we extend existing work, such as MaLA-500, by expanding the corpus for continual pre-training, coupled with large-scale training methods. 
We emphasize the creation of a massively multilingual corpus that not only increases the quantity of data but also diversifies the types of texts (e.g., code, books, scientific papers, and instructions). 
This ensures better language adaptation and broader language coverage, thus improving the representation and performance of multilingual language models, especially for underrepresented languages, ultimately creating more inclusive and versatile language models that cater to a broader linguistic diversity.
Our contribution is summarized as follows:
\begin{itemize}[nolistsep,noitemsep]
    \item We compile a massively multilingual corpus named MaLA to facilitate continual training of large language models for enhanced language adaptation across a wide range of linguistic contexts.
    \item We extend the MaLA corpus by integrating multiple curated datasets, creating a comprehensive and diverse data mix specifically for continual pre-training.
    \item We perform continual pre-training with the Llama 2 7B model\footnote{Choosing Llama 2 allows us to compare our model with many other models derived from it using continual pre-training. We plan to continue training models based on Llama 3/3.1 in the future.} on the multilingual corpus with 546 languages, resulting in the new EMMA-500 model. This model is rigorously evaluated on a diverse set of tasks and benchmarks.
\end{itemize}

\paragraph{The MaLA Corpus}
Our multilingual corpus features the following characteristics: 

\begin{itemize}[nolistsep,noitemsep]
    \item It contains 939 languages, (546 of which are used for training our EMMA-500 model) and 74 billion (B) whitespace delimited tokens in total.
    \item It has more than 300 languages with over 1 million whitespace delimited tokens and 546 languages with over 100k tokens. 
    \item It comes with four publicly available versions: (1) a noisy version after only basic pre-processing like extraction and harmonization; (2) a cleaned version after data cleaning; (3) a deduplicated version after approximate and exact deduplication; (4) a version with train and valid splits.
    \item Our augmentation to the MaLA corpus includes different types of texts such as code, books, scientific papers and instruction data, leading to a data mix with 100B+ whitespace delimited tokens.
\end{itemize}

\paragraph{Evaluation Results}
 In a comparison with decoder-only LLMs (\Cref{sec:baselines}) including Llama 2-based continual pre-trained models, LLMs that are designed to be multilingual, and the latest advanced LLMs, our model achieves strong results (\Cref{fig:overall_results}): 
 \begin{itemize}
     \item Out of models with parameter sizes from 4.5B to 13B, our model with 7B parameters has the lowest negative log-likelihood according to an intrinsic evaluation. 
     \item Our model remarkably improves the performance of commonsense reasoning, and machine translation over Llama 2-based models and multilingual baselines, and outperforms the latest advanced models in many cases.
     \item Our model improves the performance of text classification and natural language inference, outperforming all Llama 2-based models and LLMs designed to be multilingual.
     \item While math and machine reading comprehension (MRC) tasks are challenging for the Llama 2 7B model and other multilingual LLMs, our model remarkably enhances the Llama 2 base model. Our model yields improved performance on MRC over the base model but still produces quasi-random results similar to other multilingual baselines.
     \item We demonstrate that massively multilingual continued pre-training does not necessarily lead to regressions in other areas, such as code generation, if the data mix is carefully curated. Our model surpasses the Llama 2 7B base model's code generation abilities.
 \end{itemize}

\paragraph{Main Achievements} The MaLA corpus marks a significant leap forward in multilingual LLMs, distinguished by its unprecedented scale, diversity, and strategic emphasis on low-resource languages. By integrating a rich spectrum of text types---including code, scientific literature, and instructional data---it establishes a comprehensive and robust foundation for multilingual pretraining. These advancements culminate in EMMA-500, a multilingual LLM that demonstrates enhanced performance across 9 tasks and 15 benchmarks. The scope and impact of this work make it a landmark contribution, particularly in expanding NLP capabilities for underrepresented languages.

\section{The MaLA Corpus}

The MaLA corpus---MaLA standing for \textbf{Ma}ssive \textbf{L}anguage \textbf{A}daptation---is a diverse and extensive compilation of text data encompassing 939 languages sourced from a wide array of datasets.
It is developed for continual training of multilingual large language models.
The source datasets the MaLA corpus is compiled from exhibit a wide range of variance in various aspects. Examples of such elements include the data quality, the nature of the text content, how the data sources were organized into directory and file tree structures (if distributed as files rather than through an API) and the naming conventions used therein, the data formats and structures in the files or in-memory objects containing the text data, and the logic by which multilingual texts were aligned.
This section introduces the efforts made in data extraction, harmonization, pre-processing, cleaning, and deduplication in order to build the corpus. 

\subsection{Data Integration}
\label{sec:preprocessing}

To develop the MaLA corpus for training our language model, we establish a processing workflow consisting of the following key steps: 
(1) loading and curating identified data sources, 
(2) extracting and harmonizing both textual and metadata from these diverse sources into a unified format---often with tailored filtering, and 
(3) performing deduplication and further filtering on the textual data. 
In step (1), source data are either organized and loaded into memory from a file tree structure or, if available, accessed through an API. In step (2), the output is designed to be \texttt{JSON} Lines (\texttt{JSONL}) files with extracted text data and other desired content. A \texttt{JSONL} file contains multiple JSON records for storing data, each separated by a newline character. 
We selectively process only data annotated as training or development (validation) data, deliberately excluding test data.

\subsubsection{Extraction and Harmonization}

As mentioned, there is significant variability in the data quality of the source datasets used for compiling the corpus. Many of the datasets exhibit data quality challenges that would have adverse effects on model training if left unaddressed. 
These dataset-specific challenges are typically addressed during the pre-processing stage.
For example, we identify one issue involving text records consisting solely of date and timestamp information in the dataset for Languages of Russia~\citep{languages_of_russia}, likely resulting from a web scraper failing to differentiate between these elements and actual text. 
We address this by implementing a logic in the pre-processing script to detect and exclude such records. 

Due to the extensive volume of data, exhaustive examination of every source for data quality issues is impractical. Instead, we address issues only as we encounter them in our data exploration and pre-processing pipeline development efforts. This approach likely leaves some data quality problems undetected, since we do not go through the data systematically.

Despite the need for customized handling of certain dataset-specific idiosyncrasies, the core logic and structure of the pre-processing workflow remain consistent across most datasets. 
We develop a standardized pre-processing script that can be adapted with minor adjustments to accommodate different datasets.

\subsubsection{Language Code Normalization}

An essential component of our pre-processing pipeline involved converting language codes to the ISO 639-3 standard. 
This is crucial for ensuring consistent language identification across the source datasets. 
We rely on the declared language of each dataset and normalize it to the ISO standard without performing additional language identification at this step. 
This approach helps maintain uniformity while streamlining the pre-processing workflow.

For monolingual data, we primarily use the PyPI library \texttt{iso639-lang}\footnote{\url{https://pypi.org/project/iso639-lang/}} or \texttt{langcodes}\footnote{\url{https://pypi.org/project/langcodes}}. 
While converting language codes to ISO 639-3, we encounter several challenges. 
One issue is that some languages in ISO 639-3 are divided into multiple subvarieties, but our source data does not specify which subvariety is present. Our solution is to retain the original language code from the dataset, even when it does not conform to the ISO 639-3 standard. Another issue arises when certain languages are merged into other languages in the ISO 639-3 standard. In these cases, we update the language code to reflect the merged language.

Additionally, some language names or codes in the source data---referred to as ``original language names'' or ``original language codes''---are not recognized by the conversion libraries. In some cases, the reason behind this is that the original code in fact represents language families or groups of dialects (e.g., the ISO 639-2 codes ``ber'' for Berber languages and ``bih'' for Bihari languages), rather than specific languages. If so, we then retain the original codes, despite their non-compliance with ISO 639-3.
In other cases, the original language names are spelled differently from the standard recognized by the libraries. To address this mismatch, we implement a logic to detect and correct misspelled language names during pre-processing.
All these ``corner cases'' require careful attention in the pre-processing stage to ensure correct language code identification.

\subsubsection{Writing System Recognition}

In addition to normalizing language codes, we also identify the script or writing system used in the text data. We use the \texttt{GlotScript} library~\citep{kargaran2024glotscript} to recognize writing systems accordant to the ISO 15924 standard.
The process begins by sampling 100 random lines from each dataset (or the full dataset if it contains fewer than 100 lines). If \texttt{GlotScript} fails to identify a script from this sample, we attempt identification using just the first line of the sample. If this still does not yield a result, we set the script as ``None''. It is worth noting that we choose not to classify a dataset into multiple scripts, even when code-mixing (i.e., the use of multiple scripts) is present.

If script identification is unsuccessful after the initial steps, we assume the script matched a previously detected one for that language. In cases where no previous script information exists, we refer to a mapping of languages and their default scripts provided by the Glot500 corpora collection. Through this multi-step process, we are able to determine the script for every dataset without exceptions.

During script identification, we encounter several challenges. One issue is determining an appropriate length for the text chunk used for script recognition. A chunk that is too short could lead to incorrect identification if the text contains quotes or foreign language fragments using a different writing system. Conversely, using a chunk that is too long could result in excessive resource usage, slowing down processing or even causing memory exhaustion. 
Another consideration is whether to assume that a single file or dataset might contain multiple scripts. Such an assumption would require identifying the script at a more granular level, such as paragraph by paragraph or even sentence by sentence. Alternatively, we could assume that each file or dataset contained only one ``main'' script. This assumption would allow us to identify the script from a representative sample of the text for the whole file or dataset. We adopt the latter approach, recognizing a single dominant script for each dataset.
The output of this process is a label in the format language\_Script, e.g., eng\_Latn, where ``Language'' represents the ISO 639-3 language code and ``Script'' represents the ISO 15924 script code.

\subsection{Data Cleaning}
\label{sec:data_cleaning}

Most source data has already undergone data cleaning to different extents. Nonetheless, different cleaning processes have been adopted. 
We continue to clean the data to ensure consistency and accuracy for monolingual and bilingual texts.

Following the pipeline used by BigScience's pipeline for ROOTS corpus \citep{laurenccon2022bigscience}, we further adopt some necessary data cleaning to filter out text samples that might have undesirable quality. 
We first perform document modification for monolingual texts. The first step is whitespace standardization: all types of whitespace in a document are converted into a single, consistent space character. We split documents by newline characters, tabs, and spaces, strip words, and reconstruct the documents to remove very long words. However, these two steps do not apply to languages without whitespace word delimiters like Chinese, Japanese, Korean, Thai, Lao, Burmese, etc. We also remove words containing certain patterns, e.g., ``http'' and ``.com'', which are likely to be links and page source code. We then perform document filtering, including word count filtering, character repetition filtering, word repetition filtering, special characters filtering, stop words filtering, and flag words filtering. 
We re-identify the languages that are supported by the pre-trained \texttt{fastText}-based language identification model~\citep{joulin2016bag,joulin2016fasttext}. For other languages, we assume the language identification of the original data source and language code conversion are reliable.

\subsection{Data Deduplication}
\label{sec:data_dedup}

As we collect data from different sources, we deduplicate the data to remove the overlap between different sources. 
 \paragraph{MinHash Deduplication} 
For each language's dataset in each writing system, we start by using the MinHashLSH algorithm~\citep{rajaraman2011mining} to filter out similar documents. 
It is a near-deduplication technique that builds on MinHash~\citep{broder1997resemblance}, utilizing multiple hash functions for n-grams and the Jaccard similarity, and incorporates Locality-Sensitive Hashing to enhance efficiency. 
We use the implementation by \texttt{text-dedup} repository~\citep{chenghao_mou_2023_8364980}, applying 5-grams and a similarity threshold of 0.7 to identify similar documents based on the Jaccard index.

 \paragraph{Exact Deduplication}
We further deploy exact deduplication using the \texttt{text-dedup} repository again with precise matching. 
It takes each document through a hash function, i.e., MD5~\citep{rivest1992md5} in our choice, and the hash values of all documents are compared to identify duplicates.

\subsection{Key Statistics}

This section presents the final MaLA corpus obtained after data sourcing, pre-processing, cleaning, and deduplication. 
\Cref{tab:stats} first shows some basic data statistics and compares them with other multilingual corpora for pre-training language models or language adaptation.
Additional statistics are presented in \Cref{sec:additiona_stats} in the appendix.
The token counts are based on white-space delimitation, though it might not be accurate for languages like Chinese, Japanese and Korean since the entire clause is counted as one token. 
Glot500-c~\citep{imanigooghari2023glot500} has 534 languages in total, in which 454 languages are directly distributed on Huggingface~\footnote{\url{https://huggingface.co/datasets/cis-lmu/Glot500}}. 
We also omit high-resource languages in the other three datasets, i.e., MADLAD \citep{kudugunta2024madlad}, CulturaX \citep{nguyen2023culturax}, and CC100 \citep{cc100_2}, as our main focus is continual pre-training for language adaptation.
The final MaLA corpus consists of 939 languages, 546 of which have more than 100k tokens and are used for training our EMMA-500 model. 
Counting languages with more than 1 million tokens, the MaLA corpus and Glot500-c have more than 300, while MADLAD and CulturaX have 200 and 100 respectively. 
Compared with Glot500-c, the MaLA corpus contains documents with significantly higher sequence lengths, with an average token count of 90 versus 19. 
This higher sequence length is advantageous for continually training LLMs because it provides more context within each training example, allowing the model to better capture long-range dependencies and patterns in the data. 
As a result, MaLA is more effective for language adaptation.

\begin{table}[t]
\scriptsize
\centering
\caption{Data statistics of the MaLA corpus and comparison to other multilingual corpora. The number of documents and tokens is in millions.}
\label{tab:stats}
\setlength{\tabcolsep}{3pt}
\begin{tabular}{lrrrrr}
\toprule
\ch{Dataset}   & \ch{N Lang} & \ch{N Lang Counted} & \ch{N Docs} & \ch{N Tokens}  & \ch{Avg Tokens/Doc} \\
\midrule
Glot500-c & 534         & 454                & 1,815  & 35,449    & 19.53       \\
MADLAD    & 419         & 414                & 1,043  & 645,111   & 618.51      \\
CulturaX  & 167         & 161                & 2,141  & 1,029,810 & 480.99      \\
CC100     & 116         & 101                & 2,557  & 52,201    & 20.41       \\
\hline
MaLA      & 939         & 546                & 824    & 74,255    & 90.12      \\
\bottomrule
\end{tabular}
\end{table}

\section{Data Mixing and Model Training}
\label{sec:mixing}

Incorporating a diverse data mix---spanning various languages, domains, document lengths and styles---is crucial for continual training of large language models to enhance their versatility, generalization ability, and robustness across a wide range of tasks and domains.
We augment the MaLA corpus with diverse data to mitigate issues such as over-fitting to specific styles or topics or underperforming on tasks outside the training distribution.

\subsection{Data Mixing}

\paragraph{Curated Data}
We enhance the corpus with high-quality curated data, specifically high-resource languages in the monolingual part. 
We use texts from scientific papers as these provide a structured, information-dense corpus that can improve the model's ability to handle technical language and domain-specific content. 
They are (1) CSL~\citep{li2022csl}, a large-scale Chinese Scientific Literature dataset, that contains titles, abstracts, keywords and academic fields of 396,209 papers; (2) pes2o~\citep{peS2o}, a collection of full-text open-access academic papers derived from the Semantic Scholar Open Research Corpus (S2ORC)~\citep{lo-wang-2020-s2orc}.
We further add free e-books from the Gutenberg project\footnote{\url{https://www.gutenberg.org/}} compiled by \citet{project_gutenberg_HF_manu}. 
These texts enhance the range of literary styles and narrative forms, thus enhancing the model's versatility.
Adding high-resource languages into the pre-training corpora also mitigates the forgetting in model training.

\paragraph{Instruction Data}
We further augmented the training corpus by incorporating instruction-based datasets, inspired by \citet{li2023colossal,taylor2022galactica,nakamura2024aurora}.
We mix two instruction data into our training corpus.
They are: (1) xp3x (Crosslingual Public Pool of Prompts eXtended)\footnote{\url{https://huggingface.co/datasets/CohereForAI/xP3x}}, a multitask instruction collection in 277 languages~\citep{muennighoff2022crosslingual}; (2) the Aya collection~\footnote{\url{https://huggingface.co/datasets/CohereForAI/aya_collection_language_split}} that contains both human-curated and machine translated instructions in 101 languages~\citep{singh-etal-2024-aya}.
For both instruction datasets, we use their training set.

\paragraph{Code}

We additionally enrich the training corpus by sourcing code data from The Stack~\citep{kocetkov2023stack}. This is done following existing work that demonstrates the value of code data in improving the reasoning ability of language models~\citep{zhang2024unveiling, ma2024stage} while also mitigating any catastrophic forgetting of the base model's programming knowledge.

We subsample The Stack at an effective rate of 15.2\%, prioritizing high-quality source files and data science code~\footnote{\url{https://huggingface.co/datasets/AlgorithmicResearchGroup/arxiv_research_code}}. We retain the 32 most important non-data programming languages by prevalence while also adding in all \texttt{llvm} code following prior work detailing its importance in multi-lingual code generation~\citep{szafraniec2022translation, paul2024ircoder}. We also source from data-heavy formats but follow precedent~\citep{lozhkov2024starcoder} and subsample them more aggressively. For a more detailed read on filtering heuristics, we direct the reader to Appendix~\ref{appendix:code_sourcing}.

\begin{table}[t]
\centering
\scriptsize
\caption{Data mix for continual training. Code and reasoning-related data are counted by Llama 2 tokenizer and others are counted as whitespace delimited; `inst' stands for instruction and `mono' stands for monolingual texts.}
\label{tab:data_mix}
\setlength{\tabcolsep}{1.5pt}
\begin{tabular}{lrrrr}
\toprule
\ch{Data}                      & \ch{Original Counts}       & \ch{Sample Rate} & \ch{Final Counts}        & \ch{Percentage} \\
\midrule
inst high                 & 42,121,055,562  & 0.1         & 4,212,105,556  & 3.08\%      \\
inst medium-high+         & 6,486,592,274   & 0.2         & 1,297,318,455  & 0.95\%      \\
inst medium-high          & 30,651,187,534  & 0.5         & 15,325,593,767 & 11.21\%     \\
inst medium               & 1,444,764,863   & 1.0           & 1,444,764,863  & 1.06\%      \\
inst medium-low           & 47,691,495      & 5.0           & 238,457,475    & 0.17\%      \\
inst low                  & 3,064,796       & 20.0          & 61,295,920     & 0.04\%      \\
inst code/reasoning       & 612,208,775     & 1.0         & 612,208,775    & 0.45\%      \\
code			          & 221,003,976,266 & 0.1         & 20,786,882,764 & 15.20\%     \\
curated (EN pes2o)          & 56,297,354,921  & 0.2         & 11,241,574,489 & 8.22\%      \\
curated (ZH CSL \& wiki)                  & 61,787,372      & 1.0         & 61,787,372     & 0.05\%      \\
curated (Gutenberg)				  & 5,173,357,710   & 1.0         & 5,173,357,710  & 3.78\%      \\
mono high EN                   & 3,002,029,817   & 0.1         & 300,202,982    & 0.22\%      \\
mono high                 & 40,411,201,964  & 0.5         & 20,205,600,982 & 14.78\%     \\
mono medium-high          & 27,515,227,962  & 1.0         & 27,515,227,962 & 20.12\%     \\
mono medium               & 2,747,484,380   & 5.0         & 13,737,421,900 & 10.05\%     \\
mono medium-low           & 481,935,633     & 20.0        & 9,638,712,660  & 7.05\%      \\
mono low                  & 97,535,696      & 50.0        & 4,876,784,800  & 3.57\%     \\
\bottomrule
\end{tabular}
\end{table}

\begin{table*}[t]
\scriptsize
\centering
\caption{Evaluation statistics.
    Sample/Lang: average number of test samples per language; N Lang: number of languages covered; NLL: negative log-likelihood; ACC: accuracy.}
\label{tab:downstream_tasks}
\setlength{\tabcolsep}{1pt}
\begin{tabular}{llcrrc}
    \toprule
    \ch{Tasks}  & \ch{Dataset} & \ch{Metric} & \ch{Samples/Lang} & \ch{N Lang} & \ch{Domain}\\
    \midrule
    \multirow{2}{*}{Intrinsic Evaluation (\Cref{sec:intrinsic_evaluation})} & Glot500-c test \citep{imanigooghari2023glot500} & NLL & 1000 & 534 & Misc \\
     & PBC \citep{DBLP:conf/lrec/MayerC14} & NLL & 500 & 370 & Bible \\
    \midrule
     \multirow{2}{*}{Text Classification (\Cref{sec:text_classification})}& SIB200 \citep{sib-200} & ACC & 204 & 205 & Misc \\
     &Taxi1500 \citep{ma2023taxi1500} & ACC & 111 & 1507 & Bible \\
    \midrule 
    \multirow{3}{*}{Commonsense Reasoning (\Cref{sec:commonsense})}& XCOPA~\citep{ponti-etal-2020-xcopa} & ACC & 600  & 11 & Misc \\
    & XStoryCloze \citep{lin2022few} & ACC & 1870 & 11 & Misc \\
    & XWinograd~\citep{tikhonov-ryabinin-2021-heads} & ACC & 741.5 & 6 & Misc \\
     \midrule
    \multirow{1}{*}{Natural Language Inference (\Cref{sec:NLI})}& XNLI \citep{conneau2018xnli} & ACC & 2490 & 15 & Misc \\
     \midrule
    \multirow{1}{*}{Machine Translation (\Cref{sec:machine_translation})}& FLORES-200 \citep{costa2022no} & BLEU, chrF++ & 1012 & 204 & Misc \\
     \midrule
    \multirow{1}{*}{Summarization (\Cref{sec:summarization})}& XL-Sum \citep{hasan-etal-2021-xl} & ROUGE-L, BERTScore & 2537 & 44 & News \\
     \midrule
    \multirow{2}{*}{Math (\Cref{sec:math})}& MGSM direct \citep{shi2022language} & ACC & 250 & 10 & Misc \\
	& MGSM CoT \citep{shi2022language} & ACC & 250 & 10 & Misc \\
    \midrule
    \multirow{2}{*}{Machine Comprehension (\Cref{sec:MRC})}& BELEBELE \citep{bandarkar2023belebele} & ACC & 900 & 122 & Misc \\
      & ARC multilingual \citep{lai2023okapi} & ACC & 1170 & 31 & Misc \\
     \midrule
    \multirow{1}{*}{Code Generation (\Cref{sec:code_generation})}& Multipl-E~\citep{Cassano2022MultiPLEAS} & Pass@$k$ & 164 & 7 & Misc \\
    \bottomrule
    \end{tabular}
\end{table*}

\vspace{-5pt}
\paragraph{Data Mix}
Our final data mix for continual training is listed in \Cref{tab:data_mix}. 
The resource categorization refers to \Cref{sec:supported_languages} in the appendix and \texttt{inst medium-high+} is a separate category with languages with more than 500 million but less than 1B tokens. 
For monolingual text, we also have a separate category for English. 
In continual learning, where new data is introduced to an existing model, there is a risk of ``catastrophic forgetting'', where the model loses knowledge from earlier training stages. 
Although our work's primary focus is in a low-resource regime, we enhance the training corpus with a wide range of data types, including books and scientific papers in high-resource languages, code, and instruction data in our data mix.
We downsample texts in high-resource languages and upsample text in low-resource languages using different sample rates according to how resourceful the language is.  
We make our data mix diverse and balanced towards different resource groups of languages in order to retain the prior knowledge of the model while learning new information, especially in medium- and low-resource languages, thus maintaining high performance across both previously seen and new languages. 
The final data mix has around 136B tokens.

\subsection{Model Training}
We employ continual training using the causal language modelling objective for the decoder-only Llama model and exposing the pre-trained model to new data to develop our EMMA-500 model.
We adopt efficient training strategies combining optimization, memory management, precision handling, and distributed training techniques. 
Our EMMA-500 model is trained on the Leonardo supercomputer\footnote{\url{https://leonardo-supercomputer.cineca.eu}}, occupying 256 Nvidia A100 GPUs, using the GPT-NeoX framework~\citep{gpt-neox-library}.
During training, we set a global batch size of 4096 and worked with sequences of 4096 tokens. The training process ran for 12,000 steps, resulting in a total of 200 billion Llama 2 tokens.
We use the Adam optimizer \citep{kingma2015adam} with a learning rate of 0.0001, betas of [0.9, 0.95], and an epsilon of 1e-8.
We use a cosine learning rate scheduler with a warm-up of 500 iterations.
To reduce memory consumption, activation checkpointing is employed.
Precision is managed through mixed-precision techniques, using bfloat16 for computational efficiency and maintaining FP32 for gradient accumulation.

\section{Evaluation}
\label{sec:evaluation}

\subsection{Tasks, Benchmarks, and Baselines}
\label{sec:tasks_benchmarks_baselines}

 \paragraph{Tasks and Benchmarks}
We conduct a comprehensive evaluation to validate the usability of our processed data and data mixing for massively multilingual language adaptation.
We perform the intrinsic evaluation of the models' performance on next-word prediction and evaluate the model's performance on downstream tasks.
Table~\ref{tab:downstream_tasks} lists the datasets we used as downstream evaluation datasets in this work.
For math tasks, we perform direct prompting and Chain-of-Thoughts prompting on the MGSM benchmark and evaluate the performance using both exact and flexible matches. 
For code generation tasks, we perform test-case-based execution-tested evaluations using the \texttt{pass@k} metric \citep{chen2021codex}. We benchmark for \texttt{k} values of 1,10, and 25 using a generation pool of 50 samples per problem.
For massively multilingual benchmarks with more than 100 languages, we categorize languages into five groups---high, medium-high, medium, medium-low, and low---based on their token counts in the MaLA corpus as listed in \Cref{tab:languages} in \Cref{sec:supported_languages}.

\paragraph{Evaluation Software}
We use the Language Model Evaluation Harness (\texttt{lm-evaluation-harness}) framework~\citep{eval-harness} for benchmarking test sets that are already ingested in the framework. For other benchmarks, we use in-house developed evaluation scripts and other open-source implementations. 
In text classification tasks such as SIB-200 and Taxi-1500, the evaluation protocol involves calculating the probability of the next output token for each candidate category. These probabilities are then sorted in descending order, with the category having the highest probability being selected as the model's prediction. This process is implemented using the Transformers~\citep{wolf2019huggingface} library.
In tasks like Machine Translation and Open-Ended Generation, the vLLM library~\citep{kwon2023efficient} is used to accelerate inference, providing significant speed improvements.
Similarly, for code generation tasks, we use a VLLM-enabled evaluation harness package (\texttt{vllm-code-harness})\footnote{\url{https://github.com/iNeil77/vllm-code-harness}} for the execution-tested evaluations.

\paragraph{Baselines}
\label{sec:baselines}

We compare our model with three groups of decoder-only models. 
They are (1) Llama 2 models~\citep{touvron2023llama2} and continual pre-trained models based on Llama 2, such as CodeLlama~\citep{roziere2023code}, MaLA-500~\citep{lin2024mala}, LLaMAX~\citep{lu2024llamax}, Tower~\citep{alves2024tower}, and YaYi\footnote{\url{https://huggingface.co/wenge-research/yayi-7b-llama2}}; (2) other LLMs and continual pre-trained LLMs designed to be massively multilingual, including BLOOM~\citep{scao2022bloom}, mGPT~\citep{shliazhko2022mgpt}, and Occiglot\footnote{\url{https://huggingface.co/occiglot/occiglot-7b-eu5}}; 
and (3) recent LLMs with superior English capabilities like Llama 3~\citep{dubey2024llama3}, Llama 3.1\footnote{\url{https://llama.meta.com/docs/model-cards-and-prompt-formats/llama3_1}}, Qwen 2~\citep{yang2024qwen2technicalreport}, Aya 23 \citep{aryabumi2024aya23} and Gemma 2~\citep{gemmateam2024gemma2improvingopen}. 
There are also some other LLMs such as OpenAI's API models~\footnote{\url{https://platform.openai.com/docs/models}} and xLLMs-100~\citep{lai-etal-2024-llms}. However, they do not release the model weights or they limit access to them through commercial API, so we did not include them. 
The MADLAD model \citep{kudugunta2024madlad} that uses the decoder-only T5 architecture is not supported by inference engines such as the HuggingFace transformers \citep{wolf2019huggingface}. We do not compare them in this work.

\subsection{Intrinsic Evaluation}
\label{sec:intrinsic_evaluation}

For intrinsic evaluation, we compute the negative log-likelihood (NLL) of the test text given by the tested LLMs instead of using length-normalized perplexity due to different text tokenization schemes across models.
We concatenate the test set as a single input and run a sliding-window approach.\footnote{\url{https://huggingface.co/docs/transformers/en/perplexity}} 
To make a comparison across models, we use the Glot500-c test set, which covers 534 evaluated languages. 
We also test on the Parallel Bible Corpus (PBC) which could yield NLL more comparable across languages. 
\Cref{tab:ppl} and \Cref{tab:ppl_para} show the intrinsic evaluation results on Glot500-c test and PBC, respectively. As shown, our model attains lower NLL compared to all other models on both test sets and languages with different resource availability.
This suggests that the test sets are more similar to the underlying training data of our model and it can be interpreted that EMMA-500 has learned to compress massive multilingual text more efficiently.

\begin{table*}
\caption{NLL on Glot500-c test. EMMA-500 Llama 2 7B has better average performance than all baselines.}
\label{tab:ppl}
\centering
\scriptsize
\begin{tabular}{lrrrrrr}
\toprule
\ch{Model} & \ch{Avg} & \ch{High} & \ch{Medium-High} & \ch{Medium} & \ch{Medium-Low} & \ch{Low} \\
\midrule
Llama 2 7B & 190.58 & 146.43 & 176.30 & 205.86 & 210.41 & 196.45 \\
Llama 2 7B Chat & 218.87 & 173.67 & 204.78 & 239.18 & 240.97 & 223.86 \\
CodeLlama 2 7B & 193.96 & 146.43 & 180.20 & 210.09 & 212.70 & 200.60 \\
LLaMAX Llama 2 7B & 187.37 & 108.58 & 142.84 & 197.74 & 212.22 & 203.83 \\
LLaMAX Llama 2 7B Alpaca & 169.12 & 94.84 & 123.90 & 173.54 & 193.83 & 187.34 \\
MaLA-500 Llama 2 10B v1 & 155.62 & 127.51 & 153.93 & 173.02 & 166.01 & 158.12 \\
MaLA-500 Llama 2 10B v2 & 151.25 & 123.82 & 147.69 & 167.46 & 161.20 & 155.13 \\
YaYi Llama 2 7B &192.98 &149.32 &179.10 &208.96 &212.80 &198.77 \\
TowerBase Llama 2 7B & 192.98 & 150.41 & 180.19 & 209.12 & 212.18 & 198.89 \\
TowerInstruct Llama 2 7B & 199.44 & 157.33 & 186.42 & 216.92 & 218.82 & 204.93 \\
\midrule
Occiglot Mistral 7B v0.1 & 191.11 & 159.48 & 185.53 & 209.26 & 207.67 & 194.07 \\
Occiglot Mistral 7B v0.1 Instruct & 193.83 & 162.31 & 188.20 & 212.41 & 210.81 & 196.58 \\
BLOOM 7B & 202.95 & 160.33 & 195.01 & 216.15 & 220.46 & 206.89 \\
BLOOMZ 7B & 217.32 & 178.12 & 210.99 & 235.53 & 235.60 & 220.65 \\
mGPT & 340.37 & 311.14 & 337.29 & 388.97 & 367.45 & 343.14 \\
mGPT 13B &282.46 &254.78 &276.03 &327.15 &317.14 &277.13 \\
Yayi 7B & 226.67 & 181.48 & 217.34 & 243.42 & 246.58 & 231.65 \\
\midrule
Aya 23 8B &207.85 &175.18 &201.33 &227.29 &225.01 &210.90 \\
Aya expanse 8B &206.75 &162.19 &191.52 &229.32 &225.36 &211.71 \\
Llama 3 8B & 156.36 & 102.78 & 129.36 & 153.11 & 167.26 & 173.56 \\
Llama 3.1 8B & 154.59 & 101.23 & 127.57 & 150.87 & 164.81 & 172.05 \\
Gemma 7B & 692.25 & 583.39 & 721.77 & 817.40 & 729.61 & 689.60 \\
Gemma 2 9B & 320.81 & 348.26 & 351.08 & 380.49 & 338.00 & 303.62 \\
Qwen 2 7B & 188.55 & 132.47 & 171.50 & 200.26 & 210.21 & 196.78 \\
Qwen 1.5 7B & 195.52 & 141.37 & 181.41 & 212.10 & 217.16 & 202.46 \\
\midrule
EMMA-500 Llama 2 7B & 112.20 & 81.78 & 100.89 & 122.53 & 99.28 & 109.25 \\
\bottomrule
\end{tabular}
\end{table*}

\begin{table*}
\caption{NLL on PBC. EMMA-500 Llama 2 7B has better average performance than all baselines.}
\label{tab:ppl_para}
\centering
\scriptsize
\begin{tabular}{lrrrrrr}
\toprule
\ch{Model} & \ch{Avg} & \ch{High} & \ch{Medium-High} & \ch{Medium} & \ch{Medium-Low} & \ch{Low} \\
\midrule
Llama 2 7B & 122.10 & 91.30 & 99.41 & 112.31 & 133.08 & 135.34 \\
Llama 2 7B Chat & 139.14 & 108.40 & 115.78 & 129.79 & 149.54 & 152.82 \\
CodeLlama 2 7B & 123.98 & 93.27 & 101.83 & 113.47 & 134.65 & 137.52 \\
LLaMAX Llama 2 7B & 117.39 & 69.41 & 79.06 & 103.74 & 131.90 & 138.03 \\
LLaMAX Llama 2 7B Alpaca & 107.81 & 60.05 & 69.36 & 93.39 & 122.44 & 128.77 \\
MaLA-500 Llama 2 10B v1 & 103.20 & 94.04 & 98.60 & 100.53 & 105.04 & 107.65 \\
MaLA-500 Llama 2 10B v2 & 101.67 & 92.42 & 96.30 & 98.88 & 103.34 & 106.62 \\
YaYi Llama 2 7B &123.57 &93.22 &100.95 &114.01 &134.44 &136.76 \\
TowerBase Llama 2 7B & 123.70 & 93.64 & 101.44 & 114.72 & 134.41 & 136.77 \\
TowerInstruct Llama 2 7B & 127.30 & 98.21 & 105.17 & 118.65 & 137.53 & 140.28 \\
\midrule
Occiglot Mistral 7B v0.1 & 121.64 & 95.15 & 101.86 & 114.37 & 131.49 & 132.93 \\
Occiglot Mistral 7B v0.1 Instruct & 123.18 & 96.86 & 103.41 & 115.88 & 132.98 & 134.48 \\
BLOOM 7B & 129.55 & 96.62 & 111.22 & 115.33 & 138.19 & 143.03 \\
BLOOMZ 7B & 137.72 & 107.03 & 119.89 & 125.85 & 145.27 & 150.95 \\
mGPT & 225.14 & 211.35 & 203.84 & 219.75 & 229.91 & 239.82 \\
mGPT 13B &180.54 &164.98 &160.25 &179.81 &186.99 &191.31 \\
YaYi 7B & 143.80 & 108.79 & 123.20 & 130.60 & 152.37 & 158.73 \\
\midrule
Aya 23 8B &131.60 &102.16 &109.05 &121.83 &141.64 &145.46 \\
Aya expanse 8B &130.80 &96.67 &104.73 &122.42 &142.47 &145.57 \\
Llama 3 8B & 102.55 & 64.20 & 71.82 & 85.22 & 114.79 & 121.29 \\
Llama 3.1 8B & 101.43 & 62.98 & 70.68 & 83.83 & 113.36 & 120.33 \\
Gemma 7B & 460.86 & 399.22 & 427.14 & 468.66 & 463.11 & 483.29 \\
Gemma 2 9B & 197.41 & 200.76 & 192.07 & 197.88 & 196.56 & 202.69 \\
Qwen 2 7B & 120.44 & 83.34 & 94.87 & 107.84 & 133.38 & 136.52 \\
Qwen 1.5 7B & 124.02 & 89.36 & 100.55 & 113.08 & 135.54 & 139.13 \\
\midrule
EMMA-500 Llama 2 7B & 68.11 & 50.12 & 55.62 & 64.78 & 60.53 & 65.68 \\
\bottomrule
\end{tabular}
\end{table*}

\subsection{Machine Translation}
\label{sec:machine_translation}

FLORES-200 is an evaluation benchmark for translation tasks with 204 language pairs involving English and thus 408 translation directions, with a particular focus on low-resource languages. 
We assess all language models by adopting a 3-shot evaluation approach with the prompt in \Cref{sec:prompt_mt}.
The performance is measured by BLEU~\citep{papineni-etal-2002-bleu} and chrF++~\citep{popovic-2015-chrf} implemented in \texttt{sacrebleu}~\citep{post-2018-call}. The BLEU score is calculated with the \texttt{flores200} tokenizer applied to the texts and chrF++ uses word order 2. The choice of \texttt{flores200} tokenization ensures that languages that do not have a whitespace delimiter can be evaluated at the (sub-)word level. For reproducibility, we attach the BLEU and chrF++ signatures.\footnote{The BLEU signature is nrefs:1|case:mixed|eff:no|tok:flores200|smooth:exp|version:2.4.2; the chrF++ signature is nrefs:1|case:mixed|eff:yes|nc:6|nw:2|space:no|version:2.4.2}

\Cref{tab:flores200-x-eng} presents the average X-to-English (X-Eng) translation results.\footnote{We mark BLOOMZ with a ${\dag}$ because it has used FLORES in its instruction tuning data; we mark MaLA-500 with a ${\ddag}$ because it has used FLORES in its training data but with source and target sides split. Besides, as a remark, Tower, LLaMAX, and our EMMA-500 have intentionally used parallel data (not FLORES) in the training stage.} 
Our EMMA-500 model outperforms all other models on average. We achieve the best performance across all language settings, except for high-resource languages where our model slightly lags behind Llama 3/3.1, Gemma 7B, and LlaMAX 7B Alpaca. 
In the English-to-X (Eng-X) translation direction, as shown in \Cref{tab:flores200-eng-x}, the advantage of EMMA-500 is even more pronounced. 
We outperform all other models even in high-resource languages, and the advantage becomes more significant in lower-resource languages. 
Overall, we note that our model outperforms Tower models which are explicitly adjusted to perform translation tasks in high-resource languages. Further, the much larger margin between EMMA-500 and other models in Eng-X compared with X-Eng indicates that our EMMA-500 model is particularly good at generating non-English texts.

\begin{table*}[ht!]
\caption{3-shot results on FLORES-200 (X-Eng, BLEU/chrF++). EMMA-500 Llama 2 7B has better average performance than all baselines.}
\label{tab:flores200-x-eng}
\centering
\scriptsize
\begin{tabular}{lrrrrrr}
\toprule
\ch{Model} & \ch{Avg} & \ch{High} & \ch{Medium-High} & \ch{Medium} & \ch{Medium-Low} & \ch{Low} \\
\midrule
Llama 2 7B & 12.93/ 30.32 & 19.91/ 39.04 & 17.56/ 35.84 & 12.49/ 29.81 & 8.27/ 24.35 & 6.96/ 23.36 \\
Llama 2 7B Chat & 12.28/ 31.72 & 18.98/ 39.65 & 17.06/ 37.03 & 11.74/ 31.1 & 7.79/ 26.34 & 6.18/ 25.03 \\
CodeLlama 2 7B & 10.82/ 28.57 & 17.39/ 37.43 & 15.27/ 33.94 & 10.39/ 28.05 & 6.45/ 22.85 & 5.04/ 21.48 \\
LLaMAX Llama 2 7B & 1.99/ 13.66 & 3.68/ 22.18 & 2.95/ 18.15 & 1.83/ 12.84 & 0.67/ 7.2 & 1.01/ 9.04 \\
LLaMAX Llama 2 7B Alpaca & 22.29/ 42.27 & 32.83/ 54.56 & 30.04/ 51.25 & 21.7/ 41.94 & 13.06/ 31.32 & 14.24/ 32.88 \\
MaLA-500 Llama 2 10B v1$^{\ddag}$ & 2.29/ 13.6 & 4.64/ 15.95 & 3.18/ 14.64 & 2.68/ 14.23 & 1.24/ 12.58 & 0.33/ 11.18 \\
MaLA-500 Llama 2 10B v2$^{\ddag}$ & 2.87/ 15.44 & 5.58/ 18.65 & 3.81/ 16.33 & 3.55/ 16.29 & 1.63/ 14.2 & 0.55/ 12.76 \\
Yayi Llama 2 7B & 12.98/ 31.38 & 19.48/ 39.58 & 17.55/ 36.71 & 12.47/ 30.79 & 8.54/ 25.63 & 7.22/ 24.84 \\
TowerBase Llama 2 7B & 13.74/ 31.47 & 21.76/ 40.96 & 18.92/ 37.27 & 13.15/ 30.9 & 8.3/ 25.05 & 7.21/ 24.1 \\
TowerInstruct Llama 2 7B & 4.81/ 25.43 & 9.18/ 34.4 & 6.66/ 30.01 & 4.62/ 25.22 & 2.64/ 20.24 & 1.8/ 18.69 \\
\midrule
Occiglot Mistral 7B v0.1 & 13.12/ 31.13 & 19.53/ 38.93 & 17.57/ 36.27 & 13.07/ 31.2 & 9.03/ 26.15 & 6.86/ 23.83 \\
Occiglot Mistral 7B v0.1 Instruct & 11.61/ 31.65 & 16.72/ 39.28 & 15.06/ 36.48 & 11.7/ 31.73 & 8.48/ 26.88 & 6.54/ 24.7 \\
BLOOM 7B & 9.57/ 27.84 & 15.75/ 36.65 & 9.65/ 28.19 & 9.42/ 27.81 & 6.81/ 23.95 & 8.61/ 25.89 \\
BLOOMZ 7B$^{\dag}$ & 20.22/ 34.74 & 32.23/ 47.03 & 19.2/ 34.08 & 20.09/ 34.49 & 16.25/ 30.58 & 18.54/ 32.63 \\
mGPT & 5.29/ 20.69 & 9.37/ 26.64 & 8.28/ 25.29 & 3.41/ 17.87 & 2.43/ 16.07 & 2.84/ 17.28 \\
mGPT-13B & 7.42/ 24.58 & 12.61/ 31.95 & 11.11/ 30.16 & 5.72/ 22.49 & 3.57/ 18.16 & 4.11/ 20.04 \\
Yayi 7B & 4.82/ 21.36 & 5.69/ 25.18 & 4.53/ 19.97 & 4.41/ 21.52 & 3.71/ 19.18 & 6.13/ 23.12 \\
\midrule
Aya 23 8B & 13.87/ 32.36 & 19.87/ 40.18 & 18.5/ 37.91 & 12.99/ 31.99 & 8.19/ 25.6 & 9.84/ 26.56 \\
Aya expanse 8B & 13.12/ 36.86 & 17.83/ 44.92 & 15.44/ 41.09 & 13.14/ 37.16 & 9.26/ 30.39 & 10.72/ 31.97 \\
Llama 3 8B & 23.78/ 43.72 & 33.71/ 55.36 & 30.31/ 51.3 & 24.75/ 44.91 & 15.18/ 33.65 & 16.01/ 34.65 \\
Llama 3.1 8B & 24.19/ 44.1 & 34.15/ 55.7 & 30.79/ 51.7 & 24.98/ 45.26 & 15.89/ 34.24 & 16.13/ 34.85 \\
Gemma 2 9B & 23.15/ 38.87 & 33.11/ 51.36 & 30.81/ 48.53 & 25.58/ 41.23 & 15.37/ 30.03 & 11.73/ 24.15 \\
Gemma 7B & 23.79/ 43.68 & 34.23/ 55.77 & 29.87/ 50.95 & 24.0/ 44.25 & 16.16/ 34.36 & 16.03/ 34.58 \\
Qwen 1.5 7B & 15.58/ 35.87 & 24.07/ 46.29 & 19.92/ 40.74 & 15.76/ 36.27 & 9.74/ 28.81 & 9.77/ 29.13 \\
Qwen 2 7B & 17.39/ 37.61 & 27.63/ 50.06 & 22.48/ 43.28 & 18.13/ 38.63 & 9.89/ 28.54 & 10.64/ 29.99 \\
\midrule
EMMA-500 Llama 2 7B & 25.37/ 45.78 & 32.24/ 53.74 & 31.39/ 52.85 & 25.72/ 46.16 & 20.32/ 39.96 & 17.18/ 36.15 \\
\bottomrule
\end{tabular}
\end{table*}

\begin{table}[ht!]
\caption{3-shot results on FLORES-200 (Eng-X, BLEU/chrF++). EMMA-500 Llama 2 7B has better average performance than all baselines.}
\label{tab:flores200-eng-x}
\centering
\scriptsize
\begin{tabular}{lrrrrrr}
\toprule
\ch{Model} & \ch{Avg} & \ch{High} & \ch{Medium-High} & \ch{Medium} & \ch{Medium-Low} & \ch{Low} \\
\midrule
Llama 2 7B & 4.62/ 15.13 & 10.77/ 24.38 & 8.56/ 21.4 & 2.55/ 13.72 & 0.74/ 8.72 & 0.7/ 7.92 \\
Llama 2 7B Chat & 4.95/ 16.95 & 10.87/ 24.51 & 8.54/ 22.69 & 3.25/ 15.5 & 1.52/ 12.08 & 0.94/ 10.03 \\
CodeLlama 2 7B & 4.27/ 14.94 & 10.04/ 23.48 & 7.79/ 20.79 & 2.57/ 14.2 & 0.71/ 9.27 & 0.58/ 7.49 \\
LLaMAX Llama 2 7B & 0.8/ 7.42 & 1.85/ 12.06 & 1.2/ 9.74 & 0.54/ 6.55 & 0.22/ 4.52 & 0.38/ 4.81 \\
LLaMAX Llama 2 7B Alpaca & 12.51/ 28.35 & 24.8/ 41.76 & 18.69/ 38.42 & 10.1/ 27.27 & 3.79/ 16.53 & 6.68/ 18.15 \\
MaLA-500 Llama 2 10B v1$^{\ddag}$ & 0.6/ 6.08 & 1.51/ 9.0 & 1.13/ 8.19 & 0.35/ 5.99 & 0.07/ 4.5 & 0.02/ 2.9 \\
MaLA-500 Llama 2 10B v2$^{\ddag}$ & 0.54/ 6.38 & 1.4/ 9.19 & 1.02/ 8.42 & 0.24/ 5.99 & 0.07/ 5.14 & 0.02/ 3.27 \\
Yayi Llama 2 7B & 4.41/ 14.87 & 10.49/ 24.0 & 8.21/ 21.27 & 2.52/ 13.57 & 0.6/ 8.49 & 0.53/ 7.42 \\
TowerBase Llama 2 7B & 4.83/ 16.03 & 11.89/ 24.15 & 8.33/ 21.46 & 2.57/ 14.49 & 1.38/ 11.6 & 0.74/ 8.9 \\
TowerInstruct Llama 2 7B & 3.23/ 15.64 & 7.22/ 22.65 & 4.99/ 20.0 & 2.2/ 14.9 & 1.62/ 12.31 & 0.73/ 8.97 \\
\midrule
Occiglot Mistral 7B v0.1 & 4.32/ 16.1 & 10.5/ 23.74 & 6.95/ 20.91 & 2.87/ 15.44 & 1.47/ 12.0 & 0.79/ 9.09 \\
Occiglot Mistral 7B v0.1 Instruct & 3.99/ 15.8 & 9.46/ 23.17 & 6.46/ 20.73 & 2.68/ 15.29 & 1.31/ 11.31 & 0.84/ 9.04 \\
BLOOM 7B & 2.81/ 11.8 & 7.53/ 19.0 & 3.12/ 13.36 & 2.05/ 11.48 & 0.85/ 8.0 & 2.09/ 9.22 \\
BLOOMZ 7B$^{\dag}$ & 7.44/ 16.1 & 23.64/ 32.22 & 7.46/ 16.62 & 6.98/ 16.05 & 1.28/ 9.99 & 4.17/ 11.77 \\
mGPT & 2.59/ 12.56 & 5.24/ 17.04 & 4.75/ 16.92 & 1.14/ 9.75 & 0.78/ 9.24 & 0.84/ 9.08 \\
mGPT-13B & 3.88/ 14.57 & 8.32/ 21.7 & 6.84/ 20.55 & 2.06/ 12.23 & 0.9/ 8.4 & 1.33/ 9.58 \\
YaYi 7B & 4.37/ 13.5 & 13.72/ 26.28 & 4.68/ 14.31 & 3.35/ 12.89 & 0.91/ 8.51 & 2.55/ 10.08 \\
\midrule
Aya 23 8B & 6.46/ 16.15 & 11.99/ 22.87 & 9.64/ 21.38 & 3.57/ 13.86 & 2.09/ 10.15 & 5.17/ 12.25 \\
Aya expanse 8B & 6.88/ 23.89 & 12.41/ 31.11 & 8.92/ 28.5 & 5.09/ 22.45 & 3.55/ 18.29 & 5.32/ 19.53 \\
Llama 3 8B & 9.93/ 24.08 & 20.38/ 36.87 & 14.95/ 32.05 & 8.89/ 24.28 & 2.83/ 14.26 & 4.2/ 14.29 \\
Llama 3.1 8B & 10.11/ 24.69 & 20.82/ 37.39 & 15.3/ 32.82 & 8.85/ 24.85 & 2.9/ 14.83 & 4.23/ 14.81 \\
Gemma 2 9B & 12.09/ 26.48 & 24.62/ 40.69 & 17.82/ 35.51 & 10.68/ 26.58 & 3.38/ 15.02 & 5.94/ 15.98 \\
Gemma 7B & 9.05/ 23.05 & 17.58/ 34.5 & 13.62/ 30.16 & 7.96/ 22.85 & 2.64/ 14.11 & 4.47/ 14.82 \\
Qwen 1.5 7B & 5.87/ 17.77 & 14.05/ 28.6 & 8.88/ 23.57 & 3.85/ 17.07 & 1.7/ 10.85 & 2.35/ 10.21 \\
Qwen 2 7B & 5.56/ 17.17 & 13.22/ 27.65 & 8.21/ 22.36 & 4.15/ 16.93 & 1.56/ 10.47 & 2.19/ 10.11 \\
\midrule
EMMA-500 Llama 2 7B & 15.58/ 33.25 & 26.37/ 42.4 & 21.96/ 41.98 & 13.4/ 32.06 & 9.15/ 27.99 & 7.92/ 21.25 \\
\bottomrule
\end{tabular}
\end{table}

\subsection{Text Classification}
\label{sec:text_classification}

SIB-200~\citep{sib-200} and Taxi1500~\citep{ma2023taxi1500} are two prominent topic classification datasets. 
SIB-200 encompasses seven categories: science/technology, travel, politics, sports, health, entertainment, and geography. 
Taxi1500 spans 1507 languages, involving six classes: Recommendation, Faith, Description, Sin, Grace, and Violence.

We use 3-shot prompting with prompts in \Cref{sec:prompt_cls}, drawing demonstrations from the development set and testing models on the test split. 
The outcomes are tabulated in \Cref{tab:sib200_taxi1500} for SIB-200 and Taxi1500. 
For SIB-200, our EMMA-500 model outperforms all Llama2-based models, with particularly notable gains in languages with medium or fewer resources---seeing an average improvement of 47.5\%. 
Taxi-1500 could be a more challenging task since it is in the religious domain, but our model still surpasses all Llama2-based models except for MalA-500. 
However, despite these improvements in both classification tasks, our models lag behind the latest models such as Llama3 and 3.1, especially in high-resource languages. 

\begin{table*}[ht!]
\caption{3-shot results on SIB-200 and Taxi-1500 (ACC \%). EMMA-500 Llama 2 7B has better average performance than Llama 2 models and comparable performance with multilingual LLMs, and has comparable performance with the compared LLMs.}
\label{tab:sib200_taxi1500}
\setlength{\tabcolsep}{1pt}
\scriptsize
\centering
\begin{tabular}{lrrrrrrrrrrrr}
\toprule
\multicolumn{1}{c}{\multirow{2}{*}{\textbf{Model}}} & \multicolumn{6}{c}{\textbf{SIB-200}}                                                                          & \multicolumn{6}{c}{\textbf{Taxi-1500}}                                                                         \\ 
\cmidrule(lr){2-7}\cmidrule(lr){8-13}
                                                   & \ch{Avg} & \ch{High} & \ch{Med-High} & \ch{Medium} & \ch{Med-Low} & \ch{Low} 
                                                   & \ch{Avg} & \ch{High} & \ch{Med-High} & \ch{Medium} & \ch{Med-Low} & \ch{Low} \\
\midrule
Llama 2 7B                                        & 22.41   & 26.64   & 24.69   & 22.05   & 19.68   & 19.00   
                                                  & 17.54   & 19.50   & 19.49   & 18.47   & 17.46   & 17.37   \\
Llama 2 7B Chat                                   & 25.58   & 29.72   & 28.11   & 25.01   & 23.03   & 21.91   
                                                  & 15.44   & 18.73   & 17.66   & 16.61   & 15.59   & 15.22   \\
CodeLlama 2 7B                                    & 23.35   & 26.06   & 25.42   & 23.10   & 21.42   & 20.37   
                                                  & 17.03   & 17.45   & 17.41   & 17.20   & 17.05   & 17.00   \\
LLaMAX Llama 2 7B                                 & 10.61   & 12.42   & 11.60   & 10.01   & 9.45    & 9.54    
                                                  & 23.52   & 23.20   & 23.40   & 23.76   & 23.56   & 23.52   \\
LLaMAX Llama 2 7B Alpaca                          & 27.89   & 33.09   & 32.12   & 27.16   & 23.38   & 22.82   
                                                  & 15.09   & 18.70   & 16.88   & 15.99   & 15.00   & 14.91   \\
MaLA-500 Llama 2 10B v1                           & 23.25   & 23.30   & 23.64   & 22.88   & 22.76   & 23.58   
                                                  & 25.27   & 23.90   & 24.02   & 24.76   & 24.57   & 25.43   \\
MaLA-500 Llama 2 10B v2                           & 19.30   & 18.93   & 21.05   & 19.49   & 17.55   & 18.46   
                                                  & 23.39   & 21.36   & 22.30   & 21.32   & 21.72   & 23.67   \\
Yayi Llama 2 7B                                  & 24.57   & 29.04   & 27.17   & 24.42   & 21.44   & 20.69   
                                                  & 17.73   & 18.74   & 18.46   & 18.19   & 17.89   & 17.65   \\
TowerBase Llama 2 7B                             & 19.34   & 22.00   & 20.92   & 18.74   & 17.90   & 16.93   
                                                  & 17.73   & 18.49   & 18.81   & 18.67   & 18.10   & 17.61   \\
TowerInstruct Llama 2 7B                         & 20.53   & 23.21   & 21.96   & 20.26   & 19.15   & 18.04   
                                                  & 17.29   & 20.17   & 19.60   & 18.08   & 17.40   & 17.09   \\ 
\midrule
Occiglot Mistral 7B v0.1                         & 32.69   & 38.80   & 35.82   & 31.74   & 28.92   & 28.36   
                                                  & 22.26   & 24.64   & 22.91   & 22.99   & 22.33   & 22.15   \\
Occiglot Mistral 7B v0.1 Instruct                & 34.31   & 39.48   & 37.16   & 33.36   & 31.47   & 30.08   
                                                  & 18.76   & 24.30   & 20.90   & 19.41   & 19.18   & 18.48   \\
BLOOM 7B                                         & 17.81   & 23.13   & 18.05   & 17.17   & 15.76   & 17.02   
                                                  & 14.76   & 15.58   & 14.89   & 14.56   & 15.11   & 14.73   \\
BLOOMZ 7B                                        & 29.73   & 30.39   & 29.63   & 29.80   & 29.53   & 29.70   
                                                  & 16.96   & 16.93   & 16.98   & 16.96   & 16.99   & 16.95   \\
mGPT                                             & 27.11   & 28.58   & 27.99   & 26.73   & 25.89   & 26.48   
                                                  & 10.72   & 8.67    & 8.44    & 9.92    & 10.29   & 10.93   \\
mGPT-13B                                         & 33.20   & 36.69   & 34.27   & 34.48   & 29.39   & 32.26   
                                                  & 17.23   & 17.98   & 16.44   & 15.88   & 16.10   & 17.38   \\
Yayi 7B                                          & 35.76   & 40.57   & 36.20   & 35.63   & 34.72   & 33.18   
                                                  & 16.12   & 16.65   & 16.38   & 16.45   & 15.83   & 16.11   \\ 
\midrule
Aya 23 8B                                        & 41.50   & 46.78   & 45.27   & 42.43   & 35.27   & 37.87                                 
                                                  & 22.64   & 22.95   & 22.50   & 22.13 & 22.41 & 22.67   \\
Aya expanse 8B                                   & 57.01   & 65.05   & 61.51   & 56.84   & 48.59 & 53.80                                                  
                                                  & 18.73   & 20.02   & 18.60  & 18.93   & 19.39 & 18.66   \\
Llama 3 8B                                       & 63.69   & 73.45   & 70.25   & 64.62   & 53.16   & 56.96   
                                                  & 21.73   & 31.84   & 27.08   & 25.60   & 22.61   & 21.05   \\
Llama 3.1 8B                                     & 61.42   & 70.70   & 67.90   & 61.99   & 51.46   & 54.75   
                                                  & 20.20   & 27.51   & 25.21   & 24.43   & 20.97   & 19.59   \\
Gemma 7B                                         & 58.21   & 68.06   & 64.55   & 58.16   & 48.86   & 51.12   
                                                  & 18.05   & 28.68   & 28.55   & 24.99   & 20.28   & 16.89   \\
Gemma 2 9B                                       & 46.25   & 51.77   & 49.00   & 46.92   & 43.04   & 40.45   
                                                  & 13.83   & 24.29   & 24.13   & 18.74   & 14.97   & 12.83   \\
Qwen 1.5 7B                                      & 47.95   & 56.00   & 51.81   & 48.25   & 41.56   & 42.86   
                                                  & 7.29    & 12.65   & 11.45   & 8.78    & 7.30    & 6.92    \\
Qwen 2 7B                                        & 54.95   & 66.37   & 60.14   & 55.17   & 45.19   & 49.25   
                                                  & 21.87   & 27.37   & 25.57   & 24.01   & 22.33   & 21.47   \\ 
\midrule
EMMA-500 Llama 2 7B                              & 31.27   & 32.75   & 33.28   & 30.99   & 30.83   & 27.60   
                                                  & 19.82   & 23.66   & 23.33   & 22.77   & 22.00   & 19.30   \\ 
\bottomrule
\end{tabular}
\end{table*}

\subsection{Commonsense Reasoning}
\label{sec:commonsense}

We assess the models' commonsense reasoning ability using three multilingual benchmarks. XCOPA \citep{ponti-etal-2020-xcopa} is a dataset covering 11 languages that focuses on causal commonsense reasoning across multiple languages.
XStoryCloze \citep{lin2022few} is derived from the English StoryCloze dataset \citep{mostafazadeh2017lsdsem} and translated into 10 non-English languages, testing commonsense reasoning within a story. In this task, the system must choose the correct ending for a four-sentence narrative.
XWinograd~\citep{tikhonov-ryabinin-2021-heads} is a multilingual collection of Winograd Schemas~\citep{levesque2012winograd} available in six languages, aimed at evaluating cross-lingual commonsense reasoning.
We perform zero-shot evaluations. Accuracy is used as the evaluation metric.
We categorize languages into different resource groups based on language availability, accessibility, and possible corpus size. For XCOPA, we have three groups, i.e., high-resource (Italian, Turkish, Vietnamese, and Chinese), medium-resource (Swahili due to its regional influence, Estonian, Haitian Creole, Indonesian, Thai, and Tamil), and low-resource languages (Cusco-Collao Quechua).\footnote{Note that there is no perfect categorization for language resource groups.}
For XStoryCloze, the resource group is high-resource (Arabic, English, Spanish, Russian, and Chinese), medium (Hindi, Indonesian, Swahili, and Telugu), and low (Basque and Burmese). For XWinograd, there is only one group for high-resource languages. 
\Cref{tab:commonsense_agg} shows the evaluation results of zero-shot commonsense reasoning. 

Compared with Llama 2-based models on XCOPA, our model improves the average performance by a large margin---up to a 5\% increase when compared with the best-performing TowerInstruct based on Llama 2 7B.\footnote{The biggest improvements are on medium-resource languages. However, we note that the resource categorization is not perfect.}
Our model also outperforms all the multilingual LLMs. 
Recent LLMs such as Gemma and Llama 3 have stronger performance than Llama 2 models, and Gemma 2 9B performs the best. However, our model achieves better performance than Qwen, Llama 3, and 3.1. 
We gain similar results on XStoryCloze, outperforming all the models except Gemma 2 9B. 
As for XWinograd, a multilingual benchmark with high-resource only, our model achieves improved performance than Llama 2, despite not being as good as Tower models, which target high-resource languages. However, our model is comparable to other multilingual LLMs.
For low-resource languages, our model outperforms all the compared LLMs except LLaMAX Llama 2 7B Alpaca on XCOPA, where we achieve the same accuracy as it. 

\begin{table*}[ht!]
\caption{0-shot results (ACC \%) on commonsense reasoning: XCOPA, XStoryCloze, and XWinograd. EMMA-500 Llama 2 7B has better average performance than Llama 2 models and multilingual LLMs on XCOPA, XStoryCloze, and comparable performance on XWinograd.}
\label{tab:commonsense_agg}
\centering
\scriptsize
\begin{tabular}{lrrrrrrrrr}
\toprule
\multicolumn{1}{c}{\multirow[b]{2}{*}{\textbf{Model}}} & \multicolumn{4}{c}{\textbf{XCOPA}}         & \multicolumn{4}{c}{\textbf{XStoryCloze}}   & \multicolumn{1}{c}{\textbf{XWinograd}} \\
\cmidrule(lr){2-5}\cmidrule(lr){6-9}\cmidrule(lr){10-10}
 & \ch{Avg}   & \ch{High}   & \ch{Medium} & \ch{Low}    & \ch{Avg}   & \ch{High}   & \ch{Medium} & \ch{Low}    & \ch{Avg}                          \\
\midrule
                       Llama 2 7B & 56.67 & 62.10 & 53.90 & 51.60 & 57.55 & 63.38 & 54.45 & 49.21 & 72.47 \\
                  Llama 2 7B Chat & 55.85 & 61.25 & 53.13 & 50.60 & 58.41 & 64.80 & 54.77 & 49.74 & 69.45 \\
                   CodeLlama 2 7B & 54.69 & 58.70 & 52.53 & 51.60 & 55.68 & 60.68 & 52.33 & 49.90 & 70.92 \\
                LLaMAX Llama 2 7B & 54.38 & 55.50 & 54.13 & 51.40 & 60.36 & 64.34 & 58.82 & 53.47 & 67.49 \\
         LLaMAX Llama 2 7B Alpaca & 56.60 & 59.80 & 55.17 & 52.40 & 63.83 & 69.08 & 62.01 & 54.33 & 69.86 \\
          MaLA-500 Llama 2 10B v1 & 53.09 & 53.55 & 53.27 & 50.20 & 53.07 & 58.15 & 49.22 & 48.08 & 65.89 \\
          MaLA-500 Llama 2 10B v2 & 53.09 & 53.55 & 53.27 & 50.20 & 53.07 & 58.15 & 49.22 & 48.08 & 65.89 \\
                  YaYi Llama 2 7B & 56.71 & 62.10 & 54.13 & 50.60 & 58.42 & 64.98 & 54.91 & 49.04 & 74.50 \\
             TowerBase Llama 2 7B & 56.33 & 62.50 & 52.90 & 52.20 & 57.78 & 64.35 & 53.67 & 49.57 & 74.29 \\
         TowerInstruct Llama 2 7B & 57.05 & 62.90 & 54.00 & 52.00 & 59.24 & 66.83 & 54.53 & 49.70 & 74.00 \\
\hline
         Occiglot Mistral 7B v0.1 & 56.67 & 62.80 & 53.37 & 52.00 & 58.10 & 65.18 & 53.28 & 50.03 & 74.61 \\
Occiglot Mistral 7B v0.1 Instruct & 56.55 & 62.85 & 52.97 & 52.80 & 59.39 & 66.94 & 54.33 & 50.63 & 72.93 \\
                         BLOOM 7B & 56.89 & 59.95 & 55.87 & 50.80 & 59.30 & 61.99 & 59.05 & 53.08 & 70.13 \\
                        BLOOMZ 7B & 54.87 & 56.35 & 54.60 & 50.60 & 57.12 & 61.14 & 55.82 & 49.67 & 67.95 \\
                             mGPT & 55.04 & 57.10 & 54.40 & 50.60 & 54.43 & 54.96 & 54.53 & 52.91 & 59.69 \\
                         mGPT 13B & 56.18 & 59.75 & 55.13 & 48.20 & 56.44 & 57.76 & 56.35 & 53.31 & 63.59 \\
                          YaYi 7B & 56.64 & 59.55 & 55.50 & 51.80 & 60.67 & 64.90 & 59.40 & 52.65 & 69.79 \\
\hline
                        Aya 23 8B & 55.13 & 59.05 & 53.30 & 50.40 & 60.93 & 67.27 & 58.82 & 49.34 & 65.84 \\
                   Aya expanse 8B & 56.38 & 61.50 & 53.77 & 51.60 & 64.80 & 73.10 & 61.93 & 49.80 & 71.94 \\
                       Llama 3 8B & 61.71 & 68.35 & 59.03 & 51.20 & 63.41 & 68.50 & 62.03 & 53.44 & 76.84 \\
                     Llama 3.1 8B & 61.71 & 69.30 & 58.80 & 48.80 & 63.58 & 68.66 & 62.09 & 53.87 & 75.52 \\
                         Gemma 7B & 63.64 & 70.35 & 61.43 & 50.00 & 65.01 & 69.46 & 64.49 & 54.93 & 77.41 \\
                       Gemma 2 9B & 66.33 & 73.40 & 64.27 & 50.40 & 67.67 & 72.47 & 66.69 & 57.64 & 80.07 \\
                        Qwen 2 7B & 60.31 & 68.65 & 56.40 & 50.40 & 61.46 & 69.45 & 56.97 & 50.50 & 76.44 \\
                      Qwen 1.5 7B & 59.44 & 66.85 & 55.90 & 51.00 & 59.85 & 66.62 & 56.04 & 50.56 & 72.59 \\
\hline
              EMMA-500 Llama 2 7B & 63.11 & 66.60 & 62.57 & 52.40 & 66.38 & 68.92 & 65.73 & 61.32 & 72.80 \\
\bottomrule
\end{tabular}
\end{table*}

\subsection{Natural Language Inference}
\label{sec:NLI}

We evaluate on the XNLI (Cross-lingual Natural Language Inference) benchmark \citep{conneau2018xnli} where sentence pairs in different languages need to be classified as entailment, contradiction, or neutral.
We categorize the languages in XNLI into 3 groups, i.e., high-resource (German, English, Spanish, French, Russian, and Chinese), medium-resource (Arabic, Bulgarian, Greek, Hindi, Turkish, and Vietnamese), and low-resource (Swahili, Thai, and Urdu).\footnote{Again, the categorization is not perfect.}  
\Cref{tab:xnli_agg} shows the aggregated accuracy. 
Our model outperforms most baselines including Llama 2-based models and multilingual LLMs. We achieve the second-best average accuracy, slightly behind the Llama 3.1 8B model. On the low-resource end, we perform the second-best, slightly behind the Gemma 2 9B model. 

\begin{table}[ht!]
\centering
\caption{0-shot results on XNLI (ACC \%).}
\label{tab:xnli_agg}
\scriptsize
\setlength{\tabcolsep}{3pt}
\begin{tabular}{lrrrr}
\toprule
                            \ch{Model} &    \ch{Avg} &   \ch{High} &  \ch{Medium} &    \ch{Low} \\
\midrule
                       Llama 2 7B & 40.19 & 45.26 &   37.72 & 34.97 \\
                  Llama 2 7B Chat & 38.58 & 42.77 &   36.75 & 33.87 \\
                   CodeLlama 2 7B & 40.19 & 46.27 &   37.29 & 33.86 \\
                LLaMAX Llama 2 7B & 44.27 & 46.53 &   42.64 & 43.03 \\
         LLaMAX Llama 2 7B Alpaca & 45.09 & 48.47 &   42.80 & 42.89 \\
          MaLA-500 Llama 2 10B v1 & 38.11 & 42.10 &   35.85 & 34.65 \\
          MaLA-500 Llama 2 10B v2 & 38.11 & 42.10 &   35.85 & 34.65 \\
                  YaYi Llama 2 7B & 41.28 & 47.32 &   38.41 & 34.94 \\
             TowerBase Llama 2 7B & 39.84 & 46.08 &   36.33 & 34.39 \\
         TowerInstruct Llama 2 7B & 40.36 & 47.07 &   36.92 & 33.79 \\
\hline
         Occiglot Mistral 7B v0.1 & 42.35 & 49.90 &   38.39 & 35.19 \\
Occiglot Mistral 7B v0.1 Instruct & 40.81 & 47.58 &   37.18 & 34.52 \\
                         BLOOM 7B & 41.60 & 45.13 &   39.69 & 38.38 \\
                        BLOOMZ 7B & 37.13 & 40.02 &   35.56 & 34.51 \\
                             mGPT & 40.51 & 42.97 &   39.65 & 37.30 \\
                         mGPT 13B & 41.59 & 44.14 &   40.42 & 38.82 \\
                          YaYi 7B & 39.87 & 43.85 &   38.24 & 35.15 \\
\hline
                        Aya 23 8B & 43.12 & 48.51 &   41.95 & 34.67 \\
                   Aya expanse 8B & 45.56 & 50.48 &   44.24 & 38.38 \\
                       Llama 3 8B & 44.97 & 48.82 &   43.84 & 39.56 \\
                     Llama 3.1 8B & 45.62 & 49.61 &   44.04 & 40.83 \\
                         Gemma 7B & 42.58 & 46.44 &   41.00 & 38.01 \\
                       Gemma 2 9B & 46.74 & 48.50 &   45.11 & 46.49 \\
                        Qwen 2 7B & 42.77 & 47.31 &   41.35 & 36.53 \\
                      Qwen 1.5 7B & 39.47 & 40.95 &   38.80 & 37.83 \\
\hline
              EMMA-500 Llama 2 7B & 45.14 & 46.09 &   44.40 & 44.71 \\
\bottomrule
\end{tabular}
\end{table}

\subsection{Summarization}
\label{sec:summarization}

XL-Sum~\citep{hasan-etal-2021-xl} is a large-scale multilingual abstractive summarization dataset that covers 44 languages.
To evaluate the quality of the generated summaries, we use ROUGE-L~\citep{lin-2004-rouge} and BERTScore~\citep{zhang2019bertscore} as evaluation metrics.
ROUGE-L measures the longest common subsequence (LCS) between the reference and generated summaries. Recall and precision are calculated by dividing the LCS length by the reference length and the generated summary length, respectively. An F-score is then used to combine these two aspects into a single metric. BERTScore computes the semantic similarity between summaries by leveraging contextual embeddings from pre-trained language models. Specifically, we use the \texttt{bert-base-multilingual-cased}\footnote{\url{https://huggingface.co/google-bert/bert-base-multilingual-cased}} model for BERTScore to accommodate multiple languages.

The evaluation results are presented in \Cref{tab:xlsum}, we observe that our model, EMMA-500, performs comparably to other Llama2-based models like TowerInstruct Llama 2 7B. 
Comparing with recent advances, our EMMA-500 gets slightly better performance than Llama 3 and Gemma.
While EMMA-500 shows strength in certain language categories, it does not significantly outperform these models overall. This suggests that, although EMMA-500 is effective in multilingual summarization tasks, there is room for improvement to achieve more consistent performance across all languages. It is also important to note that this test set includes only 44 languages, limiting the evaluation of EMMA-500's capabilities in low-resource languages.

\begin{table*}[ht]
\caption{Results on XL-Sum (ROUGE-L/BERTScore \%).}
\label{tab:xlsum}
\centering
\scriptsize
\begin{tabular}{lrrrrrr}
\toprule
\ch{Model} & \ch{Avg} & \ch{High} & \ch{Medium-High} & \ch{Medium} & \ch{Medium-Low} & \ch{Low} \\
\midrule
Llama 2 7B & 7.11/66.52 & 7.35/66.79 & 6.65/66.28 & 6.50/64.67 & 9.73/61.90 & 8.22/71.32 \\
Llama 2 7B Chat & 8.84/68.44 & 9.48/68.48 & 8.27/68.15 & 7.57/66.82 & 12.14/64.16 & 10.18/73.54 \\
CodeLlama 2 7B & 7.15/65.74 & 7.28/65.08 & 6.80/65.90 & 6.37/62.96 & 9.93/63.54 & 8.26/71.32 \\
LLaMAX Llama 2 7B & 5.29/64.59 & 5.07/64.83 & 5.19/64.71 & 4.94/61.43 & 7.19/61.85 & 5.94/69.14 \\
LLaMAX Llama 2 7B Alpaca & 10.11/69.24 & 10.42/69.70 & 9.41/68.68 & 9.45/68.14 & 12.65/64.86 & 12.15/73.80 \\
MaLA-500 Llama 2 10B v1 & 5.45/63.96 & 5.97/64.03 & 5.21/63.88 & 4.77/61.12 & 7.05/63.83 & 5.59/68.16 \\
MaLA-500 Llama 2 10B v2 & 5.44/64.28 & 5.86/64.45 & 5.22/64.07 & 4.86/61.35 & 6.86/64.33 & 5.60/68.82 \\
Yayi Llama 2 7B & 7.98/67.21 & 8.67/67.39 & 7.48/67.29 & 7.05/65.59 & 10.95/60.55 & 8.55/71.44 \\
TowerBase Llama 2 7B & 7.65/67.09 & 7.92/67.26 & 7.25/67.06 & 6.64/65.02 & 9.90/63.33 & 9.16/71.21 \\
TowerInstruct Llama 2 7B & 8.89/68.46 & 9.44/68.60 & 8.59/68.51 & 7.64/66.28 & 11.90/63.71 & 9.45/72.89 \\
\midrule
Occiglot Mistral 7B v0.1 & 7.33/66.20 & 7.53/66.40 & 6.82/65.87 & 6.80/64.31 & 9.71/62.74 & 8.77/71.11 \\
Occiglot Mistral 7B v0.1 Instruct & 8.31/66.96 & 9.03/66.81 & 7.77/67.07 & 7.93/63.86 & 10.41/63.58 & 8.59/72.52 \\
BLOOM 7B & 6.99/64.78 & 6.78/65.06 & 6.97/64.77 & 6.12/61.89 & 9.79/58.16 & 7.61/70.93 \\
BLOOMZ 7B & 11.15/69.82 & 13.92/68.73 & 8.92/69.99 & 10.24/69.05 & 12.03/66.18 & 14.86/74.12 \\
mGPT & 3.85/59.74 & 4.82/59.91 & 3.60/60.10 & 2.35/56.02 & 6.81/59.22 & 3.67/63.30 \\
mGPT-13B & 4.96/62.20 & 5.85/62.02 & 4.59/62.54 & 3.80/58.97 & 7.92/59.13 & 4.90/66.99 \\
Yayi 7B & 12.06/69.74 & 13.98/68.93 & 10.50/69.49 & 11.05/68.91 & 12.45/65.87 & 15.29/75.26 \\
\midrule
Aya 23 8B & 8.68/66.79 & 10.03/67.66 & 8.24/66.26 & 6.80/68.73 & 12.63/63.26 & 8.47/65.68 \\
Aya expanse 8B & 10.51/68.73 & 11.61/68.56 & 9.66/68.54 & 10.21/71.02 & 12.88/65.38 & 10.90/68.02 \\
Llama 3 8B & 8.47/67.08 & 8.06/66.79 & 8.10/66.95 & 8.36/64.85 & 9.65/65.08 & 10.52/72.20 \\
Llama 3.1 8B & 8.57/66.97 & 8.01/66.41 & 8.40/66.84 & 8.48/65.21 & 9.05/65.80 & 10.45/71.68 \\
Gemma 2 9B & 7.38/65.45 & 7.24/64.00 & 6.87/65.81 & 7.47/62.95 & 9.22/63.57 & 8.89/71.46 \\
Gemma 7B & 6.70/64.52 & 6.86/63.48 & 6.24/64.91 & 6.63/67.21 & 6.73/59.86 & 8.30/63.35 \\
Qwen 1.5 7B & 9.58/69.13 & 10.61/68.78 & 8.58/68.89 & 9.61/67.83 & 10.79/64.92 & 10.74/74.40 \\
Qwen 2 7B & 10.18/69.35 & 11.58/69.36 & 9.31/68.98 & 9.63/68.03 & 10.64/64.82 & 11.15/74.47 \\
\midrule
EMMA-500 Llama 2 7B & 8.58/67.20 & 8.22/65.87 & 8.24/67.26 & 8.00/69.48 & 11.47/65.55 & 10.37/67.41 \\
\bottomrule
\end{tabular}
\end{table*}

\subsection{Math}
\label{sec:math}

We evaluate the ability of LLMs to solve grade-school math problems across multiple languages on the MGSM (Multilingual Grade School Math) benchmark \citep{shi2022language}. MGSM is an extension of the GSM8K (Grade School Math 8K) dataset \citep{cobbe2021training} by translating 250 of the original GSM8K problems into ten languages. 
We also split these ten languages into three groups, i.e., high-resource (Spanish, French, German, Russian, Chinese, and Japanese), medium-resource (Thai, Swahili, and Bengali), and low-resource (Telugu).
\Cref{tab:mgsm_agg} shows the results for 3-shot prompting with a flexible match to obtain the answers in model generation.
We evaluate all the models by directly prompting the questions (denoted as direct) and the questions with answers followed by Chain-of-Thoughts prompt in the same languages as the subset being evaluated (denoted as CoT) \citep{shi2022language}. 
The results show that Llama 2 7B is a weak model in the MGSM math task. The base model and its variants failed in most settings. 
Multilingual LLMs such as BLOOM, XGLM, and YaYi are also subpar at this task. 
Recent LLMs like Aya, Llama 3, Qwen, and Gemma obtain reasonable performance, and Qwen series models are the best.
Our model improves the Llama 2 7B model remarkably in both prompt strategies.
For direct prompting, our model has an average accuracy of 0.1702, 7\% higher than the Llama 2 7B chat model. 
For CoT-based prompting, our model increases the score of the Llama 2 7B chat model from 0.1 to 0.3 (20\% higher) and slightly outperforms Llama 3 and 3.1 8B models. 

\begin{table*}[ht!]
\centering
\caption{3-shot results (ACC \%) on MGSM obtained by direct and CoT prompting.}
\label{tab:mgsm_agg}
\scriptsize
\begin{tabular}{lrrrrrrrr}
\toprule
\multicolumn{1}{c}{\multirow[b]{2}{*}{\textbf{Model}}} & \multicolumn{4}{c}{\textbf{Direct Prompting}} & \multicolumn{4}{c}{\textbf{CoT Prompting}} \\
\cmidrule(lr){2-5}\cmidrule(lr){6-9}
&   \ch{Avg} &   \ch{High} &  \ch{Medium} &    \ch{Low} &   \ch{Avg} &   \ch{High} &  \ch{Medium} &    \ch{Low} \\
\midrule
                       Llama 2 7B &  6.69 &  8.07 &    2.13 &  1.20 &  6.36 &  7.60 &    2.13 &  0.80 \\
                  Llama 2 7B Chat & 10.22 & 13.73 &    2.13 &  0.80 & 10.91 & 13.53 &    2.80 &  1.60 \\
                   CodeLlama 2 7B &  5.93 &  7.07 &    2.93 &  1.20 &  6.64 &  8.73 &    2.67 &  2.00 \\
                LLaMAX Llama 2 7B &  3.35 &  4.00 &    2.00 &  0.80 &  3.62 &  4.33 &    2.27 &  2.40 \\
         LLaMAX Llama 2 7B Alpaca &  5.05 &  5.20 &    4.00 &  1.60 &  6.35 &  8.07 &    4.13 &  0.80 \\
          MaLA-500 Llama 2 10B v1 &  0.91 &  1.33 &    0.27 &  0.00 &  0.73 &  1.27 &    0.00 &  0.00 \\
          MaLA-500 Llama 2 10B v2 &  0.91 &  1.33 &    0.27 &  0.00 &  0.73 &  1.27 &    0.00 &  0.00 \\
                  YaYi Llama 2 7B &  7.09 &  9.47 &    1.47 &  1.20 &  7.22 &  8.73 &    2.27 &  1.20 \\
             TowerBase Llama 2 7B &  6.15 &  8.33 &    1.73 &  0.80 &  6.16 &  8.60 &    2.40 &  0.80 \\
         TowerInstruct Llama 2 7B &  7.24 &  9.53 &    1.73 &  2.00 &  8.24 & 10.47 &    1.87 &  1.20 \\
\hline
         Occiglot Mistral 7B v0.1 & 13.31 & 16.87 &    4.53 &  1.60 & 14.07 & 18.80 &    3.60 &  1.20 \\
Occiglot Mistral 7B v0.1 Instruct & 22.76 & 29.80 &    7.47 &  2.80 & 22.16 & 30.40 &    7.87 &  2.80 \\
                         BLOOM 7B &  2.87 &  2.60 &    2.80 &  3.60 &  2.29 &  2.20 &    1.47 &  2.00 \\
                        BLOOMZ 7B &  2.55 &  2.67 &    2.40 &  1.20 &  2.15 &  1.67 &    3.07 &  2.00 \\
                             mGPT &  1.35 &  1.67 &    0.53 &  0.00 &  1.42 &  1.93 &    0.67 &  0.00 \\
                         mGPT 13B &  1.31 &  1.80 &    0.67 &  0.00 &  1.53 &  1.67 &    1.20 &  0.00 \\
                          YaYi 7B &  2.76 &  2.93 &    1.47 &  2.40 &  3.02 &  2.93 &    2.40 &  2.00 \\
\hline
                        Aya 23 8B & 22.29 & 30.67 &    3.47 &  0.80 & 24.71 & 35.07 &    5.47 &  2.40 \\
                   Aya expanse 8B & 43.02 & 55.47 &   18.93 &  5.20 & 41.45 & 54.67 &   18.53 &  5.20 \\
                       Llama 3 8B & 27.45 & 27.87 &   26.13 &  5.60 & 28.13 & 28.53 &   26.67 &  5.20 \\
                     Llama 3.1 8B & 28.36 & 29.00 &   26.13 &  4.40 & 27.31 & 27.27 &   25.47 &  8.40 \\
                         Gemma 7B & 38.22 & 36.60 &   38.27 & 27.20 & 35.78 & 34.67 &   37.07 & 26.80 \\
                       Gemma 2 9B & 32.95 & 28.00 &   35.73 & 30.80 & 44.69 & 36.07 &   52.00 & 47.20 \\
                        Qwen 2 7B & 48.95 & 54.40 &   38.80 & 14.80 & 51.47 & 58.93 &   39.07 & 14.40 \\
                      Qwen 1.5 7B & 31.56 & 40.00 &   16.00 &  4.00 & 30.36 & 40.60 &   14.80 &  2.40 \\
\hline
              EMMA-500 Llama 2 7B & 17.02 & 19.20 &   11.87 &  2.40 & 18.09 & 20.00 &   13.20 &  2.80 \\
\bottomrule
\end{tabular}
\end{table*}

\subsection{Machine Reading Comprehension}
\label{sec:MRC}

BELEBELE \citep{bandarkar2023belebele} is a machine reading comprehension dataset covering 122 languages including high- and low-resource languages. Each question offers four multiple-choice answers based on a short passage from the FLORES-200 dataset. 
This benchmark is very challenging, even the English version of it presents remarkable challenges for advanced models. 
\Cref{tab:belebele_agg} shows the zero-shot results in different resource groups. 
Our continual pre-training improves the Llama 2 7B base model. But Llama 2 7 B-based models mostly fail in this task and get quasi-random results. 
Recent advanced models like Aya expanse, Llama 3.1 and Qwen 2 get reasonable results. 

\begin{table*}[ht!]
\centering
\caption{0-shot results (ACC \%) on BELEBELE.}
\label{tab:belebele_agg}
\scriptsize
\begin{tabular}{lrrrrrrr}
\toprule
                            \ch{Model} &    \ch{Avg} &   \ch{High} &  \ch{Medium-High} &  \ch{Medium} &  \ch{Medium-Low} &    \ch{Low}  \\
\midrule
                       Llama 2 7B & 26.27 & 26.76 &        26.35 &   26.07 &       26.41 & 26.27 \\
                  Llama 2 7B Chat & 29.05 & 31.84 &        29.97 &   28.95 &       29.47 & 29.09 \\
                   CodeLlama 2 7B & 27.38 & 27.37 &        27.33 &   27.30 &       27.30 & 27.38 \\
                LLaMAX Llama 2 7B & 23.09 & 23.23 &        23.15 &   23.07 &       23.10 & 23.08 \\
         LLaMAX Llama 2 7B Alpaca & 24.48 & 25.46 &        24.82 &   24.41 &       24.60 & 24.49 \\
          MaLA-500 Llama 2 10B v1 & 22.96 & 23.02 &        22.98 &   22.97 &       22.98 & 22.97 \\
          MaLA-500 Llama 2 10B v2 & 22.96 & 23.02 &        22.98 &   22.97 &       22.98 & 22.97 \\
                  YaYi Llama 2 7B & 28.32 & 29.64 &        28.67 &   28.11 &       28.37 & 28.26 \\
             TowerBase Llama 2 7B & 26.36 & 27.43 &        26.85 &   26.29 &       26.48 & 26.34 \\
         TowerInstruct Llama 2 7B & 27.93 & 29.88 &        28.51 &   27.57 &       28.19 & 27.92 \\
\hline
         Occiglot Mistral 7B v0.1 & 30.16 & 32.25 &        30.94 &   30.02 &       30.40 & 30.15 \\
Occiglot Mistral 7B v0.1 Instruct & 32.05 & 34.14 &        32.62 &   31.74 &       32.40 & 32.08 \\
                         BLOOM 7B & 24.11 & 24.25 &        24.52 &   24.12 &       24.11 & 24.08 \\
                        BLOOMZ 7B & 39.32 & 45.43 &        43.67 &   41.51 &       40.08 & 39.51 \\
                             mGPT & 23.96 & 24.01 &        23.81 &   24.00 &       23.94 & 23.98 \\
                          YaYi 7B & 37.97 & 44.37 &        42.71 &   40.49 &       38.72 & 38.09 \\
\hline
                        Aya 23 8B & 40.08 & 43.85 &        41.71 &   39.22 &       40.93 & 39.81 \\
                   Aya expanse 8B & 46.98 & 52.22 &        48.99 &   46.57 &       48.36 & 46.93 \\
                       Llama 3 8B & 40.73 & 46.07 &        42.92 &   41.03 &       41.87 & 40.88 \\
                     Llama 3.1 8B & 45.19 & 52.50 &        48.01 &   45.65 &       46.74 & 45.34 \\
                         Gemma 7B & 43.37 & 52.63 &        47.83 &   44.82 &       45.43 & 43.94 \\
                       Gemma 2 9B & 54.49 & 64.10 &        58.90 &   55.83 &       56.85 & 55.05 \\
                        Qwen 2 7B & 49.31 & 57.62 &        52.20 &   50.04 &       51.16 & 49.48 \\
                      Qwen 1.5 7B & 41.83 & 48.86 &        44.79 &   42.18 &       43.00 & 41.78 \\
\hline
              EMMA-500 Llama 2 7B & 26.75 & 28.32 &        28.18 &   27.58 &       27.14 & 26.94 \\
\bottomrule
\end{tabular}
\end{table*}

We then move to a more challenging task, the ARC multilingual test, which is a machine-translated benchmark~\citep{lai2023okapi} from the ARC dataset \citep{Clark2018ThinkYH} that contains English science exam questions for multiple grade levels. 
We test the five-shot performance, with results shown in \Cref{tab:arc_agg}.
The evaluation results show a similar pattern to BELEBELE that EMMA-500 improves Llama 2 but the 7B model is not capable of this challenging task, all Llama2 7B-based models obtain close to random results, and recent advances like Llama 3, Gemma, and Qwen get much better results. 

\begin{table}[ht!]
\centering
\caption{5-shot results (ACC \%) on ARC multilingual.}
\label{tab:arc_agg}
\scriptsize
\setlength{\tabcolsep}{3pt}
\begin{tabular}{lrrrr}
\toprule
                            \ch{Model} &    \ch{Avg} &   \ch{High} &  \ch{Medium} &    \ch{Low} \\
\midrule
                       Llama 2 7B & 27.5608 & 33.1244 & 27.3065 & 21.0196 \\
                  Llama 2 7B Chat & 28.0155 & 33.6893 & 27.7891 & 21.2910 \\
                   CodeLlama 2 7B & 25.2283 & 28.8639 & 24.6368 & 21.6450 \\
                LLaMAX Llama 2 7B & 26.0905 & 30.0022 & 25.9174 & 21.4821 \\
         LLaMAX Llama 2 7B Alpaca & 31.0602 & 36.8898 & 31.8518 & 22.4866 \\
          MaLA-500 Llama 2 10B v1 & 21.1612 & 21.9174 & 20.4836 & 21.3172 \\
          MaLA-500 Llama 2 10B v2 & 21.1612 & 21.9174 & 20.4836 & 21.3172 \\
                  YaYi Llama 2 7B & 28.3975 & 34.2968 & 28.3460 & 21.1068 \\
             TowerBase Llama 2 7B & 27.9355 & 35.3215 & 26.8248 & 20.5081 \\
         TowerInstruct Llama 2 7B & 30.1019 & 38.8803 & 28.8544 & 21.1562 \\
\hline
         Occiglot Mistral 7B v0.1 & 29.7667 & 38.3930 & 28.5078 & 21.0296 \\
Occiglot Mistral 7B v0.1 Instruct & 30.8844 & 40.2927 & 29.6522 & 21.1264 \\
                         BLOOM 7B & 23.6501 & 26.2685 & 22.7178 & 21.8920 \\
                        BLOOMZ 7B & 23.9497 & 26.9360 & 22.7439 & 22.1762 \\
                             mGPT & 20.2422 & 20.1129 & 19.6471 & 21.3709 \\
                         mGPT 13B & 21.7584 & 22.9872 & 21.1366 & 21.2329 \\
                          YaYi 7B & 24.4413 & 27.9617 & 23.2904 & 21.9109 \\
\hline
                        Aya 23 8B & 31.0838 & 40.0507 & 30.0173 & 21.6080 \\
                   Aya expanse 8B & 36.5617 & 47.8696 & 36.1507 & 23.0948 \\
                       Llama 3 8B & 34.7957 & 42.4323 & 35.5256 & 24.0638 \\
                     Llama 3.1 8B & 34.9336 & 42.4323 & 35.8939 & 23.9995 \\
                         Gemma 7B & 38.6827 & 46.4619 & 40.4660 & 26.0609 \\
                       Gemma 2 9B & 44.1534 & 54.5890 & 46.1756 & 27.8228 \\
                        Qwen 2 7B & 33.8204 & 43.8754 & 32.6423 & 23.1660 \\
                      Qwen 1.5 7B & 28.9263 & 35.5527 & 28.1392 & 21.9223 \\
\hline
              EMMA-500 Llama 2 7B & 29.5282 & 34.1004 & 29.8165 & 23.3444 \\
\bottomrule
\end{tabular}
\end{table}

\subsection{Code Generation}
\label{sec:code_generation}

We conduct code generation evaluations on the Multipl-E~\citep{Cassano2022MultiPLEAS} benchmark in the interest of measuring the effects of massively multilingual continual pre-training on a model's code generation utility and detecting if any catastrophic forgetting~\citep{luo2023catastrophic} has occurred on this front. Importantly, this also has implications for a model's reasoning~\citep{yang2024wand} and entity-tracking abilities~\citep{kim2024entity}.

Table~\ref{tab:me_agg} outlines comparisons against strong, openly available multilingual models such as Qwen~\citep{yang2024qwen2technicalreport} and Aya~\citep{ustun2024aya} along with other continually pre-trained models. Our main takeaway is that by carefully curating high-quality code data as part of the data mix, it is possible to not only avoid the catastrophic forgetting that has plagued existing continually pre-trained models (MaLA-500, LLaMAX and TowerBase) but also surpass the base model itself. However, the pre-training phase is still the most important, and closing the gap with stronger base models like Qwen and Gemma remains elusive.

\begin{table}[ht!]
\centering
\caption{Results on Multipl-E. For language-level breakdowns, refer to \Cref{tab:me_p1,tab:me_p10,tab:me_p25} in the appendix.}
\label{tab:me_agg}
\scriptsize
\setlength{\tabcolsep}{3pt}
\begin{tabular}{lrrr}
\toprule
    \ch{Model} & \ch{Avg Pass@1} & \ch{Avg Pass@10} & \ch{Avg Pass@25} \\
\midrule
    Llama 2 7B & 8.92\% & 19.45\% & 25.68\% \\
    CodeLlama 2 7B & 28.43\% & 50.83\% & 63.92\% \\
    LLaMAX 2 7B & 0.35\% & 1.61\% & 2.67\% \\
    MaLA-500 Llama 2 10B V2 & 0.0\% & 0.0\% & 0.0\% \\
    TowerBase Llama 2 7B & 3.61\% & 6.65\% & 8.97\% \\
\midrule
    Occiglot Mistral 7B v0.1 & 21.26\% & 31.37\% & 45.86\% \\ 
    Bloom 7B & 5.34\% & 10.49\% & 14.65\% \\
    BloomZ 2 7B & 5.85\% & 11.40\% & 15.76\% \\
    Aya23 8B & 9.19\% & 23.52\% & 32.09\% \\
\midrule
    Mistral 7B v0.3 & 26.10\% & 48.68\% & 59.05\% \\
    Llama 3 8B & 30.09\% & 53.82\% & 64.01\% \\
    LLaMAX 3 8B & 3.00\% & 7.23\% & 10.67\% \\
    Gemma 7B & 28.55\% & 54.27\% & 64.75\% \\
    CodeGemma 7B & 31.51\% & 63.13\% & 72.65\% \\
    Qwen 1.5 7B & 21.05\% & 37.19\% & 47.63\% \\
    Qwen 2 7B & 38.68\% & 62.63\% & 71.55\% \\
\midrule
    EMMA-500 Llama 2 7B & 11.38\% & 19.02\% & 26.16\% \\

\bottomrule
\end{tabular}
\end{table}

\section{Related Work}
\label{sec:related}

\paragraph{Multilingual LLMs}

Multilingual large language models (LLMs) have made significant progress in processing and understanding multiple languages within a unified framework. Models like mT5 \citep{xue2021mt5} and XGLM \citep{lin2022few} leverage both monolingual and multilingual datasets to perform tasks such as translation and text summarization across a wide spectrum of languages. However, the predominant focus on English has led to disparities in performance, particularly for low-resource languages. Recent work on multilingual LLMs, such as BLOOM \citep{scao2022bloom}, has shown that adapting these English-centric models through vocabulary extension based on multilingual corpora and continual pre-training (CPT) can improve performance across languages, especially low-resource ones. These models highlight the importance of efficient tokenization and adaptation, which can bridge the performance gap between high-resource and low-resource languages.

\paragraph{Multilingual Corpora}

The availability and use of multilingual corpora play a crucial role in training multilingual LLMs. CC100 Corpus \citep{conneau2020unsupervised}, launched in 2020, encompasses hundreds of billions of tokens and over 100 languages. Further, CC100-XL Corpus \citep{lin2022few}, created for the training of XGLM, extends across 68 Common Crawl Snapshots and 134 languages, aiming to balance language presentation and improve performance in few-shot and zero-shot tasks. The ROOTS Corpus \citep{laurenccon2022bigscience}, released in July 2022, supports BLOOM with approximately 341 billion tokens across 46 natural languages. It emphasizes underrepresented languages such as Swahili and Catalan, drawing from diverse sources including web crawls, books and academic publications. Besides, Occiglot Fineweb \footnote{\url{https://occiglot.eu/posts/occiglot-fineweb/}}, which began to be released in early 2024, consists of around 230 million documents in 10 European languages. It combines curated and cleaned web data to support efficient training for both high- and low-resource European languages. Additionally, recent efforts parallel corpus construction from web crawls, such as ParaCrawl \citep{banon-etal-2020-paracrawl} and CCMatrix \citep{schwenk-etal-2021-ccmatrix}, have contributed to large-scale multilingual training too.

\paragraph{Continual Pre-training}

Continual pre-training has become a popular strategy to adapt LLMs to new languages and domains without retraining from scratch. The process involves updating model parameters incrementally using a relatively small amount of new data from target languages. Recent studies, such as those by \cite{tejaswi2024exploringdesignchoicesbuilding}, have demonstrated the effectiveness of CPT in improving the adaptability of LLMs in low-resource language settings. Techniques such as vocabulary augmentation and targeted domain-specific pre-training have been shown to significantly improve both efficiency and performance, particularly when large multilingual models are adapted to new languages. Despite its benefits, CPT can lead to catastrophic forgetting, where models lose previously learned information. To address the potential degraded performance issue, \cite{ibrahim2024simplescalablestrategiescontinually} presents a simple yet effective approach to continual pre-train models, demonstrating that with a combination of learning rate re-warming, re-decaying, and replay of previous data, it is possible to match the performance of fully re-trained models.  Also, recent studies have delved into the effectiveness of continual pre-training with parallel data. As highlighted in the study by \cite{gilabert2024investigatingtranslationcapabilitieslarge}, the use of a Catalan-centric parallel dataset has enabled the training of models good at trasnalting in various directions.  Besides, the research by \cite{kondo2024enhancingtranslationaccuracylarge}, proposed a two-phase continual training approach with parallel data. In the first phase, a pre-trained LLM is continually pre-trained on parallel data, followed by a second phase of supervised fine-tuning with a small amount of high-quality parallel data. Their experiments with a 3.8B-parameter model across various data formats revealed that alternating between source and target sentences during continual pre-training is crucial for enhancing translation accuracy in the corresponding direction.

\section{Conclusion}
\label{sec:conclusion}
This paper addresses critical advancements and challenges in adapting language models to more than 500 languages, focusing on enhancing their performance across diverse languages.
We compile the MaLA corpus, a multilingual dataset for continual pre-training of multilingual language models.
By expanding and augmenting existing corpora, we train the EMMA-500 model.
It demonstrates notable improvements in a range of tasks such as next-token prediction, commonsense reasoning, machine translation, and text classification, showing remarkable improvements in low-resource languages.
Our results show that a well-curated, massively multilingual corpora can advance model capabilities. 
This work sets a new benchmark for inclusive and effective multilingual language models and paves the way for future research to address the disparities between high-resource and low-resource languages.

\section*{Limitations}

\paragraph{Multilingual Benchmark} Multilingual language models are typically designed to serve users of different languages, which reflect the diverse cultural and linguistic backgrounds of their speakers. However, many available multilingual benchmarks, including some used in this study, have been created through human or machine translation. They tend to feature topics and knowledge primarily from English-speaking communities and carry imperfections due to translation, which affects the integrity of LLM evaluation \citep{chen2024good}. We highlight this discrepancy and call for collaborative efforts to develop large-scale natively-created multilingual test sets that more accurately assess the breadth of languages and cultures these models aim to cater to.

\paragraph{Human Evaluation} While human evaluation is valuable, it comes with challenges such as subjectivity, inconsistency, and high costs, especially in a massively multilingual setting. Recruiting linguistically skilled annotators is often impractical, particularly for low-resource languages, making large-scale human evaluation unsustainable. Even assessing a subset of languages remains prohibitively expensive. While our work is constrained by these limitations, we recognize the importance of human evaluation in complementing automated metrics. However, given these challenges, we choose automatic tools rather than human evaluation to provide a more scalable and consistent alternative, despite their imperfections. 

\paragraph{Model Performance} Our EMMA-500 model demonstrates enhanced multilingual performance relative to its base model, Llama 2 7B, as well as other continually pre-trained variants. However, it does not surpass some of the latest models, such as Llama 3, Gemma 2, and Qwen 2, on certain benchmarks. These models are likely to be a better starting point for continual pre-training compared to Llama 2 due to their larger pre-training data. We intend to pursue multilingual extension based on these newer models in the future. On the other hand, it would be practical to assess the released corpus directly during pre-training from scratch. 
Moreover, multilingual instruction tuning to prepare these models for better task performance and natural interactions would be useful.
Despite its strong multilingual capabilities, our model still faces challenges in mathematical reasoning and machine reading comprehension. As evaluated on a machine-translated benchmark, its performance on math tasks remains limited, likely due to the inherent difficulty of numerical reasoning and the potential impact of translation artifacts on problem clarity. Similarly, while our model shows improvements in many NLP tasks, it struggles with machine comprehension, which requires deep contextual understanding and reasoning. These limitations highlight areas for future improvement, such as incorporating specialized training data or exploring alternative architectures better suited for these tasks.

\paragraph{Real-world Usage}
This research focuses on expanding multilingual corpora and enhancing a model's multilingual capabilities through continual pre-training. However, it is not yet ready for real-world deployment. The model has not undergone human preference alignment or red-teaming for safety and robustness. While our work makes progress in improving multilingual NLP, further refinement, including alignment with human preferences and thorough adversarial testing, would be necessary before practical deployment.

\section*{Acknowledgment}
The work has received funding from the European Union's Horizon Europe research and innovation programme under grant agreement No 101070350 and from UK Research and Innovation (UKRI) under the UK government's Horizon Europe funding guarantee [grant number 10052546], and funding from UTTER's Financial Support for Third Parties under the EU's Horizon Europe Research and Innovation Actions (UTTER, contract 101070631).
The authors wish to acknowledge CSC - IT Center for Science, Finland, the Leonardo and LUMI supercomputers, owned by the EuroHPC Joint Undertaking, for providing computational resources.
We thank Indraneil Paul for his contributions to preparing the code data and evaluating our model on code generation tasks. 
\bibliography{ref-mala-lm}

\begin{thebibliography}{142}
\providecommand{\natexlab}[1]{#1}

\bibitem[{Aakanksha et~al.(2024)Aakanksha, Ahmadian, Ermis, Goldfarb-Tarrant,
  Kreutzer, Fadaee, and Hooker}]{aakanksha2024aya_redteaming}
Aakanksha, Arash Ahmadian, Beyza Ermis, Seraphina Goldfarb-Tarrant, Julia
  Kreutzer, Marzieh Fadaee, and Sara Hooker. 2024.
\newblock \href {https://arxiv.org/abs/2406.18682} {The multilingual alignment
  prism: Aligning global and local preferences to reduce harm}.
\newblock \emph{Preprint}, arXiv:2406.18682.

\bibitem[{Abate et~al.(2018)Abate, Melese, Tachbelie, Meshesha, Atinafu,
  Mulugeta, Assabie, Abera, Ephrem, Abebe, Tsegaye, Lemma, Andargie, and
  Shifaw}]{abate-etal-2018-parallel}
Solomon~Teferra Abate, Michael Melese, Martha~Yifiru Tachbelie, Million
  Meshesha, Solomon Atinafu, Wondwossen Mulugeta, Yaregal Assabie, Hafte Abera,
  Binyam Ephrem, Tewodros Abebe, Wondimagegnhue Tsegaye, Amanuel Lemma, Tsegaye
  Andargie, and Seifedin Shifaw. 2018.
\newblock \href {https://aclanthology.org/C18-1262} {Parallel corpora for
  bi-lingual {E}nglish-{E}thiopian languages statistical machine translation}.
\newblock In \emph{Proceedings of the 27th International Conference on
  Computational Linguistics}.

\bibitem[{Abdelali et~al.(2021)Abdelali, Mubarak, Samih, Hassan, and
  Darwish}]{abdelali-etal-2021-qadi}
Ahmed Abdelali, Hamdy Mubarak, Younes Samih, Sabit Hassan, and Kareem Darwish.
  2021.
\newblock \href {https://aclanthology.org/2021.wanlp-1.1} {{QADI}: {A}rabic
  dialect identification in the wild}.
\newblock In \emph{Proceedings of the Sixth Arabic Natural Language Processing
  Workshop}.

\bibitem[{Abu~Kwaik et~al.(2018)Abu~Kwaik, Saad, Chatzikyriakidis, and
  Dobnik}]{abu-kwaik-etal-2018-shami}
Kathrein Abu~Kwaik, Motaz Saad, Stergios Chatzikyriakidis, and Simon Dobnik.
  2018.
\newblock \href {https://aclanthology.org/L18-1576} {{S}hami: A corpus of
  {L}evantine {A}rabic dialects}.
\newblock In \emph{Proceedings of the Eleventh International Conference on
  Language Resources and Evaluation ({LREC} 2018)}.

\bibitem[{Adelani et~al.(2022)Adelani, Alabi, Fan, Kreutzer, Shen, Reid,
  Ruiter, Klakow, Nabende, Chang, Gwadabe, Sackey, Dossou, Emezue, Leong,
  Beukman, Muhammad, Jarso, Yousuf, Niyongabo~Rubungo, Hacheme, Wairagala,
  Nasir, Ajibade, Ajayi, Gitau, Abbott, Ahmed, Ochieng, Aremu, Ogayo, Mukiibi,
  Ouoba~Kabore, Kalipe, Mbaye, Tapo, Memdjokam~Koagne, Munkoh-Buabeng, Wagner,
  Abdulmumin, Awokoya, Buzaaba, Sibanda, Bukula, and
  Manthalu}]{adelani-etal-2022-thousand}
David Adelani, Jesujoba Alabi, Angela Fan, Julia Kreutzer, Xiaoyu Shen, Machel
  Reid, Dana Ruiter, Dietrich Klakow, Peter Nabende, Ernie Chang, Tajuddeen
  Gwadabe, Freshia Sackey, Bonaventure F.~P. Dossou, Chris Emezue, Colin Leong,
  Michael Beukman, Shamsuddeen Muhammad, Guyo Jarso, Oreen Yousuf, Andre
  Niyongabo~Rubungo, Gilles Hacheme, Eric~Peter Wairagala, Muhammad~Umair
  Nasir, Benjamin Ajibade, Tunde Ajayi, Yvonne Gitau, Jade Abbott, Mohamed
  Ahmed, Millicent Ochieng, Anuoluwapo Aremu, Perez Ogayo, Jonathan Mukiibi,
  Fatoumata Ouoba~Kabore, Godson Kalipe, Derguene Mbaye, Allahsera~Auguste
  Tapo, Victoire Memdjokam~Koagne, Edwin Munkoh-Buabeng, Valencia Wagner, Idris
  Abdulmumin, Ayodele Awokoya, Happy Buzaaba, Blessing Sibanda, Andiswa Bukula,
  and Sam Manthalu. 2022.
\newblock \href {https://doi.org/10.18653/v1/2022.naacl-main.223} {A few
  thousand translations go a long way! leveraging pre-trained models for
  {A}frican news translation}.
\newblock In \emph{Proceedings of the 2022 Conference of the North American
  Chapter of the Association for Computational Linguistics: Human Language
  Technologies}.

\bibitem[{Adelani et~al.(2021)Adelani, Ruiter, Alabi, Adebonojo, Ayeni,
  Adeyemi, Awokoya, and Espa{\~n}a-Bonet}]{adelani-etal-2021-effect}
David Adelani, Dana Ruiter, Jesujoba Alabi, Damilola Adebonojo, Adesina Ayeni,
  Mofe Adeyemi, Ayodele~Esther Awokoya, and Cristina Espa{\~n}a-Bonet. 2021.
\newblock \href {https://aclanthology.org/2021.mtsummit-research.6} {The effect
  of domain and diacritics in {Y}oruba{--}{E}nglish neural machine
  translation}.
\newblock In \emph{Proceedings of Machine Translation Summit XVIII: Research
  Track}.

\bibitem[{Adelani et~al.(2023)Adelani, Liu, Shen, Vassilyev, Alabi, Mao, Gao,
  and Lee}]{sib-200}
David~Ifeoluwa Adelani, Hannah Liu, Xiaoyu Shen, Nikita Vassilyev, Jesujoba~O.
  Alabi, Yanke Mao, Haonan Gao, and En{-}Shiun~Annie Lee. 2023.
\newblock \href {https://doi.org/10.48550/arXiv.2309.07445} {{SIB-200:} {A}
  simple, inclusive, and big evaluation dataset for topic classification in
  200+ languages and dialects}.
\newblock \emph{CoRR}.

\bibitem[{Agerri et~al.(2018)Agerri, G{\'o}mez~Guinovart, Rigau, and
  Solla~Portela}]{agerri-etal-2018-developing}
Rodrigo Agerri, Xavier G{\'o}mez~Guinovart, German Rigau, and Miguel~Anxo
  Solla~Portela. 2018.
\newblock \href {https://aclanthology.org/L18-1367} {Developing new linguistic
  resources and tools for the {G}alician language}.
\newblock In \emph{Proceedings of the Eleventh International Conference on
  Language Resources and Evaluation ({LREC} 2018)}.

\bibitem[{Alsarsour et~al.(2018)Alsarsour, Mohamed, Suwaileh, and
  Elsayed}]{alsarsour-etal-2018-dart}
Israa Alsarsour, Esraa Mohamed, Reem Suwaileh, and Tamer Elsayed. 2018.
\newblock \href {https://aclanthology.org/L18-1579} {{DART}: A large dataset of
  dialectal {A}rabic tweets}.
\newblock In \emph{Proceedings of the Eleventh International Conference on
  Language Resources and Evaluation ({LREC} 2018)}.

\bibitem[{Alves et~al.(2024)Alves, Pombal, Guerreiro, Martins, Alves, Farajian,
  Peters, Rei, Fernandes, Agrawal et~al.}]{alves2024tower}
Duarte~M Alves, Jos{\'e} Pombal, Nuno~M Guerreiro, Pedro~H Martins, Jo{\~a}o
  Alves, Amin Farajian, Ben Peters, Ricardo Rei, Patrick Fernandes, Sweta
  Agrawal, et~al. 2024.
\newblock \href {https://arxiv.org/abs/2402.17733} {Tower: An open multilingual
  large language model for translation-related tasks}.
\newblock \emph{arXiv preprint arXiv:2402.17733}.

\bibitem[{Anastasopoulos et~al.(2020)Anastasopoulos, Cattelan, Dou, Federico,
  Federmann, Genzel, Guzm{\'a}n, Hu, Hughes, Koehn, Lazar, Lewis, Neubig, Niu,
  {\"O}ktem, Paquin, Tang, and Tur}]{anastasopoulos-etal-2020-tico}
Antonios Anastasopoulos, Alessandro Cattelan, Zi-Yi Dou, Marcello Federico,
  Christian Federmann, Dmitriy Genzel, Franscisco Guzm{\'a}n, Junjie Hu,
  Macduff Hughes, Philipp Koehn, Rosie Lazar, Will Lewis, Graham Neubig,
  Mengmeng Niu, Alp {\"O}ktem, Eric Paquin, Grace Tang, and Sylwia Tur. 2020.
\newblock \href {https://doi.org/10.18653/v1/2020.nlpcovid19-2.5} {{TICO}-19:
  the translation initiative for {CO}vid-19}.
\newblock In \emph{Proceedings of the 1st Workshop on {NLP} for {COVID}-19
  (Part 2) at {EMNLP} 2020}.

\bibitem[{Andonian et~al.(2023)Andonian, Anthony, Biderman, Black, Gali, Gao,
  Hallahan, Levy-Kramer, Leahy, Nestler, Parker, Pieler, Phang, Purohit,
  Schoelkopf, Stander, Songz, Tigges, Th{\'e}rien, Wang, and
  Weinbach}]{gpt-neox-library}
Alex Andonian, Quentin Anthony, Stella Biderman, Sid Black, Preetham Gali, Leo
  Gao, Eric Hallahan, Josh Levy-Kramer, Connor Leahy, Lucas Nestler, Kip
  Parker, Michael Pieler, Jason Phang, Shivanshu Purohit, Hailey Schoelkopf,
  Dashiell Stander, Tri Songz, Curt Tigges, Benjamin Th{\'e}rien, Phil Wang,
  and Samuel Weinbach. 2023.
\newblock \href {https://doi.org/10.5281/zenodo.5879544} {{GPT-NeoX: Large
  Scale Autoregressive Language Modeling in PyTorch}}.

\bibitem[{Aryabumi et~al.(2024)Aryabumi, Dang, Talupuru, Dash, Cairuz, Lin,
  Venkitesh, Smith, Campos, Tan et~al.}]{aryabumi2024aya23}
Viraat Aryabumi, John Dang, Dwarak Talupuru, Saurabh Dash, David Cairuz, Hangyu
  Lin, Bharat Venkitesh, Madeline Smith, Jon~Ander Campos, Yi~Chern Tan, et~al.
  2024.
\newblock Aya 23: Open weight releases to further multilingual progress.
\newblock \emph{arXiv preprint arXiv:2405.15032}.

\bibitem[{Aulamo et~al.(2023)Aulamo, Bogoychev, Ji, Nail,
  Ram{\'\i}rez-S{\'a}nchez, Tiedemann, Van Der~Linde, and
  Zaragoza}]{aulamo2023hplt}
Mikko Aulamo, Nikolay Bogoychev, Shaoxiong Ji, Graeme Nail, Gema
  Ram{\'\i}rez-S{\'a}nchez, J{\"o}rg Tiedemann, Jelmer Van Der~Linde, and Jaume
  Zaragoza. 2023.
\newblock \href {https://aclanthology.org/2023.eamt-1.61/} {Hplt: High
  performance language technologies}.
\newblock In \emph{Proceedings of the 24th Annual Conference of the European
  Association for Machine Translation}.

\bibitem[{Bafna(2022)}]{bafna2022empirical}
Niyati Bafna. 2022.
\newblock Empirical models for an indic language continuum.

\bibitem[{Bandarkar et~al.(2023)Bandarkar, Liang, Muller, Artetxe, Shukla,
  Husa, Goyal, Krishnan, Zettlemoyer, and Khabsa}]{bandarkar2023belebele}
Lucas Bandarkar, Davis Liang, Benjamin Muller, Mikel Artetxe, Satya~Narayan
  Shukla, Donald Husa, Naman Goyal, Abhinandan Krishnan, Luke Zettlemoyer, and
  Madian Khabsa. 2023.
\newblock \href {https://aclanthology.org/2024.acl-long.44/} {The {BELEBELE}
  benchmark: a parallel reading comprehension dataset in 122 language
  variants}.
\newblock \emph{arXiv preprint arXiv:2308.16884}.

\bibitem[{Ba{\~n}{\'o}n et~al.(2020)Ba{\~n}{\'o}n, Chen, Haddow, Heafield,
  Hoang, Espl{\`a}-Gomis, Forcada, Kamran, Kirefu, Koehn, Ortiz~Rojas,
  Pla~Sempere, Ram{\'\i}rez-S{\'a}nchez, Sarr{\'\i}as, Strelec, Thompson,
  Waites, Wiggins, and Zaragoza}]{banon-etal-2020-paracrawl}
Marta Ba{\~n}{\'o}n, Pinzhen Chen, Barry Haddow, Kenneth Heafield, Hieu Hoang,
  Miquel Espl{\`a}-Gomis, Mikel~L. Forcada, Amir Kamran, Faheem Kirefu, Philipp
  Koehn, Sergio Ortiz~Rojas, Leopoldo Pla~Sempere, Gema
  Ram{\'\i}rez-S{\'a}nchez, Elsa Sarr{\'\i}as, Marek Strelec, Brian Thompson,
  William Waites, Dion Wiggins, and Jaume Zaragoza. 2020.
\newblock \href {https://doi.org/10.18653/v1/2020.acl-main.417} {{P}ara{C}rawl:
  Web-scale acquisition of parallel corpora}.
\newblock In \emph{Proceedings of the 58th Annual Meeting of the Association
  for Computational Linguistics}.

\bibitem[{Ba{\~{n}}{\'{o}}n et~al.(2022)Ba{\~{n}}{\'{o}}n, Espl{\`{a}}{-}Gomis,
  Forcada, Garc{\'{\i}}a{-}Romero, Kuzman, Ljubesic, van Noord, Sempere,
  Ram{\'{\i}}rez{-}S{\'{a}}nchez, Rupnik, Suchomel, Toral, van~der Werff, and
  Zaragoza}]{macocu}
Marta Ba{\~{n}}{\'{o}}n, Miquel Espl{\`{a}}{-}Gomis, Mikel~L. Forcada, Cristian
  Garc{\'{\i}}a{-}Romero, Taja Kuzman, Nikola Ljubesic, Rik van Noord,
  Leopoldo~Pla Sempere, Gema Ram{\'{\i}}rez{-}S{\'{a}}nchez, Peter Rupnik,
  V{\'{\i}}t Suchomel, Antonio Toral, Tobias van~der Werff, and Jaume Zaragoza.
  2022.
\newblock \href {https://aclanthology.org/2022.eamt-1.41} {Macocu: Massive
  collection and curation of monolingual and bilingual data: focus on
  under-resourced languages}.
\newblock In \emph{Proceedings of the 23rd Annual Conference of the European
  Association for Machine Translation, {EAMT} 2022, Ghent, Belgium, June 1-3,
  2022}.

\bibitem[{Broder(1997)}]{broder1997resemblance}
Andrei~Z Broder. 1997.
\newblock On the resemblance and containment of documents.
\newblock In \emph{Proceedings. Compression and Complexity of SEQUENCES 1997
  (Cat. No. 97TB100171)}.

\bibitem[{Camacho-Collados et~al.(2016)Camacho-Collados, Delli~Bovi, Raganato,
  and Navigli}]{camacho-collados-etal-2016-large}
Jos{\'e} Camacho-Collados, Claudio Delli~Bovi, Alessandro Raganato, and Roberto
  Navigli. 2016.
\newblock \href {https://aclanthology.org/L16-1269} {A large-scale multilingual
  disambiguation of glosses}.
\newblock In \emph{Proceedings of the Tenth International Conference on
  Language Resources and Evaluation ({LREC}'16)}.

\bibitem[{Cassano et~al.(2022)Cassano, Gouwar, Nguyen, Nguyen, Phipps-Costin,
  Pinckney, Yee, Zi, Anderson, Feldman et~al.}]{Cassano2022MultiPLEAS}
Federico Cassano, John Gouwar, Daniel Nguyen, Sydney Nguyen, Luna
  Phipps-Costin, Donald Pinckney, Ming-Ho Yee, Yangtian Zi, Carolyn~Jane
  Anderson, Molly~Q Feldman, et~al. 2022.
\newblock Multipl-e: A scalable and extensible approach to benchmarking neural
  code generation.
\newblock \emph{arXiv preprint arXiv:2208.08227}.

\bibitem[{Chang et~al.(2023)Chang, Arnett, Tu, and
  Bergen}]{curse-of-multilinguality}
Tyler~A. Chang, Catherine Arnett, Zhuowen Tu, and Benjamin~K. Bergen. 2023.
\newblock \href {https://arxiv.org/abs/2311.09205} {When is multilinguality a
  curse? {L}anguage modeling for 250 high- and low-resource languages}.
\newblock \emph{arXiv preprint}.

\bibitem[{Chen et~al.(2021)Chen, Tworek, Jun, Yuan, Pinto, Kaplan, Edwards,
  Burda, Joseph, Brockman et~al.}]{chen2021codex}
Mark Chen, Jerry Tworek, Heewoo Jun, Qiming Yuan, Henrique Ponde De~Oliveira
  Pinto, Jared Kaplan, Harri Edwards, Yuri Burda, Nicholas Joseph, Greg
  Brockman, et~al. 2021.
\newblock Evaluating large language models trained on code.
\newblock \emph{arXiv preprint arXiv:2107.03374}.

\bibitem[{Chen et~al.(2024)Chen, Yu, Guo, and Haddow}]{chen2024good}
Pinzhen Chen, Simon Yu, Zhicheng Guo, and Barry Haddow. 2024.
\newblock \href {https://arxiv.org/abs/2406.12822} {Is it good data for
  multilingual instruction tuning or just bad multilingual evaluation for large
  language models?}
\newblock \emph{arXiv preprint arXiv:2406.12822}.

\bibitem[{Clark et~al.(2018)Clark, Cowhey, Etzioni, Khot, Sabharwal, Schoenick,
  and Tafjord}]{Clark2018ThinkYH}
Peter Clark, Isaac Cowhey, Oren Etzioni, Tushar Khot, Ashish Sabharwal, Carissa
  Schoenick, and Oyvind Tafjord. 2018.
\newblock Think you have solved question answering? try arc, the ai2 reasoning
  challenge.
\newblock \emph{ArXiv}.

\bibitem[{Cobbe et~al.(2021)Cobbe, Kosaraju, Bavarian, Hilton, Nakano, Hesse,
  and Schulman}]{cobbe2021training}
Karl Cobbe, Vineet Kosaraju, Mohammad Bavarian, Jacob Hilton, Reiichiro Nakano,
  Christopher Hesse, and John Schulman. 2021.
\newblock \href {https://arxiv.org/abs/2110.14168} {Training verifiers to solve
  math word problems}.
\newblock \emph{Preprint}, arXiv:2110.14168.

\bibitem[{Conneau et~al.(2020)Conneau, Khandelwal, Goyal, Chaudhary, Wenzek,
  Guzm{\'a}n, Grave, Ott, Zettlemoyer, and Stoyanov}]{conneau2020unsupervised}
Alexis Conneau, Kartikay Khandelwal, Naman Goyal, Vishrav Chaudhary, Guillaume
  Wenzek, Francisco Guzm{\'a}n, {\'E}douard Grave, Myle Ott, Luke Zettlemoyer,
  and Veselin Stoyanov. 2020.
\newblock Unsupervised cross-lingual representation learning at scale.
\newblock In \emph{Proceedings of the 58th Annual Meeting of the Association
  for Computational Linguistics}.

\bibitem[{Conneau et~al.(2018)Conneau, Rinott, Lample, Williams, Bowman,
  Schwenk, and Stoyanov}]{conneau2018xnli}
Alexis Conneau, Ruty Rinott, Guillaume Lample, Adina Williams, Samuel~R.
  Bowman, Holger Schwenk, and Veselin Stoyanov. 2018.
\newblock Xnli: Evaluating cross-lingual sentence representations.
\newblock In \emph{Proceedings of the 2018 Conference on Empirical Methods in
  Natural Language Processing}.

\bibitem[{{Corpora and Tools}(n.d.)}]{languages_of_russia}
{Corpora and Tools}. n.d.
\newblock \href {http://web-corpora.net/wsgi3/minorlangs/download} {Languages
  of {Russia}: Collections of texts in small languages}.

\bibitem[{Costa-juss{\`a} et~al.(2022)Costa-juss{\`a}, Cross, {\c{C}}elebi,
  Elbayad, Heafield, Heffernan, Kalbassi, Lam, Licht, Maillard
  et~al.}]{costa2022no}
Marta~R Costa-juss{\`a}, James Cross, Onur {\c{C}}elebi, Maha Elbayad, Kenneth
  Heafield, Kevin Heffernan, Elahe Kalbassi, Janice Lam, Daniel Licht, Jean
  Maillard, et~al. 2022.
\newblock No language left behind: Scaling human-centered machine translation.
\newblock \emph{arXiv preprint arXiv:2207.04672}.

\bibitem[{de~Gibert et~al.(2024)de~Gibert, Nail, Arefyev, Ba{\~n}{\'o}n,
  van~der Linde, Ji, Zaragoza-Bernabeu, Aulamo, Ram{\'\i}rez-S{\'a}nchez,
  Kutuzov et~al.}]{de2024new}
Ona de~Gibert, Graeme Nail, Nikolay Arefyev, Marta Ba{\~n}{\'o}n, Jelmer
  van~der Linde, Shaoxiong Ji, Jaume Zaragoza-Bernabeu, Mikko Aulamo, Gema
  Ram{\'\i}rez-S{\'a}nchez, Andrey Kutuzov, et~al. 2024.
\newblock A new massive multilingual dataset for high-performance language
  technologies.
\newblock In \emph{Proceedings of LREC-COLING}.

\bibitem[{Devlin et~al.(2019)Devlin, Chang, Lee, and
  Toutanova}]{devlin2019bert}
Jacob Devlin, Ming-Wei Chang, Kenton Lee, and Kristina Toutanova. 2019.
\newblock {BERT}: Pre-training of deep bidirectional transformers for language
  understanding.
\newblock In \emph{Proceedings of the 2019 Conference of the North American
  Chapter of the Association for Computational Linguistics: Human Language
  Technologies}.

\bibitem[{Dubey et~al.(2024)Dubey, Jauhri, Pandey, Kadian, Al-Dahle, Letman,
  Mathur, Schelten, Yang, Fan et~al.}]{dubey2024llama3}
Abhimanyu Dubey, Abhinav Jauhri, Abhinav Pandey, Abhishek Kadian, Ahmad
  Al-Dahle, Aiesha Letman, Akhil Mathur, Alan Schelten, Amy Yang, Angela Fan,
  et~al. 2024.
\newblock The llama 3 herd of models.
\newblock \emph{arXiv preprint arXiv:2407.21783}.

\bibitem[{Dunn(2020)}]{earthlings}
Jonathan Dunn. 2020.
\newblock \href {https://doi.org/10.1007/s10579-020-09489-2} {Mapping
  languages: the corpus of global language use}.
\newblock \emph{Lang. Resour. Evaluation}.

\bibitem[{EdTeKLA(2022)}]{indigenous-languages}
EdTeKLA. 2022.
\newblock \href {https://github.com/EdTeKLA/IndigenousLanguages_Corpora}
  {Indigenous languages corpora}.

\bibitem[{El-Haj(2020)}]{el-haj-2020-habibi}
Mahmoud El-Haj. 2020.
\newblock \href {https://aclanthology.org/2020.lrec-1.165} {Habibi - a multi
  dialect multi national {A}rabic song lyrics corpus}.
\newblock In \emph{Proceedings of the Twelfth Language Resources and Evaluation
  Conference}.

\bibitem[{El-Haj et~al.(2018)El-Haj, Rayson, and
  Aboelezz}]{el-haj-etal-2018-arabic}
Mahmoud El-Haj, Paul Rayson, and Mariam Aboelezz. 2018.
\newblock \href {https://aclanthology.org/L18-1573} {{A}rabic dialect
  identification in the context of bivalency and code-switching}.
\newblock In \emph{Proceedings of the Eleventh International Conference on
  Language Resources and Evaluation ({LREC} 2018)}.

\bibitem[{Faysse(2023)}]{project_gutenberg_HF_manu}
Manuel Faysse. 2023.
\newblock \href {https://huggingface.co/datasets/manu/project_gutenberg}
  {Dataset card for "project gutenberg"}.

\bibitem[{Gaim et~al.(2021)Gaim, Yang, and Park}]{Gaim2021TLMD}
Fitsum Gaim, Wonsuk Yang, and Jong~C. Park. 2021.
\newblock \href {https://doi.org/10.5281/zenodo.5139094} {Tlmd: Tigrinya
  language modeling dataset (1.0.0)}.
\newblock Dataset.

\bibitem[{Gao et~al.(2023)Gao, Tow, Biderman, Black, DiPofi, Foster, Golding,
  Hsu, McDonell, Muennighoff et~al.}]{eval-harness}
Leo Gao, Jonathan Tow, Stella Biderman, Sid Black, Anthony DiPofi, Charles
  Foster, Laurence Golding, Jeffrey Hsu, Kyle McDonell, Niklas Muennighoff,
  et~al. 2023.
\newblock A framework for few-shot language model evaluation.
\newblock Zenodo.

\bibitem[{Gilabert et~al.(2024)Gilabert, Escolano, Savall, Fornaciari, Mash,
  Liao, and Melero}]{gilabert2024investigatingtranslationcapabilitieslarge}
Javier~García Gilabert, Carlos Escolano, Aleix~Sant Savall, Francesca De~Luca
  Fornaciari, Audrey Mash, Xixian Liao, and Maite Melero. 2024.
\newblock \href {https://arxiv.org/abs/2406.09140} {Investigating the
  translation capabilities of large language models trained on parallel data
  only}.
\newblock \emph{Preprint}, arXiv:2406.09140.

\bibitem[{Goldhahn et~al.(2012)Goldhahn, Eckart, and
  Quasthoff}]{goldhahn-etal-2012-building}
Dirk Goldhahn, Thomas Eckart, and Uwe Quasthoff. 2012.
\newblock \href
  {http://www.lrec-conf.org/proceedings/lrec2012/pdf/327_Paper.pdf} {Building
  large monolingual dictionaries at the {L}eipzig corpora collection: From 100
  to 200 languages}.
\newblock In \emph{Proceedings of the Eighth International Conference on
  Language Resources and Evaluation ({LREC}'12)}.

\bibitem[{G{\'o}ngora et~al.(2021)G{\'o}ngora, Giossa, and
  Chiruzzo}]{gongora-etal-2021-experiments}
Santiago G{\'o}ngora, Nicol{\'a}s Giossa, and Luis Chiruzzo. 2021.
\newblock \href {https://doi.org/10.18653/v1/2021.americasnlp-1.16}
  {Experiments on a {G}uarani corpus of news and social media}.
\newblock In \emph{Proceedings of the First Workshop on Natural Language
  Processing for Indigenous Languages of the Americas}.

\bibitem[{G{\'o}ngora et~al.(2022)G{\'o}ngora, Giossa, and
  Chiruzzo}]{gongora-etal-2022-use}
Santiago G{\'o}ngora, Nicol{\'a}s Giossa, and Luis Chiruzzo. 2022.
\newblock \href {https://doi.org/10.18653/v1/2022.computel-1.16} {Can we use
  word embeddings for enhancing {G}uarani-{S}panish machine translation?}
\newblock In \emph{Proceedings of the Fifth Workshop on the Use of
  Computational Methods in the Study of Endangered Languages}.

\bibitem[{Gowda et~al.(2021)Gowda, Zhang, Mattmann, and
  May}]{gowda-etal-2021-many}
Thamme Gowda, Zhao Zhang, Chris Mattmann, and Jonathan May. 2021.
\newblock \href {https://doi.org/10.18653/v1/2021.acl-demo.37}
  {Many-to-{E}nglish machine translation tools, data, and pretrained models}.
\newblock In \emph{Proceedings of the 59th Annual Meeting of the Association
  for Computational Linguistics and the 11th International Joint Conference on
  Natural Language Processing: System Demonstrations}.

\bibitem[{{Grupo de Inteligencia Artificial
  PUCP}(n.d.)}]{multilingual-data-peru}
{Grupo de Inteligencia Artificial PUCP}. n.d.
\newblock \href {https://github.com/iapucp/multilingual-data-peru} {Monolingual
  and parallel corpora of peruvian languages}.

\bibitem[{Hasan et~al.(2021)Hasan, Bhattacharjee, Islam, Mubasshir, Li, Kang,
  Rahman, and Shahriyar}]{hasan-etal-2021-xl}
Tahmid Hasan, Abhik Bhattacharjee, Md.~Saiful Islam, Kazi Mubasshir, Yuan-Fang
  Li, Yong-Bin Kang, M.~Sohel Rahman, and Rifat Shahriyar. 2021.
\newblock \href {https://doi.org/10.18653/v1/2021.findings-acl.413} {{XL}-sum:
  Large-scale multilingual abstractive summarization for 44 languages}.
\newblock In \emph{Findings of the Association for Computational Linguistics:
  ACL-IJCNLP 2021}.

\bibitem[{Ibrahim et~al.(2024)Ibrahim, Thérien, Gupta, Richter, Anthony,
  Lesort, Belilovsky, and
  Rish}]{ibrahim2024simplescalablestrategiescontinually}
Adam Ibrahim, Benjamin Thérien, Kshitij Gupta, Mats~L. Richter, Quentin
  Anthony, Timothée Lesort, Eugene Belilovsky, and Irina Rish. 2024.
\newblock \href {https://arxiv.org/abs/2403.08763} {Simple and scalable
  strategies to continually pre-train large language models}.
\newblock \emph{Preprint}, arXiv:2403.08763.

\bibitem[{Ifeoluwa~Adelani et~al.(2021)Ifeoluwa~Adelani, Abbott, Neubig,
  D'souza, Kreutzer, Lignos, Palen-Michel, Buzaaba, Rijhwani, Ruder
  et~al.}]{masakhane_tacl_2021}
David Ifeoluwa~Adelani, Jade Abbott, Graham Neubig, Daniel D'souza, Julia
  Kreutzer, Constantine Lignos, Chester Palen-Michel, Happy Buzaaba, Shruti
  Rijhwani, Sebastian Ruder, et~al. 2021.
\newblock Masakhaner: Named entity recognition for african languages.
\newblock \emph{arXiv e-prints}.

\bibitem[{Imani et~al.(2023)Imani, Lin, Kargaran, Severini, Sabet, Kassner, Ma,
  Schmid, Martins, Yvon, and Sch{\"u}tze}]{imanigooghari2023glot500}
Ayyoob Imani, Peiqin Lin, Amir~Hossein Kargaran, Silvia Severini, Masoud~Jalili
  Sabet, Nora Kassner, Chunlan Ma, Helmut Schmid, Andr{\'e} F.~T. Martins,
  Fran{\c c}ois Yvon, and Hinrich Sch{\"u}tze. 2023.
\newblock \href {https://aclanthology.org/2023.acl-long.61/} {Glot500: Scaling
  multilingual corpora and language models to 500 languages}.
\newblock \emph{arXiv preprint}.

\bibitem[{Joulin et~al.(2016{\natexlab{a}})Joulin, Grave, Bojanowski, Douze,
  J{\'e}gou, and Mikolov}]{joulin2016fasttext}
Armand Joulin, Edouard Grave, Piotr Bojanowski, Matthijs Douze, H{\'e}rve
  J{\'e}gou, and Tomas Mikolov. 2016{\natexlab{a}}.
\newblock Fasttext.zip: Compressing text classification models.
\newblock \emph{arXiv preprint arXiv:1612.03651}.

\bibitem[{Joulin et~al.(2016{\natexlab{b}})Joulin, Grave, Bojanowski, and
  Mikolov}]{joulin2016bag}
Armand Joulin, Edouard Grave, Piotr Bojanowski, and Tomas Mikolov.
  2016{\natexlab{b}}.
\newblock Bag of tricks for efficient text classification.
\newblock \emph{arXiv preprint arXiv:1607.01759}.

\bibitem[{Kakwani et~al.(2020)Kakwani, Kunchukuttan, Golla, N.C.,
  Bhattacharyya, Khapra, and Kumar}]{kakwani-etal-2020-indicnlpsuite}
Divyanshu Kakwani, Anoop Kunchukuttan, Satish Golla, Gokul N.C., Avik
  Bhattacharyya, Mitesh~M. Khapra, and Pratyush Kumar. 2020.
\newblock \href {https://doi.org/10.18653/v1/2020.findings-emnlp.445}
  {{I}ndic{NLPS}uite: Monolingual corpora, evaluation benchmarks and
  pre-trained multilingual language models for {I}ndian languages}.
\newblock In \emph{Findings of the Association for Computational Linguistics:
  EMNLP 2020}.

\bibitem[{Kargaran et~al.(2023)Kargaran, Imani, Yvon, and
  Sch{\"u}tze}]{kargaran2023glotlid}
Amir~Hossein Kargaran, Ayyoob Imani, Fran{\c{c}}ois Yvon, and Hinrich
  Sch{\"u}tze. 2023.
\newblock \href {https://openreview.net/forum?id=dl4e3EBz5j} {{GlotLID}:
  Language identification for low-resource languages}.
\newblock In \emph{The 2023 Conference on Empirical Methods in Natural Language
  Processing}.

\bibitem[{Kargaran et~al.(2024)Kargaran, Yvon, and
  Sch{\"u}tze}]{kargaran2024glotscript}
Amir~Hossein Kargaran, Fran{\c{c}}ois Yvon, and Hinrich Sch{\"u}tze. 2024.
\newblock Glotscript: A resource and tool for low resource writing system
  identification.
\newblock In \emph{Proceedings of the 2024 Joint International Conference on
  Computational Linguistics, Language Resources and Evaluation (LREC-COLING
  2024)}.

\bibitem[{Kim et~al.(2024)Kim, Schuster, and Toshniwal}]{kim2024entity}
Najoung Kim, Sebastian Schuster, and Shubham Toshniwal. 2024.
\newblock \href {https://doi.org/10.48550/ARXIV.2405.21068} {Code pretraining
  improves entity tracking abilities of language models}.
\newblock \emph{CoRR}.

\bibitem[{Kingma and Ba(2015)}]{kingma2015adam}
Diederik~P. Kingma and Jimmy Ba. 2015.
\newblock Adam: A method for stochastic optimization.
\newblock In \emph{International Conference for Learning Representations}.

\bibitem[{Kocetkov et~al.(2023)Kocetkov, Li, Allal, Li, Mou, Jernite, Mitchell,
  Ferrandis, Hughes, Wolf, Bahdanau, von Werra, and
  de~Vries}]{kocetkov2023stack}
Denis Kocetkov, Raymond Li, Loubna~Ben Allal, Jia Li, Chenghao Mou, Yacine
  Jernite, Margaret Mitchell, Carlos~Mu{\~{n}}oz Ferrandis, Sean Hughes, Thomas
  Wolf, Dzmitry Bahdanau, Leandro von Werra, and Harm de~Vries. 2023.
\newblock \href {https://openreview.net/forum?id=pxpbTdUEpD} {The stack: 3 {TB}
  of permissively licensed source code}.
\newblock \emph{Trans. Mach. Learn. Res.}

\bibitem[{Kondo et~al.(2024)Kondo, Utsuro, and
  Nagata}]{kondo2024enhancingtranslationaccuracylarge}
Minato Kondo, Takehito Utsuro, and Masaaki Nagata. 2024.
\newblock \href {https://arxiv.org/abs/2407.03145} {Enhancing translation
  accuracy of large language models through continual pre-training on parallel
  data}.
\newblock \emph{Preprint}, arXiv:2407.03145.

\bibitem[{Koto and Koto(2020)}]{koto-koto-2020-towards}
Fajri Koto and Ikhwan Koto. 2020.
\newblock \href {https://aclanthology.org/2020.paclic-1.17} {Towards
  computational linguistics in {M}inangkabau language: Studies on sentiment
  analysis and machine translation}.
\newblock In \emph{Proceedings of the 34th Pacific Asia Conference on Language,
  Information and Computation}.

\bibitem[{Kudugunta et~al.(2024)Kudugunta, Caswell, Zhang, Garcia, Xin,
  Kusupati, Stella, Bapna, and Firat}]{kudugunta2024madlad}
Sneha Kudugunta, Isaac Caswell, Biao Zhang, Xavier Garcia, Derrick Xin, Aditya
  Kusupati, Romi Stella, Ankur Bapna, and Orhan Firat. 2024.
\newblock Madlad-400: A multilingual and document-level large audited dataset.
\newblock \emph{Advances in Neural Information Processing Systems}.

\bibitem[{Kunchukuttan et~al.(2018)Kunchukuttan, Mehta, and
  Bhattacharyya}]{kunchukuttan-etal-2018-iit}
Anoop Kunchukuttan, Pratik Mehta, and Pushpak Bhattacharyya. 2018.
\newblock \href {https://aclanthology.org/L18-1548} {The {IIT} {B}ombay
  {E}nglish-{H}indi parallel corpus}.
\newblock In \emph{Proceedings of the Eleventh International Conference on
  Language Resources and Evaluation ({LREC} 2018)}.

\bibitem[{Kwon et~al.(2023)Kwon, Li, Zhuang, Sheng, Zheng, Yu, Gonzalez, Zhang,
  and Stoica}]{kwon2023efficient}
Woosuk Kwon, Zhuohan Li, Siyuan Zhuang, Ying Sheng, Lianmin Zheng, Cody~Hao Yu,
  Joseph~E. Gonzalez, Hao Zhang, and Ion Stoica. 2023.
\newblock Efficient memory management for large language model serving with
  pagedattention.
\newblock In \emph{Proceedings of the ACM SIGOPS 29th Symposium on Operating
  Systems Principles}.

\bibitem[{Lai et~al.(2023)Lai, Nguyen, Ngo, Nguyen, Dernoncourt, Rossi, and
  Nguyen}]{lai2023okapi}
Viet~Dac Lai, Chien~Van Nguyen, Nghia~Trung Ngo, Thuat Nguyen, Franck
  Dernoncourt, Ryan~A. Rossi, and Thien~Huu Nguyen. 2023.
\newblock Okapi: Instruction-tuned large language models in multiple languages
  with reinforcement learning from human feedback.
\newblock In \emph{Proceedings of the 2023 Conference on Empirical Methods in
  Natural Language Processing: System Demonstrations}.

\bibitem[{Lai et~al.(2024)Lai, Mesgar, and Fraser}]{lai-etal-2024-llms}
Wen Lai, Mohsen Mesgar, and Alexander Fraser. 2024.
\newblock \href {https://aclanthology.org/2024.findings-acl.488} {{LLM}s beyond
  {E}nglish: Scaling the multilingual capability of {LLM}s with cross-lingual
  feedback}.
\newblock In \emph{Findings of the Association for Computational Linguistics
  ACL 2024}.

\bibitem[{Lauren{\c{c}}on et~al.(2022)Lauren{\c{c}}on, Saulnier, Wang, Akiki,
  del Moral, Le~Scao, Von~Werra, Mou, Ponferrada, Nguyen
  et~al.}]{laurenccon2022bigscience}
Hugo Lauren{\c{c}}on, Lucile Saulnier, Thomas Wang, Christopher Akiki,
  Albert~Villanova del Moral, Teven Le~Scao, Leandro Von~Werra, Chenghao Mou,
  Eduardo~Gonz{\'a}lez Ponferrada, Huu Nguyen, et~al. 2022.
\newblock The bigscience roots corpus: A 1.6 tb composite multilingual dataset.
\newblock In \emph{Thirty-sixth Conference on Neural Information Processing
  Systems Datasets and Benchmarks Track}.

\bibitem[{Leong et~al.(2022)Leong, Nemecek, Mansdorfer, Filighera, Owodunni,
  and Whitenack}]{bloom-lm}
Colin Leong, Joshua Nemecek, Jacob Mansdorfer, Anna Filighera, Abraham
  Owodunni, and Daniel Whitenack. 2022.
\newblock \href {https://aclanthology.org/2022.emnlp-main.590} {Bloom library:
  Multimodal datasets in 300+ languages for a variety of downstream tasks}.
\newblock In \emph{Proceedings of the 2022 Conference on Empirical Methods in
  Natural Language Processing, {EMNLP} 2022, Abu Dhabi, United Arab Emirates,
  December 7-11, 2022}.

\bibitem[{Levesque et~al.(2012)Levesque, Davis, and
  Morgenstern}]{levesque2012winograd}
Hector Levesque, Ernest Davis, and Leora Morgenstern. 2012.
\newblock The {Winograd} schema challenge.
\newblock In \emph{Thirteenth international conference on the principles of
  knowledge representation and reasoning}.

\bibitem[{Li et~al.(2023)Li, Liu, Bian, Fang, Huang, Liu, Wang, and
  You}]{li2023colossal}
Shenggui Li, Hongxin Liu, Zhengda Bian, Jiarui Fang, Haichen Huang, Yuliang
  Liu, Boxiang Wang, and Yang You. 2023.
\newblock Colossal-ai: A unified deep learning system for large-scale parallel
  training.
\newblock In \emph{Proceedings of the 52nd International Conference on Parallel
  Processing}.

\bibitem[{Li et~al.(2022)Li, Zhang, Zhao, Shen, Liu, Mao, and
  Zhang}]{li2022csl}
Yudong Li, Yuqing Zhang, Zhe Zhao, Linlin Shen, Weijie Liu, Weiquan Mao, and
  Hui Zhang. 2022.
\newblock {CSL}: A large-scale chinese scientific literature dataset.
\newblock In \emph{Proceedings of the 29th International Conference on
  Computational Linguistics}.

\bibitem[{Life(2014)}]{evenki-newspaper}
Evenki Life. 2014.
\newblock \href
  {https://drive.google.com/file/d/1he2q6RncA_NKHPIJjSzlkK-2qgEFTiCG/view}
  {Evenki life newspaper}.

\bibitem[{Lin(2004)}]{lin-2004-rouge}
Chin-Yew Lin. 2004.
\newblock \href {https://aclanthology.org/W04-1013} {{ROUGE}: A package for
  automatic evaluation of summaries}.
\newblock In \emph{Text Summarization Branches Out}.

\bibitem[{Lin et~al.(2024)Lin, Ji, Tiedemann, Martins, and
  Sch{\"u}tze}]{lin2024mala}
Peiqin Lin, Shaoxiong Ji, J{\"o}rg Tiedemann, Andr{\'e}~FT Martins, and Hinrich
  Sch{\"u}tze. 2024.
\newblock {MaLA}-500: Massive language adaptation of large language models.
\newblock \emph{arXiv preprint arXiv:2401.13303}.

\bibitem[{Lin et~al.(2022)Lin, Mihaylov, Artetxe, Wang, Chen, Simig, Ott,
  Goyal, Bhosale, Du et~al.}]{lin2022few}
Xi~Victoria Lin, Todor Mihaylov, Mikel Artetxe, Tianlu Wang, Shuohui Chen,
  Daniel Simig, Myle Ott, Naman Goyal, Shruti Bhosale, Jingfei Du, et~al. 2022.
\newblock Few-shot learning with multilingual generative language models.
\newblock In \emph{Proceedings of the 2022 Conference on Empirical Methods in
  Natural Language Processing}.

\bibitem[{Lo et~al.(2020)Lo, Wang, Neumann, Kinney, and
  Weld}]{lo-wang-2020-s2orc}
Kyle Lo, Lucy~Lu Wang, Mark Neumann, Rodney Kinney, and Daniel Weld. 2020.
\newblock \href {https://doi.org/10.18653/v1/2020.acl-main.447} {{S}2{ORC}: The
  semantic scholar open research corpus}.
\newblock In \emph{Proceedings of the 58th Annual Meeting of the Association
  for Computational Linguistics}.

\bibitem[{Lozhkov et~al.(2024)Lozhkov, Li, Allal, Cassano, Lamy-Poirier, Tazi,
  Tang, Pykhtar, Liu, Wei et~al.}]{lozhkov2024starcoder}
Anton Lozhkov, Raymond Li, Loubna~Ben Allal, Federico Cassano, Joel
  Lamy-Poirier, Nouamane Tazi, Ao~Tang, Dmytro Pykhtar, Jiawei Liu, Yuxiang
  Wei, et~al. 2024.
\newblock Starcoder 2 and the stack v2: The next generation.
\newblock \emph{arXiv preprint arXiv:2402.19173}.

\bibitem[{Lu et~al.(2024)Lu, Zhu, Li, Qiao, and Yuan}]{lu2024llamax}
Yinquan Lu, Wenhao Zhu, Lei Li, Yu~Qiao, and Fei Yuan. 2024.
\newblock {LLaMAX}: Scaling linguistic horizons of {LLM} by enhancing
  translation capabilities beyond 100 languages.
\newblock \emph{arXiv preprint arXiv:2407.05975}.

\bibitem[{Luo et~al.(2023)Luo, Yang, Meng, Li, Zhou, and
  Zhang}]{luo2023catastrophic}
Yun Luo, Zhen Yang, Fandong Meng, Yafu Li, Jie Zhou, and Yue Zhang. 2023.
\newblock \href {https://doi.org/10.48550/ARXIV.2308.08747} {An empirical study
  of catastrophic forgetting in large language models during continual
  fine-tuning}.
\newblock \emph{CoRR}.

\bibitem[{Ma et~al.(2023)Ma, ImaniGooghari, Ye, Asgari, and
  Sch{\"u}tze}]{ma2023taxi1500}
Chunlan Ma, Ayyoob ImaniGooghari, Haotian Ye, Ehsaneddin Asgari, and Hinrich
  Sch{\"u}tze. 2023.
\newblock \href {https://arxiv.org/abs/2305.08487} {Taxi1500: A multilingual
  dataset for text classification in 1500 languages}.
\newblock \emph{Preprint}, arXiv:2305.08487.

\bibitem[{Ma et~al.(2024)Ma, Liu, Yu, Zhang, Jiang, Wang, and Li}]{ma2024stage}
Yingwei Ma, Yue Liu, Yue Yu, Yuanliang Zhang, Yu~Jiang, Changjian Wang, and
  Shanshan Li. 2024.
\newblock \href {https://openreview.net/forum?id=KIPJKST4gw} {At which training
  stage does code data help llms reasoning?}
\newblock In \emph{The Twelfth International Conference on Learning
  Representations, {ICLR} 2024, Vienna, Austria, May 7-11, 2024}.

\bibitem[{Majli{\v s}(2011)}]{w2c}
Martin Majli{\v s}. 2011.
\newblock \href {http://hdl.handle.net/11858/00-097C-0000-0022-6133-9} {{W2C}
  – web to corpus – corpora}.
\newblock {LINDAT}/{CLARIAH}-{CZ} digital library at the Institute of Formal
  and Applied Linguistics ({{\'U}FAL}), Faculty of Mathematics and Physics,
  Charles University.

\bibitem[{Masakhane(2023)}]{lacuna-project}
Masakhane. 2023.
\newblock \href {https://github.com/masakhane-io/lacuna_pos_ner} {Lacuna
  project}.

\bibitem[{Mayer and Cysouw(2014)}]{DBLP:conf/lrec/MayerC14}
Thomas Mayer and Michael Cysouw. 2014.
\newblock \href
  {http://www.lrec-conf.org/proceedings/lrec2014/summaries/220.html} {Creating
  a massively parallel bible corpus}.
\newblock In \emph{Proceedings of the Ninth International Conference on
  Language Resources and Evaluation, {LREC} 2014, Reykjavik, Iceland, May
  26-31, 2014}.

\bibitem[{Mirzakhalov et~al.(2021)Mirzakhalov, Babu, Ataman, Kariev, Tyers,
  Abduraufov, Hajili, Ivanova, Khaytbaev, Laverghetta~Jr., Moydinboyev, Onal,
  Pulatova, Wahab, Firat, and Chellappan}]{mirzakhalov-etal-2021-large}
Jamshidbek Mirzakhalov, Anoop Babu, Duygu Ataman, Sherzod Kariev, Francis
  Tyers, Otabek Abduraufov, Mammad Hajili, Sardana Ivanova, Abror Khaytbaev,
  Antonio Laverghetta~Jr., Bekhzodbek Moydinboyev, Esra Onal, Shaxnoza
  Pulatova, Ahsan Wahab, Orhan Firat, and Sriram Chellappan. 2021.
\newblock \href {https://doi.org/10.18653/v1/2021.emnlp-main.475} {A
  large-scale study of machine translation in {T}urkic languages}.
\newblock In \emph{Proceedings of the 2021 Conference on Empirical Methods in
  Natural Language Processing}.

\bibitem[{Moran et~al.(2022)Moran, Bentz, Gutierrez-Vasques, Pelloni, and
  Samardzic}]{moran-etal-2022-teddi}
Steven Moran, Christian Bentz, Ximena Gutierrez-Vasques, Olga Pelloni, and
  Tanja Samardzic. 2022.
\newblock \href {https://aclanthology.org/2022.lrec-1.123} {{T}e{DD}i sample:
  Text data diversity sample for language comparison and multilingual {NLP}}.
\newblock In \emph{Proceedings of the Thirteenth Language Resources and
  Evaluation Conference}.

\bibitem[{Morishita et~al.(2020)Morishita, Suzuki, and
  Nagata}]{morishita-etal-2020-jparacrawl}
Makoto Morishita, Jun Suzuki, and Masaaki Nagata. 2020.
\newblock \href {https://aclanthology.org/2020.lrec-1.443} {{JP}ara{C}rawl: A
  large scale web-based {E}nglish-{J}apanese parallel corpus}.
\newblock In \emph{Proceedings of the Twelfth Language Resources and Evaluation
  Conference}.

\bibitem[{Mostafazadeh et~al.(2017)Mostafazadeh, Roth, Louis, Chambers, and
  Allen}]{mostafazadeh2017lsdsem}
Nasrin Mostafazadeh, Michael Roth, Annie Louis, Nathanael Chambers, and James
  Allen. 2017.
\newblock Lsdsem 2017 shared task: The story cloze test.
\newblock In \emph{Proceedings of the 2nd Workshop on Linking Models of
  Lexical, Sentential and Discourse-level Semantics}.

\bibitem[{Mou et~al.(2023)Mou, Ha, Enevoldsen, and
  Liu}]{chenghao_mou_2023_8364980}
Chenghao Mou, Chris Ha, Kenneth Enevoldsen, and Peiyuan Liu. 2023.
\newblock \href {https://doi.org/10.5281/zenodo.8364980}
  {Chenghaomou/text-dedup: Reference snapshot}.

\bibitem[{Muennighoff et~al.(2022)Muennighoff, Wang, Sutawika, Roberts,
  Biderman, Scao, Bari, Shen, Yong, Schoelkopf
  et~al.}]{muennighoff2022crosslingual}
Niklas Muennighoff, Thomas Wang, Lintang Sutawika, Adam Roberts, Stella
  Biderman, Teven~Le Scao, M~Saiful Bari, Sheng Shen, Zheng-Xin Yong, Hailey
  Schoelkopf, et~al. 2022.
\newblock Crosslingual generalization through multitask finetuning.
\newblock \emph{arXiv preprint arXiv:2211.01786}.

\bibitem[{Mukiibi et~al.(2022)Mukiibi, Katumba, Nakatumba-Nabende, Hussein, and
  Meyer}]{mukiibi2022makerere}
Jonathan Mukiibi, Andrew Katumba, Joyce Nakatumba-Nabende, Ali Hussein, and
  Joshua Meyer. 2022.
\newblock The makerere radio speech corpus: A luganda radio corpus for
  automatic speech recognition.
\newblock In \emph{Proceedings of the Thirteenth Language Resources and
  Evaluation Conference}.

\bibitem[{Nakamura et~al.(2024)Nakamura, Mishra, Tedeschi, Chai, Stillerman,
  Friedrich, Yadav, Laud, Chien, Zhuo et~al.}]{nakamura2024aurora}
Taishi Nakamura, Mayank Mishra, Simone Tedeschi, Yekun Chai, Jason~T
  Stillerman, Felix Friedrich, Prateek Yadav, Tanmay Laud, Vu~Minh Chien,
  Terry~Yue Zhuo, et~al. 2024.
\newblock Aurora-m: The first open source multilingual language model
  red-teamed according to the us executive order.
\newblock \emph{arXiv preprint arXiv:2404.00399}.

\bibitem[{Nakazawa et~al.(2022)Nakazawa, Mino, Goto, Dabre, Higashiyama,
  Parida, Kunchukuttan, Morishita, Bojar, Chu, Eriguchi, Abe, Oda, and
  Kurohashi}]{nakazawa-etal-2022-overview}
Toshiaki Nakazawa, Hideya Mino, Isao Goto, Raj Dabre, Shohei Higashiyama,
  Shantipriya Parida, Anoop Kunchukuttan, Makoto Morishita, Ond{\v{r}}ej Bojar,
  Chenhui Chu, Akiko Eriguchi, Kaori Abe, Yusuke Oda, and Sadao Kurohashi.
  2022.
\newblock \href {https://aclanthology.org/2022.wat-1.1} {Overview of the 9th
  workshop on {A}sian translation}.
\newblock In \emph{Proceedings of the 9th Workshop on Asian Translation}.

\bibitem[{Nakazawa et~al.(2021)Nakazawa, Nakayama, Ding, Dabre, Higashiyama,
  Mino, Goto, Pa~Pa, Kunchukuttan, Parida, Bojar, Chu, Eriguchi, Abe, Oda, and
  Kurohashi}]{nakazawa-etal-2021-overview}
Toshiaki Nakazawa, Hideki Nakayama, Chenchen Ding, Raj Dabre, Shohei
  Higashiyama, Hideya Mino, Isao Goto, Win Pa~Pa, Anoop Kunchukuttan,
  Shantipriya Parida, Ond{\v{r}}ej Bojar, Chenhui Chu, Akiko Eriguchi, Kaori
  Abe, Yusuke Oda, and Sadao Kurohashi. 2021.
\newblock \href {https://doi.org/10.18653/v1/2021.wat-1.1} {Overview of the 8th
  workshop on {A}sian translation}.
\newblock In \emph{Proceedings of the 8th Workshop on Asian Translation
  (WAT2021)}.

\bibitem[{Neubig(2011)}]{neubig11kftt}
Graham Neubig. 2011.
\newblock The {Kyoto} free translation task.
\newblock http://www.phontron.com/kftt.

\bibitem[{Nguyen et~al.(2023)Nguyen, Nguyen, Lai, Man, Ngo, Dernoncourt, Rossi,
  and Nguyen}]{nguyen2023culturax}
Thuat Nguyen, Chien~Van Nguyen, Viet~Dac Lai, Hieu Man, Nghia~Trung Ngo, Franck
  Dernoncourt, Ryan~A. Rossi, and Thien~Huu Nguyen. 2023.
\newblock \href {https://arxiv.org/abs/2309.09400} {Culturax: A cleaned,
  enormous, and multilingual dataset for large language models in 167
  languages}.
\newblock \emph{Preprint}, arXiv:2309.09400.

\bibitem[{Ogueji et~al.(2021)Ogueji, Zhu, and Lin}]{afriberta}
Kelechi Ogueji, Yuxin Zhu, and Jimmy Lin. 2021.
\newblock \href {https://aclanthology.org/2021.mrl-1.11} {Small data? no
  problem! exploring the viability of pretrained multilingual language models
  for low-resourced languages}.
\newblock In \emph{Proceedings of the 1st Workshop on Multilingual
  Representation Learning}.

\bibitem[{OSCAR(2023)}]{OSCAR2301}
OSCAR. 2023.
\newblock \href {https://huggingface.co/datasets/oscar-corpus/OSCAR-2301}
  {Oscar (open super-large crawled aggregated corpus) 2301}.

\bibitem[{Palen-Michel et~al.(2022)Palen-Michel, Kim, and
  Lignos}]{palen-michel-etal-2022-multilingual}
Chester Palen-Michel, June Kim, and Constantine Lignos. 2022.
\newblock \href {https://aclanthology.org/2022.lrec-1.224} {Multilingual open
  text release 1: Public domain news in 44 languages}.
\newblock In \emph{Proceedings of the Thirteenth Language Resources and
  Evaluation Conference}.

\bibitem[{Papineni et~al.(2002)Papineni, Roukos, Ward, and
  Zhu}]{papineni-etal-2002-bleu}
Kishore Papineni, Salim Roukos, Todd Ward, and Wei-Jing Zhu. 2002.
\newblock \href {https://doi.org/10.3115/1073083.1073135} {{B}leu: a method for
  automatic evaluation of machine translation}.
\newblock In \emph{Proceedings of the 40th Annual Meeting of the Association
  for Computational Linguistics}.

\bibitem[{Paul et~al.(2024)Paul, Glavas, and Gurevych}]{paul2024ircoder}
Indraneil Paul, Goran Glavas, and Iryna Gurevych. 2024.
\newblock \href {https://aclanthology.org/2024.acl-long.802} {Ircoder:
  Intermediate representations make language models robust multilingual code
  generators}.
\newblock In \emph{Proceedings of the 62nd Annual Meeting of the Association
  for Computational Linguistics (Volume 1: Long Papers), {ACL} 2024, Bangkok,
  Thailand, August 11-16, 2024}.

\bibitem[{Ponti et~al.(2020)Ponti, Glava{\v{s}}, Majewska, Liu, Vuli{\'c}, and
  Korhonen}]{ponti-etal-2020-xcopa}
Edoardo~Maria Ponti, Goran Glava{\v{s}}, Olga Majewska, Qianchu Liu, Ivan
  Vuli{\'c}, and Anna Korhonen. 2020.
\newblock {XCOPA}: A multilingual dataset for causal commonsense reasoning.
\newblock In \emph{Proceedings of the 2020 Conference on Empirical Methods in
  Natural Language Processing (EMNLP)}.

\bibitem[{Popovi{\'c}(2015)}]{popovic-2015-chrf}
Maja Popovi{\'c}. 2015.
\newblock \href {https://doi.org/10.18653/v1/W15-3049} {chr{F}: character
  n-gram {F}-score for automatic {MT} evaluation}.
\newblock In \emph{Proceedings of the Tenth Workshop on Statistical Machine
  Translation}.

\bibitem[{Post(2018)}]{post-2018-call}
Matt Post. 2018.
\newblock \href {https://www.aclweb.org/anthology/W18-6319} {A call for clarity
  in reporting {BLEU} scores}.
\newblock In \emph{Proceedings of the Third Conference on Machine Translation:
  Research Papers}.

\bibitem[{Raffel et~al.(2020)Raffel, Shazeer, Roberts, Lee, Narang, Matena,
  Zhou, Li, and Liu}]{raffel2020exploring_t5}
Colin Raffel, Noam Shazeer, Adam Roberts, Katherine Lee, Sharan Narang, Michael
  Matena, Yanqi Zhou, Wei Li, and Peter~J Liu. 2020.
\newblock Exploring the limits of transfer learning with a unified text-to-text
  transformer.
\newblock \emph{Journal of machine learning research}.

\bibitem[{Rajaraman and Ullman(2011)}]{rajaraman2011mining}
Anand Rajaraman and Jeffrey~D Ullman. 2011.
\newblock \emph{Mining of massive datasets}.
\newblock Autoedicion.

\bibitem[{Rivest(1992)}]{rivest1992md5}
Ronald Rivest. 1992.
\newblock The md5 message-digest algorithm.

\bibitem[{Roziere et~al.(2023)Roziere, Gehring, Gloeckle, Sootla, Gat, Tan,
  Adi, Liu, Remez, Rapin et~al.}]{roziere2023code}
Baptiste Roziere, Jonas Gehring, Fabian Gloeckle, Sten Sootla, Itai Gat,
  Xiaoqing~Ellen Tan, Yossi Adi, Jingyu Liu, Tal Remez, J{\'e}r{\'e}my Rapin,
  et~al. 2023.
\newblock Code llama: Open foundation models for code.
\newblock \emph{arXiv preprint arXiv:2308.12950}.

\bibitem[{Rozis and Skadi{\c{n}}{\v{s}}(2017)}]{rozis-skadins-2017-tilde}
Roberts Rozis and Raivis Skadi{\c{n}}{\v{s}}. 2017.
\newblock \href {https://aclanthology.org/W17-0235} {Tilde {MODEL} -
  multilingual open data for {EU} languages}.
\newblock In \emph{Proceedings of the 21st Nordic Conference on Computational
  Linguistics}.

\bibitem[{Sajjad et~al.(2020)Sajjad, Abdelali, Durrani, and
  Dalvi}]{sajjad-etal-2020-arabench}
Hassan Sajjad, Ahmed Abdelali, Nadir Durrani, and Fahim Dalvi. 2020.
\newblock \href {https://doi.org/10.18653/v1/2020.coling-main.447}
  {{A}ra{B}ench: Benchmarking dialectal {A}rabic-{E}nglish machine
  translation}.
\newblock In \emph{Proceedings of the 28th International Conference on
  Computational Linguistics}.

\bibitem[{Scao et~al.(2022)Scao, Fan, Akiki, Pavlick, Ili{\'c}, Hesslow,
  Castagn{\'e}, Luccioni, Yvon, Gall{\'e} et~al.}]{scao2022bloom}
Teven~Le Scao, Angela Fan, Christopher Akiki, Ellie Pavlick, Suzana Ili{\'c},
  Daniel Hesslow, Roman Castagn{\'e}, Alexandra~Sasha Luccioni, Fran{\c{c}}ois
  Yvon, Matthias Gall{\'e}, et~al. 2022.
\newblock {BLOOM}: A {176B}-parameter open-access multilingual language model.
\newblock \emph{arXiv preprint}.

\bibitem[{Schwenk et~al.(2021{\natexlab{a}})Schwenk, Chaudhary, Sun, Gong, and
  Guzm{\'a}n}]{schwenk-etal-2021-wikimatrix}
Holger Schwenk, Vishrav Chaudhary, Shuo Sun, Hongyu Gong, and Francisco
  Guzm{\'a}n. 2021{\natexlab{a}}.
\newblock \href {https://doi.org/10.18653/v1/2021.eacl-main.115}
  {{W}iki{M}atrix: Mining 135{M} parallel sentences in 1620 language pairs from
  {W}ikipedia}.
\newblock In \emph{Proceedings of the 16th Conference of the European Chapter
  of the Association for Computational Linguistics: Main Volume}.

\bibitem[{Schwenk et~al.(2021{\natexlab{b}})Schwenk, Wenzek, Edunov, Grave,
  Joulin, and Fan}]{schwenk-etal-2021-ccmatrix}
Holger Schwenk, Guillaume Wenzek, Sergey Edunov, Edouard Grave, Armand Joulin,
  and Angela Fan. 2021{\natexlab{b}}.
\newblock \href {https://doi.org/10.18653/v1/2021.acl-long.507} {{CCM}atrix:
  Mining billions of high-quality parallel sentences on the web}.
\newblock In \emph{Proceedings of the 59th Annual Meeting of the Association
  for Computational Linguistics and the 11th International Joint Conference on
  Natural Language Processing (Volume 1: Long Papers)}.

\bibitem[{Shi et~al.(2022)Shi, Suzgun, Freitag, Wang, Srivats, Vosoughi, Chung,
  Tay, Ruder, Zhou, Das, and Wei}]{shi2022language}
Freda Shi, Mirac Suzgun, Markus Freitag, Xuezhi Wang, Suraj Srivats, Soroush
  Vosoughi, Hyung~Won Chung, Yi~Tay, Sebastian Ruder, Denny Zhou, Dipanjan Das,
  and Jason Wei. 2022.
\newblock \href {https://arxiv.org/abs/2210.03057} {Language models are
  multilingual chain-of-thought reasoners}.
\newblock \emph{Preprint}, arXiv:2210.03057.

\bibitem[{Shliazhko et~al.(2022)Shliazhko, Fenogenova, Tikhonova, Mikhailov,
  Kozlova, and Shavrina}]{shliazhko2022mgpt}
Oleh Shliazhko, Alena Fenogenova, Maria Tikhonova, Vladislav Mikhailov,
  Anastasia Kozlova, and Tatiana Shavrina. 2022.
\newblock {mGPT}: Few-shot learners go multilingual.
\newblock \emph{arXiv preprint arXiv:2204.07580}.

\bibitem[{Singh(2008)}]{singh-2008-named}
Anil~Kumar Singh. 2008.
\newblock \href {https://aclanthology.org/I08-5003} {Named entity recognition
  for south and south {E}ast {A}sian languages: Taking stock}.
\newblock In \emph{Proceedings of the {IJCNLP}-08 Workshop on Named Entity
  Recognition for South and South East {A}sian Languages}.

\bibitem[{Singh et~al.(2024)Singh, Vargus, Dsouza, Karlsson, Mahendiran, Ko,
  Shandilya, Patel, Mataciunas, OMahony et~al.}]{singh-etal-2024-aya}
Shivalika Singh, Freddie Vargus, Daniel Dsouza, B{\"o}rje~F Karlsson, Abinaya
  Mahendiran, Wei-Yin Ko, Herumb Shandilya, Jay Patel, Deividas Mataciunas,
  Laura OMahony, et~al. 2024.
\newblock \href {https://aclanthology.org/2024.acl-long.620} {Aya dataset: An
  open-access collection for multilingual instruction tuning}.
\newblock In \emph{Proceedings of the 62nd Annual Meeting of the Association
  for Computational Linguistics (Volume 1: Long Papers)}.

\bibitem[{Soldaini and Lo(2023)}]{peS2o}
Luca Soldaini and Kyle Lo. 2023.
\newblock {peS2o (Pretraining Efficiently on S2ORC) Dataset}.
\newblock Technical report, {Allen Institute for AI}.
\newblock ODC-By, \url{https://github.com/allenai/pes2o}.

\bibitem[{Su{\'a}rez et~al.(2019)Su{\'a}rez, Sagot, and
  Romary}]{suarez2019asynchronous}
Pedro Javier~Ortiz Su{\'a}rez, Beno{\^\i}t Sagot, and Laurent Romary. 2019.
\newblock Asynchronous pipeline for processing huge corpora on medium to low
  resource infrastructures.
\newblock In \emph{7th Workshop on the Challenges in the Management of Large
  Corpora (CMLC-7)}.

\bibitem[{Szafraniec et~al.(2023)Szafraniec, Rozi{\`{e}}re, Leather, Labatut,
  Charton, and Synnaeve}]{szafraniec2022translation}
Marc Szafraniec, Baptiste Rozi{\`{e}}re, Hugh Leather, Patrick Labatut,
  Fran{\c{c}}ois Charton, and Gabriel Synnaeve. 2023.
\newblock \href {https://openreview.net/forum?id=XomEU3eNeSQ} {Code translation
  with compiler representations}.
\newblock In \emph{The Eleventh International Conference on Learning
  Representations, {ICLR} 2023, Kigali, Rwanda, May 1-5, 2023}.

\bibitem[{Taylor et~al.(2022)Taylor, Kardas, Cucurull, Scialom, Hartshorn,
  Saravia, Poulton, Kerkez, and Stojnic}]{taylor2022galactica}
Ross Taylor, Marcin Kardas, Guillem Cucurull, Thomas Scialom, Anthony
  Hartshorn, Elvis Saravia, Andrew Poulton, Viktor Kerkez, and Robert Stojnic.
  2022.
\newblock Galactica: A large language model for science.
\newblock \emph{arXiv preprint arXiv:2211.09085}.

\bibitem[{Team et~al.(2024)Team, Riviere, Pathak, Sessa, Hardin, Bhupatiraju,
  Hussenot, Mesnard, Shahriari, Ram{\'e}
  et~al.}]{gemmateam2024gemma2improvingopen}
Gemma Team, Morgane Riviere, Shreya Pathak, Pier~Giuseppe Sessa, Cassidy
  Hardin, Surya Bhupatiraju, L{\'e}onard Hussenot, Thomas Mesnard, Bobak
  Shahriari, Alexandre Ram{\'e}, et~al. 2024.
\newblock Gemma 2: Improving open language models at a practical size.
\newblock \emph{arXiv preprint arXiv:2408.00118}.

\bibitem[{Tejaswi et~al.(2024)Tejaswi, Gupta, and
  Choi}]{tejaswi2024exploringdesignchoicesbuilding}
Atula Tejaswi, Nilesh Gupta, and Eunsol Choi. 2024.
\newblock \href {https://arxiv.org/abs/2406.14670} {Exploring design choices
  for building language-specific llms}.
\newblock \emph{Preprint}, arXiv:2406.14670.

\bibitem[{Thuat~Nguyen and Nguyen(2024)}]{nguyen2024culturay}
Huu~Nguyen Thuat~Nguyen and Thien Nguyen. 2024.
\newblock Culturay: A large cleaned multilingual dataset of 75 languages.

\bibitem[{Tiedemann(2012)}]{opus_2012}
J{\"o}rg Tiedemann. 2012.
\newblock \href
  {http://www.lrec-conf.org/proceedings/lrec2012/pdf/463_Paper.pdf} {Parallel
  data, tools and interfaces in {OPUS}}.
\newblock In \emph{Proceedings of the Eighth International Conference on
  Language Resources and Evaluation ({LREC}'12)}.

\bibitem[{Tiedemann(2020)}]{tatoeba}
J{\"o}rg Tiedemann. 2020.
\newblock \href {https://www.aclweb.org/anthology/2020.wmt-1.139} {The
  {T}atoeba {T}ranslation {C}hallenge {--} {R}ealistic data sets for low
  resource and multilingual {MT}}.
\newblock In \emph{Proceedings of the Fifth Conference on Machine Translation}.

\bibitem[{Tikhonov and Ryabinin(2021)}]{tikhonov-ryabinin-2021-heads}
Alexey Tikhonov and Max Ryabinin. 2021.
\newblock {I}t{'}s {A}ll in the {H}eads: {U}sing {A}ttention {H}eads as a
  {B}aseline for {C}ross-{L}ingual {T}ransfer in {C}ommonsense {R}easoning.
\newblock In \emph{Findings of the Association for Computational Linguistics:
  ACL-IJCNLP 2021}.

\bibitem[{Touvron et~al.(2023)Touvron, Martin, Stone, Albert, Almahairi,
  Babaei, Bashlykov, Batra, Bhargava, Bhosale et~al.}]{touvron2023llama2}
Hugo Touvron, Louis Martin, Kevin Stone, Peter Albert, Amjad Almahairi, Yasmine
  Babaei, Nikolay Bashlykov, Soumya Batra, Prajjwal Bhargava, Shruti Bhosale,
  et~al. 2023.
\newblock Llama 2: Open foundation and fine-tuned chat models.
\newblock \emph{arXiv preprint}.

\bibitem[{{\"U}st{\"u}n et~al.(2024){\"U}st{\"u}n, Aryabumi, Yong, Ko, D'souza,
  Onilude, Bhandari, Singh, Ooi, Kayid et~al.}]{ustun2024aya}
Ahmet {\"U}st{\"u}n, Viraat Aryabumi, Zheng-Xin Yong, Wei-Yin Ko, Daniel
  D'souza, Gbemileke Onilude, Neel Bhandari, Shivalika Singh, Hui-Lee Ooi, Amr
  Kayid, et~al. 2024.
\newblock Aya model: An instruction finetuned open-access multilingual language
  model.
\newblock \emph{arXiv preprint arXiv:2402.07827}.

\bibitem[{Vaswani et~al.(2017)Vaswani, Shazeer, Parmar, Uszkoreit, Jones,
  Gomez, Kaiser, and Polosukhin}]{vaswani2017attention}
Ashish Vaswani, Noam Shazeer, Niki Parmar, Jakob Uszkoreit, Llion Jones,
  Aidan~N Gomez, {\L}ukasz Kaiser, and Illia Polosukhin. 2017.
\newblock Attention is all you need.
\newblock In \emph{Advances in neural information processing systems}.

\bibitem[{Wang et~al.(2023)Wang, Tu, Chen, Yuan, Huang, Jiao, and
  Lyu}]{wang2023all}
Wenxuan Wang, Zhaopeng Tu, Chang Chen, Youliang Yuan, Jen-tse Huang, Wenxiang
  Jiao, and Michael~R Lyu. 2023.
\newblock All languages matter: On the multilingual safety of large language
  models.
\newblock \emph{arXiv preprint arXiv:2310.00905}.

\bibitem[{Wenzek et~al.(2020)Wenzek, Lachaux, Conneau, Chaudhary, Guzm{\'a}n,
  Joulin, and Grave}]{cc100_2}
Guillaume Wenzek, Marie-Anne Lachaux, Alexis Conneau, Vishrav Chaudhary,
  Francisco Guzm{\'a}n, Armand Joulin, and Edouard Grave. 2020.
\newblock \href {https://aclanthology.org/2020.lrec-1.494} {{CCN}et: Extracting
  high quality monolingual datasets from web crawl data}.
\newblock In \emph{Proceedings of the Twelfth Language Resources and Evaluation
  Conference}.

\bibitem[{{Wikimedia Foundation}(n.d.)}]{wikidump}
{Wikimedia Foundation}. n.d.
\newblock \href {https://dumps.wikimedia.org} {Wikimedia downloads}.

\bibitem[{Wilie et~al.(2020)Wilie, Vincentio, Winata, Cahyawijaya, Li, Lim,
  Soleman, Mahendra, Fung, Bahar, and Purwarianti}]{wilie2020indonlu}
Bryan Wilie, Karissa Vincentio, Genta~Indra Winata, Samuel Cahyawijaya, X.~Li,
  Zhi~Yuan Lim, S.~Soleman, R.~Mahendra, Pascale Fung, Syafri Bahar, and
  A.~Purwarianti. 2020.
\newblock Indonlu: Benchmark and resources for evaluating indonesian natural
  language understanding.
\newblock In \emph{Proceedings of the 1st Conference of the Asia-Pacific
  Chapter of the Association for Computational Linguistics and the 10th
  International Joint Conference on Natural Language Processing}.

\bibitem[{Wolf et~al.(2019)Wolf, Debut, Sanh, Chaumond, Delangue, Moi, Cistac,
  Rault, Louf, Funtowicz et~al.}]{wolf2019huggingface}
Thomas Wolf, Lysandre Debut, Victor Sanh, Julien Chaumond, Clement Delangue,
  Anthony Moi, Pierric Cistac, Tim Rault, R{\'e}mi Louf, Morgan Funtowicz,
  et~al. 2019.
\newblock Huggingface's transformers: State-of-the-art natural language
  processing.
\newblock \emph{arXiv preprint}.

\bibitem[{Xue et~al.(2021)Xue, Constant, Roberts, Kale, Al-Rfou, Siddhant,
  Barua, and Raffel}]{xue2021mt5}
Linting Xue, Noah Constant, Adam Roberts, Mihir Kale, Rami Al-Rfou, Aditya
  Siddhant, Aditya Barua, and Colin Raffel. 2021.
\newblock {mT5}: A massively multilingual pre-trained text-to-text transformer.
\newblock In \emph{Proceedings of the 2021 Conference of the North American
  Chapter of the Association for Computational Linguistics: Human Language
  Technologies}.

\bibitem[{Yang et~al.(2024{\natexlab{a}})Yang, Yang, Hui, Zheng, Yu, Zhou, Li,
  Li, Liu, Huang et~al.}]{yang2024qwen2technicalreport}
An~Yang, Baosong Yang, Binyuan Hui, Bo~Zheng, Bowen Yu, Chang Zhou, Chengpeng
  Li, Chengyuan Li, Dayiheng Liu, Fei Huang, et~al. 2024{\natexlab{a}}.
\newblock Qwen2 technical report.
\newblock \emph{arXiv preprint arXiv:2407.10671}.

\bibitem[{Yang et~al.(2024{\natexlab{b}})Yang, Liu, Wu, Yang, Fung, Li, Huang,
  Cao, Wang, Wang, Ji, and Zhai}]{yang2024wand}
Ke~Yang, Jiateng Liu, John Wu, Chaoqi Yang, Yi~R. Fung, Sha Li, Zixuan Huang,
  Xu~Cao, Xingyao Wang, Yiquan Wang, Heng Ji, and Chengxiang Zhai.
  2024{\natexlab{b}}.
\newblock \href {https://doi.org/10.48550/ARXIV.2401.00812} {If {LLM} is the
  wizard, then code is the wand: {A} survey on how code empowers large language
  models to serve as intelligent agents}.
\newblock \emph{CoRR}.

\bibitem[{Zevallos et~al.(2022)Zevallos, Ortega, Chen, Castro, Bel, Toshio,
  Venturas, Aradiel, and Melgarejo}]{zevallos-etal-2022-introducing}
Rodolfo Zevallos, John Ortega, William Chen, Richard Castro, N{\'u}ria Bel,
  Cesar Toshio, Renzo Venturas, Hilario Aradiel, and Nelsi Melgarejo. 2022.
\newblock \href {https://doi.org/10.18653/v1/2022.deeplo-1.1} {Introducing
  {Q}u{BERT}: A large monolingual corpus and {BERT} model for {S}outhern
  {Q}uechua}.
\newblock In \emph{Proceedings of the Third Workshop on Deep Learning for
  Low-Resource Natural Language Processing}.

\bibitem[{Zhang et~al.(2024{\natexlab{a}})Zhang, Tao, Huang, Lin, Chen, and
  Feng}]{zhang2024mc}
Chen Zhang, Mingxu Tao, Quzhe Huang, Jiuheng Lin, Zhibin Chen, and Yansong
  Feng. 2024{\natexlab{a}}.
\newblock Mc$^2$: Towards transparent and culturally-aware nlp for minority
  languages in china.
\newblock \emph{arXiv preprint arXiv:2311.08348}.

\bibitem[{Zhang et~al.(2019)Zhang, Kishore, Wu, Weinberger, and
  Artzi}]{zhang2019bertscore}
Tianyi Zhang, Varsha Kishore, Felix Wu, Kilian~Q Weinberger, and Yoav Artzi.
  2019.
\newblock Bertscore: Evaluating text generation with bert.
\newblock \emph{arXiv preprint arXiv:1904.09675}.

\bibitem[{Zhang et~al.(2024{\natexlab{b}})Zhang, Chen, Ye, Yang, Chen, Wang,
  and Petzold}]{zhang2024unveiling}
Xinlu Zhang, Zhiyu~Zoey Chen, Xi~Ye, Xianjun Yang, Lichang Chen, William~Yang
  Wang, and Linda~Ruth Petzold. 2024{\natexlab{b}}.
\newblock \href {https://doi.org/10.48550/ARXIV.2405.20535} {Unveiling the
  impact of coding data instruction fine-tuning on large language models
  reasoning}.
\newblock \emph{CoRR}.

\bibitem[{Zhu et~al.(2018)Zhu, Lu, Zheng, Guo, Zhang, Wang, and
  Yu}]{zhu2018texygen}
Yaoming Zhu, Sidi Lu, Lei Zheng, Jiaxian Guo, Weinan Zhang, Jun Wang, and Yong
  Yu. 2018.
\newblock Texygen: A benchmarking platform for text generation models.
\newblock In \emph{The 41st international ACM SIGIR conference on research \&
  development in information retrieval}.

\end{thebibliography}

\pagebreak

\begin{appendices}
\section{Questions and Answers}

\subsection{Novelty}
\noindent \textbf{Q:} What is the main goal of this paper?\\
\noindent \textbf{A:} The main goal is to improve multilingual language models by creating a massive, diverse multilingual corpus (MaLA) and applying continual pre-training to enhance linguistic inclusivity and task performance. \\

\noindent \textbf{Q:} What are the main achievements and the scientific novelty?\\
\noindent \textbf{A:} Our main achievements and scientific novelty are written in the introduction (\Cref{sec:introduction}).
To summarize the key contributions of this paper: the MaLA corpus's scale, diversity, and focus on low-resource languages represent a significant advancement over previous efforts. 
By integrating a variety of text types---such as code, scientific papers, and instructions---this work provides a more comprehensive foundation for multilingual pre-training. 
These improvements lead to a better multilingual LLM (EMMA-500) that boosts multilingual capabilities evaluated on 9 tasks and 15 benchmarks. 
The scale and impact of this work make it a valuable contribution to the field of multilingual NLP, especially for low-resource languages. \\

\subsection{Data}

\noindent \textbf{Q:} How was the data mix tuned?  \\
\noindent \textbf{A:} The tuning of the data mix is a separate research topic. Our paper focuses on providing a valuable resource through data processing and model training. While we did not conduct specific ablations on the data mix, we carefully curated the corpus to ensure diversity and balance across languages and text types. Further exploration of this topic could be beneficial for future work. \\

\noindent \textbf{Q:} Since whitespace cleaning only applies to whitespace-tokenized languages, are the other languages cleaner per se or need cleaning in other ways?  \\
\noindent \textbf{A:} The whitespace cleaning process primarily applies to whitespace-tokenized languages, helping standardize the text for better processing. For languages that do not rely on whitespace tokenization (such as Chinese, Japanese, or Korean), the text may not require this specific cleaning but might need other preprocessing steps, such as character normalization or segmentation. \\

\noindent \textbf{Q:} How were existing datasets and their languages to be included chosen? \\
\noindent \textbf{A:} The selection of existing datasets and their languages for inclusion was based on several factors:
\begin{itemize}
    \item Language Coverage: We prioritized datasets that cover a wide range of languages, with a focus on including both high-resource and low-resource languages to ensure broad multilingual representation.
    \item Data Quality: Datasets with high-quality, diverse content, such as books, scientific papers, and code, were selected to ensure the corpus has varied text types, improving model generalization.
    \item Task Relevance: We considered datasets that aligned well with our goals for continual pre-training and multilingual adaptation across tasks like translation, classification, and reasoning.
    \item Availability: Datasets that were publicly available and accessible were preferred, ensuring transparency and reproducibility.
\end{itemize}

\subsection{Evaluation}

\noindent \textbf{Q:}  How do we choose the number of shots in in-context learning?\\
\noindent \textbf{A:} 
The choice of the number of shots in evaluation is based on the results of the Llama 2 model, which achieves reasonably good scores on high-resource languages with this setup. We do not grid-search the number of shots. We release the model weights. The readers can choose the number of shots as wish to evaluate. \\

\noindent \textbf{Q:} How would the model perform with fewer or more shots? Is the number of shots optimized for EMMA-500?  \\
\noindent \textbf{A:} The choice of 3-shot in machine translation was based on results from the LLaMA 2 model, where it showed reasonable performance on high-resource languages. The number of shots was not optimized through a grid search for EMMA-500. While fewer or more shots could impact performance, further experimentation would be needed to determine the optimal number for different tasks and languages. However, this is not the primary focus of our work, which is intended as a resource paper. The 3-shot choice already effectively showcases the model's performance across tasks, providing a good balance between data efficiency and task completion.\\

\noindent \textbf{Q:} Why no human evaluation?\\
\noindent \textbf{A:} Human evaluation, while valuable, has limitations such as subjectivity, inconsistency, and scalability. Evaluators may have biases, leading to variability in judgments across individuals and cultures. Additionally, evaluating multilingual and open-ended tasks requires diverse, linguistically skilled annotators, which can be expensive and time-consuming. These challenges make automated metrics a practical complement, even though they may lack nuanced judgment. While resource constraints have limited our ability to perform human evaluation for this work, we recognize its importance in complementing automatic metrics. However, we must emphasize the limitations and unsustainability of human evaluation, particularly in a massively multilingual setting. Even evaluating a subset of languages or datasets, most likely low-resource languages, remains prohibitively expensive. Also, finding qualified evaluators for low-resource languages is often impractical and costly. Automatic tools, while not perfect, offer a scalable and consistent method for data collection and evaluation. Determining whether automatic or human evaluation is better is not the primary focus of this paper.\\

\noindent \textbf{Q:} How reliable is machine-generated benchmark?\\
\noindent \textbf{A:} Machine-generated benchmarks have significant value as they offer scalability, consistency, and objectivity, making it feasible to evaluate models across numerous tasks and languages efficiently. They are particularly useful for comparative analysis and identifying measurable improvements. 
There are also limitations: they often miss nuanced qualities like creativity, contextual appropriateness, and cultural sensitivity, especially in open-ended or multilingual tasks. Biases in the generation process or evaluation metrics can also lead to misleading results. \\

\noindent \textbf{Q:} Why no safety-related evaluation?\\
\noindent \textbf{A:} We find safety-related evaluation a separate research topic. This work looks at CPT and releases a corpus and a multilingual base model rather than an instruction-tuned model. While our work focuses on multilingual data expansion and model evaluation across a broad range of languages and tasks, safety evaluation was not a primary focus as the model is not directly developed for end users. We recognize the importance of safety benchmarks like XSafety \citep{wang2023all} and Aya redteaming \citep{aakanksha2024aya_redteaming} that only cover a few languages, and they might not fit the evaluation in a massively multilingual setting best. \\

\subsection{Comparison with Baselines}
\noindent \textbf{Q:} Is it unfair to compare models that do not officially support certain languages?\\
\noindent \textbf{A:} Our evaluation focuses on the performance of models across a broad range of languages, including those that may not be officially supported. This approach highlights the model's ability to generalize and handle underrepresented languages, even if they are not specifically optimized for them. There is no widely accepted standard for which models should be used as the ``fair'' baseline for such comparisons. \\

\noindent \textbf{Q:} What is the relation to the MaLA-500 model/work? What are the key improvements over it?\\
\noindent \textbf{A:} The key improvements include:
\begin{itemize}
    \item Larger, More Diverse Corpus: our corpus includes 939 languages and 74 billion tokens.
    \item Higher Token Count per Document: our corpus features longer documents (average of 90 tokens per sequence) compared to Glot500-c used by the MaLA-500 model (average of 19 tokens), providing better context and enabling the model to capture long-range dependencies.
    \item More Diverse Data: We also include a wide range of text types, including code, books, and scientific papers.
    \item Better Performance: Evaluated across more benchmarks, EMMA-500 shows improvements in multilingual tasks like intrinsic tasks, translation, commonsense reasoning, and text classification.
\end{itemize}

\noindent \textbf{Q:} Why is Tower instruct performing so poorly (especially since it is built for translation)? \\
\noindent \textbf{A:}  According to its technical report, Tower is heavily optimized (through CPT and SFT) for only around 10 languages centered on translation-related data. Its language capabilities might not generalize well across low-resource languages or specific language pairs. This is essentially our model's advantage over Tower.\\

\subsection{Others}

\noindent \textbf{Q:} Will this corpus, model weight, benchmark and processing scripts be public? \\
\noindent \textbf{A:}  Yes, we release the MaLA corpus (including raw, cleaned, and deduplicated versions and splits for training and validation), model weights, codes, and model generation. Benchmarks, processing scripts, and training software are open-sourced by the original authors. Some benchmarks, such as PBC, will require permission to access data from the original authors. \\

\noindent \textbf{Q:} Why is the global batch size set to 16M, rather than the commonly used 4M tokens like LLaMA-2?  \\
\noindent \textbf{A:}  This is a technical choice that allows us to efficiently use our computing. We note that increasing batch size during training is now seen in LLM pre-training, e.g., Llama-3 doubled their batch size twice from 4M to 16M during training. \\

\noindent \textbf{Q:} How do we categorize language into different resource groups?\label{q:language_categorization} \\ 
\noindent \textbf{A:} We classify languages into five groups---high, medium-high, medium, medium-low, and low---based on their token counts in the MaLA corpus.
\Cref{tab:languages} shows the specific languages in different categories. 
Since the MaLA corpus is compiled from multiple corpora, its token distribution reflects the availability of textual data across languages. Grouping languages based on token counts provides a practical and data-driven way to distinguish between high- and low-resource languages, aligning with real-world data availability rather than relying on predefined taxonomies. Besides, We provide standard ISO language codes along with their corresponding writing systems, which ensures clarity, consistency, and interoperability across different linguistic resources and NLP applications.\\

\noindent \textbf{Q:} Why don't we use Joshi et al.'s language taxonomy\footnote{\url{https://microsoft.github.io/linguisticdiversity/assets/lang2tax.txt}}? \\
\noindent \textbf{A:} While we acknowledge the importance of language taxonomies like Joshi et al.'s, we choose not to rely on it for this work for several reasons:
\begin{itemize}
    \item Focus on Practical Impact: Our work emphasizes advancing multilingual NLP through comprehensive corpus creation and improved model performance, rather than adhering to a specific taxonomy. We focused on covering as many languages as possible, particularly underrepresented ones, using ISO codes for consistency in data processing.
    \item Feasibility vs. Contribution: Manually aligning our language set with Joshi's taxonomy, especially given the absence of ISO codes, is a labor-intensive task with limited relevance to the core contributions of this paper. While it is feasible, we believe our efforts are better directed toward demonstrating the utility of the MaLA corpus and models through tangible multilingual improvements.
\end{itemize}
Our MaLA corpus's classification system utilizes ISO codes to offer a standardized way to reference languages, reducing ambiguity, especially for languages with multiple names or dialects, and includes writing systems that further enhance precision, particularly for languages that use multiple scripts. 
This approach facilitates better data organization, model training, and cross-referencing with external linguistic datasets.
Besides, we report per-language performance for some benchmarks in the Appendix, allowing readers to categorize the performance into different groups of languages as they wish. To avoid making the Appendix unnecessarily long, we will also share all results, including models' generation, via GitHub, enabling the community to conduct further analyses with their preferred language taxonomy.
\\

\noindent \textbf{Q:} Why do we not add which languages they officially support and separate the aggregated metrics for officially supported and unseen languages? \\
\noindent \textbf{A:} We can separate metrics for models that explicitly mention supported languages, but this is not possible for models like LLaMA and GEMMA 2, which do not provide such details. We report per-language performance for some benchmarks in the Appendix, allowing readers to categorize the performance into different groups of languages as they wish. To avoid making the Appendix unnecessarily long, we will also share all results, including models' generation, via GitHub, enabling the community to conduct further analyses with their preferred metrics. \\

\section{Data Sources}

\subsection{Monolingual Data}

\subsubsection{List of Data Sources} 
The MaLA corpus harvests a wide range of datasets in multiple domains. 
\Cref{table:monolingual_table} in the appendix lists the corpora and collections we used as monolingual data sources.
The major ones include AfriBERTa \citep{afriberta}, Bloom library \citep{bloom-lm}, CC100 \citep{conneau2020unsupervised,cc100_2}
CulturaX \citep{nguyen2023culturax}, CulturaY \citep{nguyen2024culturay}, the Curse of Multilinguality \citep{curse-of-multilinguality}, Evenki Life~\citep{evenki-newspaper}, Glot500 \citep{imanigooghari2023glot500}, GlotSparse \citep{kargaran2023glotlid}, monoHPLT of HPLT v1.2 \citep{de2024new} from the HPLT project~\citep{aulamo2023hplt}, Indigenous Languages Corpora \citep{indigenous-languages}, Indo4B \citep{wilie2020indonlu}, 
Lacuna Project \citep{lacuna-project}, Languages of Russia \citep{languages_of_russia}, MADLAD-400 \citep{kudugunta2024madlad}, Makerere Radio Speech Corpus \citep{mukiibi2022makerere}, masakhane-ner1.0 \citep{masakhane_tacl_2021}, MC2 \citep{zhang2024mc}, mC4 \citep{raffel2020exploring_t5}, multilingual-data-peru \citep{multilingual-data-peru}, OSCAR 2301 \citep{OSCAR2301} from the OSCAR project \footnote{\url{https://oscar-project.org/}}, Tatoeba challenge monolingual collection \citep{tatoeba}, The Leipzig Corpora \citep{goldhahn-etal-2012-building}, Tigrinya Language Modeling \citep{Gaim2021TLMD}, Wikipedia 20231101 \citep{wikidump} and Wikisource 20231201 \citep{wikidump}.
We exclude high-resource languages in CulturaX, HPLT, MADLAD-400, CC100, mC4, and OSCAR 2301.
We exclude Gahuza and Pidgin in the AfriBERTa dataset.
We filter out texts that mainly contain a date or timestamp in the Languages of Russia dataset.
For Glot500-c, we filter out texts, which may come from train or test tests from datasets for machine translation, such as Flores200, Tatoeba, and mtdata. 
Despite the translation data being split into source and target languages in the Glot500-c corpus, we decide to filter them to avoid potential data leakage, especially since we use Flores200 as the evaluation benchmark.

\Cref{table:monolingual_table} lists the corpora and collections we use as monolingual data sources in this work. Monolingual data sources simply contain text data in a single language.

\paragraph{Metadata} For the monolingual data sources, we define the contents of the \texttt{JSONL} output file from the pre-processing workflow to consist of the fields \texttt{url}, \texttt{text}, \texttt{collection}, \texttt{source}, and \texttt{original\_code}.
The contents of these fields are as follows. The field \texttt{text} contains the language data in a granularity specific to the given corpus. If the granularity was sentence-level, then we could expect the sentences in the corpus to generally be independent of each other, while only parts of the sentences---such as phrases, clauses, and words---exhibit serial dependence. If the granularity was paragraph-level, then we could expect the sentences within paragraphs to have serial dependence, while paragraphs to largely be independent of each other. 
The field \texttt{url} contains a URL indicating the web address from which the text data has been extracted, if available. The field \texttt{collection} contains the name of the collection, i.e., a corpus or a set of corpora, which the text is extracted from, whereas the field  \texttt{source} contains the name of a more specific part of the collection, such as the name of an individual corpus or a file in the collection the text was extracted from. Lastly, the field \texttt{original\_code} contains the language code of the text data as it is designated in the data source, e.g., in the directory structure, the filenames, or the data object returned by an API call.

\begin{table*}[ht!]
\scriptsize
\centering
\caption{Datasets used as monolingual source data.\label{table:monolingual_table}}
\setlength{\tabcolsep}{3pt}
\resizebox{\linewidth}{!}{
\begin{tabular}{p{7cm}p{1.2cm}p{3cm}p{10cm}r}
\toprule
Name & Languages & Domains & URL & Year \\
\midrule
AfriBERTa~\citep{afriberta}	&	10	&	news	&	\url{https://huggingface.co/datasets/castorini/afriberta-corpus}	&	2021	\\
Bloom library~\citep{bloom-lm}	&	363	&	religious, books	&	\url{https://huggingface.co/datasets/sil-ai/bloom-lm}	&	2022	\\
CC100~\citep{conneau2020unsupervised}	&	100	&	web	&	\url{https://huggingface.co/datasets/cc100}	&	2020	\\
CulturaX~\citep{nguyen2023culturax}	&	167	&	web	&	\url{https://huggingface.co/datasets/uonlp/CulturaX}	&	2023	\\
CulturaY~\citep{nguyen2024culturay}	&	75	&	web	&	\url{https://huggingface.co/datasets/ontocord/CulturaY}	&	2024	\\
curse-of-multilinguality~\citep{curse-of-multilinguality}	&	200	&	misc	&	\url{https://github.com/tylerachang/curse-of-multilinguality}	&	2023	\\
Evenki Life~\citep{evenki-newspaper}	&	1	&	newspapers	&	\url{https://drive.google.com/file/d/1he2q6RncA_NKHPIJjSzlkK-2qgEFTiCG/view}	&	2014	\\
Glot500~\citep{imanigooghari2023glot500}	&	511	&	misc	&	\url{https://huggingface.co/datasets/cis-lmu/Glot500}	&	2023	\\
GlotSparse~\citep{kargaran2023glotlid}	&	10	&	news	&	\url{https://huggingface.co/datasets/cis-lmu/GlotSparse}	&	2023	\\
HPLT v1.2~\citep{de2024new}	&	75	&	web	&	\url{https://hplt-project.org/datasets/v1.2}	&	2024	\\
Indigenous Languages Corpora~\citep{indigenous-languages}	&	1	&	UNK	&	\url{https://github.com/EdTeKLA/IndigenousLanguages_Corpora}	&	2022	\\
Indo4B~\citep{wilie2020indonlu}	&	1	&	misc	&	\url{https://github.com/IndoNLP/indonlu?tab=readme-ov-file}	&	2020	\\
Lacuna Project~\citep{lacuna-project}	&	20	&	UNK	&	\url{https://github.com/masakhane-io/lacuna_pos_ner}	&	2023	\\
Languages of Russia~\citep{languages_of_russia}	&	46	&	social media, web	&	\url{http://web-corpora.net/wsgi3/minorlangs/download}	&	UNK	\\
Madlad-400~\citep{kudugunta2024madlad}	&	419	&	web	&	\url{https://huggingface.co/datasets/allenai/MADLAD-400}	&	2023	\\
Makerere Radio Speech Corpus~\citep{mukiibi2022makerere}	&	1	&	transcription	&	\url{https://zenodo.org/records/5855017}	&	2022	\\
masakhane-ner1.0~\citep{masakhane_tacl_2021}	&	12	&	UNK	&	\url{https://github.com/masakhane-io/masakhane-ner}	&	2021	\\
MC2~\citep{zhang2024mc}	&	4	&	web	&	\url{https://huggingface.co/datasets/pkupie/mc2_corpus}	&	2024	\\
mC4~\citep{raffel2020exploring_t5}	&	101	&	web	&	\url{https://huggingface.co/datasets/allenai/c4}	&	2020	\\
multilingual-data-peru~\citep{multilingual-data-peru}	&	4	&	UNK	&	\url{https://github.com/iapucp/multilingual-data-peru}	&	2020	\\
OSCAR 2301~\citep{OSCAR2301}	&	152	&	web	&	\url{https://huggingface.co/datasets/oscar-corpus/OSCAR-2301}	&	2023	\\
The Leipzig Corpora~\citep{goldhahn-etal-2012-building}	&	136	&	newspapers, wikipedia	&	\url{https://huggingface.co/datasets/imvladikon/leipzig_corpora_collection}	&	2012	\\
Tigrinya Language Modeling~\citep{Gaim2021TLMD}	&	1	&	news, blogs, books	&	\url{https://zenodo.org/records/5139094}	&	2021	\\
Wikipedia 20231101~\citep{wikidump}	&	323	&	wikipedia	&	\url{https://huggingface.co/datasets/wikimedia/wikipedia}	&	2023	\\
Wikisource 20231201~\citep{wikidump}	&	73	&	books	&	\url{https://huggingface.co/datasets/wikimedia/wikisource}	&	2023	\\
Tatoeba challenge monolingual~\citep{tatoeba}	&	280	&	wikimedia	&	\url{https://github.com/Helsinki-NLP/Tatoeba-Challenge/blob/master/data/MonolingualData-v2020-07-28.md}	&	2020	\\
\bottomrule
\end{tabular}
}
\end{table*}

Glot500-c uses the following datasets:
AI4Bharat,\footnote{\url{https://ai4bharat.org/}}
AIFORTHAI-LotusCorpus,\footnote{\url{https://github.com/korakot/corpus/releases/download/v1.0/AIFORTHAI-LotusCorpus.zip}}
Add \citep{el-haj-etal-2018-arabic},
AfriBERTa \citep{afriberta},
AfroMAFT \citep{adelani-etal-2022-thousand,xue2021mt5},
Anuvaad,\footnote{\url{https://github.com/project-anuvaad/anuvaad-parallel-corpus}}
AraBench \citep{sajjad-etal-2020-arabench},
AUTSHUMATO,\footnote{\url{https://autshumato.sourceforge.net/}}
Bloom \citep{bloom-lm}, 
CC100 \citep{conneau2020unsupervised}, 
CCNet \citep{cc100_2}, 
CMU\_Haitian\_Creole,\footnote{\url{http://www.speech.cs.cmu.edu/haitian/text/}}
CORP.NCHLT,\footnote{\url{https://repo.sadilar.org/handle/20.500.12185/7}}
Clarin,\footnote{\url{https://www.clarin.si/}}
DART \citep{alsarsour-etal-2018-dart},
Earthlings \citep{earthlings}, 
FFR,\footnote{\url{https://github.com/bonaventuredossou/ffr-v1/tree/master/FFR-Dataset}}
Flores200 \citep{costa2022no}, 
GiossaMedia \citep{gongora-etal-2022-use, gongora-etal-2021-experiments},
Glosses \citep{camacho-collados-etal-2016-large},
Habibi \citep{el-haj-2020-habibi},
HinDialect \citep{bafna2022empirical}, 
HornMT,\footnote{\url{https://github.com/asmelashteka/HornMT}}
IITB \citep{kunchukuttan-etal-2018-iit},
IndicNLP \citep{nakazawa-etal-2021-overview},
Indiccorp \citep{kakwani-etal-2020-indicnlpsuite}, 
isiZulu,\footnote{\url{https://zenodo.org/record/5035171}}
JParaCrawl \citep{morishita-etal-2020-jparacrawl}, 
KinyaSMT,\footnote{\url{https://github.com/pniyongabo/kinyarwandaSMT}}
LeipzigData \citep{goldhahn-etal-2012-building}, 
Lindat,\footnote{\url{https://lindat.cz/faq-repository}}
Lingala\_Song\_Lyrics,\footnote{\url{https://github.com/espoirMur/songs_lyrics_webscrap}}
Lyrics,\footnote{\url{https://lyricstranslate.com/}}
MC4 \citep{raffel2020exploring_t5}, 
MTData \citep{gowda-etal-2021-many}, 
MaCoCu \citep{macocu}, 
Makerere MT Corpus,\footnote{\url{https://zenodo.org/record/5089560}}
Masakhane community,\footnote{\url{https://github.com/masakhane-io/masakhane-community}}
Mburisano\_Covid,\footnote{\url{https://repo.sadilar.org/handle/20.500.12185/536}}
Menyo20K \citep{adelani-etal-2021-effect},
Minangkabau corpora \citep{koto-koto-2020-towards}, 
MoT \citep{palen-michel-etal-2022-multilingual},
NLLB\_seed \citep{costa2022no},
Nart/abkhaz,\footnote{\url{https://huggingface.co/datasets/Nart/abkhaz_text}}
OPUS \citep{opus_2012}, 
OSCAR \citep{suarez2019asynchronous}, 
ParaCrawl \citep{banon-etal-2020-paracrawl},
Parallel Corpora for Ethiopian Languages \citep{abate-etal-2018-parallel}, 
Phontron \citep{neubig11kftt},
QADI \citep{abdelali-etal-2021-qadi},
Quechua-IIC \citep{zevallos-etal-2022-introducing}, 
SLI\_GalWeb.1.0 \citep{agerri-etal-2018-developing},
Shami \citep{abu-kwaik-etal-2018-shami},
Stanford NLP,\footnote{\url{https://nlp.stanford.edu/}}
StatMT,\footnote{\url{https://statmt.org/}}
TICO \citep{anastasopoulos-etal-2020-tico},
TIL \citep{mirzakhalov-etal-2021-large},
Tatoeba,\footnote{\url{https://tatoeba.org/en/}}
TeDDi \citep{moran-etal-2022-teddi},
Tilde \citep{rozis-skadins-2017-tilde},
W2C \citep{w2c}, 
WAT \citep{nakazawa-etal-2022-overview},
WikiMatrix \citep{schwenk-etal-2021-wikimatrix}, 
Wikipedia,\footnote{\url{https://huggingface.co/datasets/wikipedia}}
Workshop on NER for South and South East Asian Languages \citep{singh-2008-named},
XLSum \citep{hasan-etal-2021-xl}.
We filter out Flores200 when processing the Glot500-c dataset.
Glot500-c includes texts in languages, i.e., Azerbaijani, Gujarati, Igbo, Oromo, Rundi, Tigrinya and Yoruba, from the XLSum dataset.

\subsection{Code}
\label{appendix:code_sourcing}

All the programming language splits are filtered using the following conditions:

\begin{itemize}
    \item For files forked more than 25 times, we retain them if the average line length is less than 120, the maximum line length is less than 300, and the alphanumeric fraction is more than 30\%.
    \item For files forked between 15 and 25 times, we retain them if the average line length is less than 90, the maximum line length is less than 150, and the alphanumeric fraction is more than 40\%.
    \item For files forked less than 15 times, we retain them if the average line length is less than 80, the maximum line length is less than 120, and the alphanumeric fraction is more than 45\%.
\end{itemize}

Subsequently, an aggressive MinHash deduplication pipeline with a threshold of 0.5 and a shingle size of 20 is applied. Finally, the resultant language splits are then capped at 5 million samples each.

\section{Additional Statistics of MaLA Corpus}
\label{sec:additiona_stats}

\subsection{Supported Languages}
\label{sec:supported_languages}
\begin{figure*}[ht!]
    \centering
    \begin{subfigure}[b]{\linewidth}
        \centering
        \includegraphics[width=\textwidth]{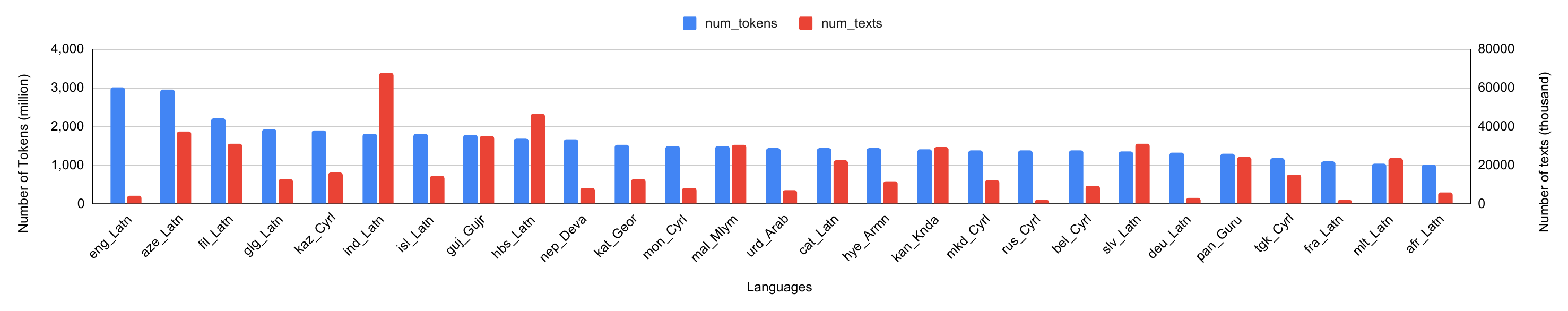}
        \caption{High}
        \label{fig:mala_counts_high}
    \end{subfigure}
    \hfill
    \begin{subfigure}[b]{\linewidth}
        \centering
        \includegraphics[width=\textwidth]{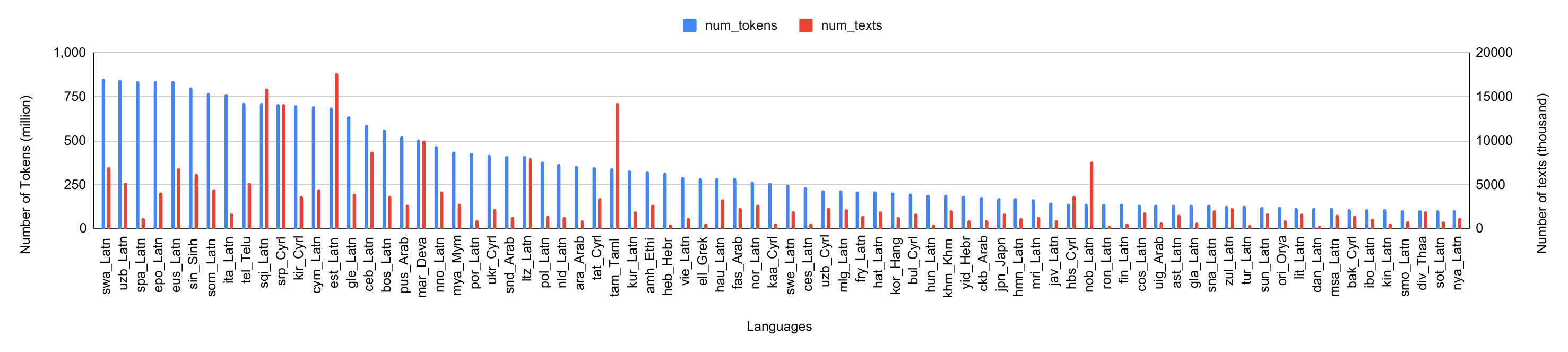}
        \caption{Medium-high}
        \label{fig:mala_counts_medium_high}
    \end{subfigure}
    \hfill
    \begin{subfigure}[b]{\linewidth}
        \centering
        \includegraphics[width=\textwidth]{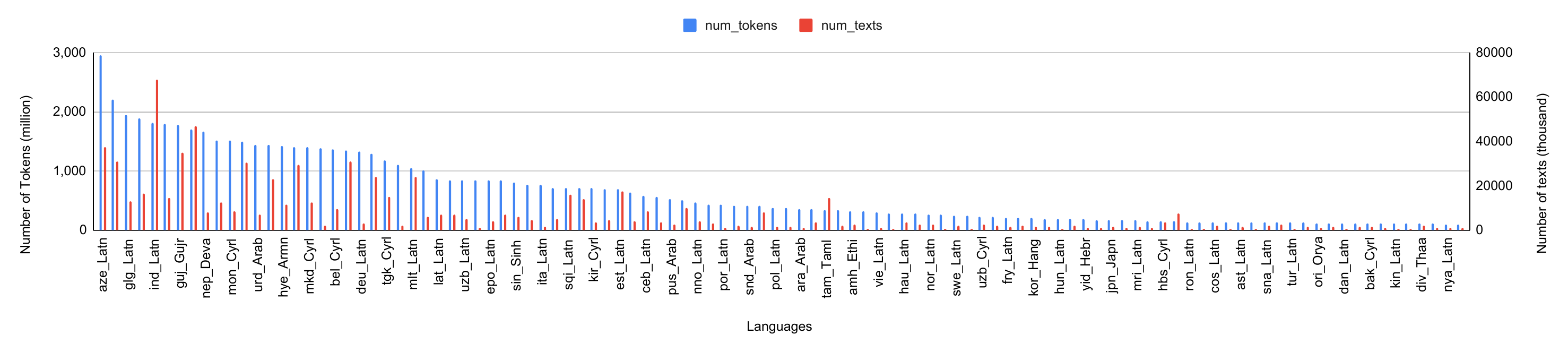}
        \caption{Medium}
        \label{fig:mala_counts_medium}
    \end{subfigure}
    \hfill
    \begin{subfigure}[b]{\linewidth}
        \centering
        \includegraphics[width=\textwidth]{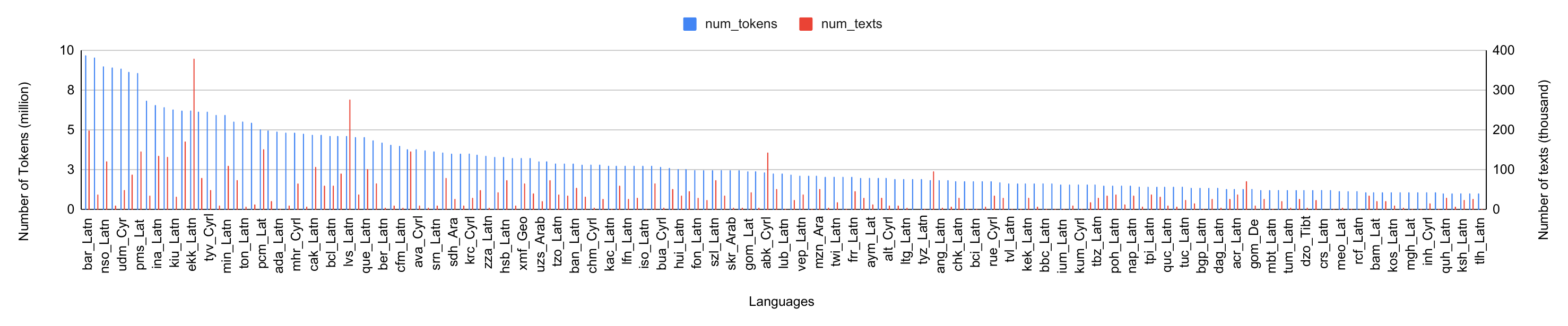}
        \caption{Medium-low}
        \label{fig:mala_counts_medium_low}
    \end{subfigure}
    \hfill
    \begin{subfigure}[b]{\linewidth}
        \centering
        \includegraphics[width=\textwidth]{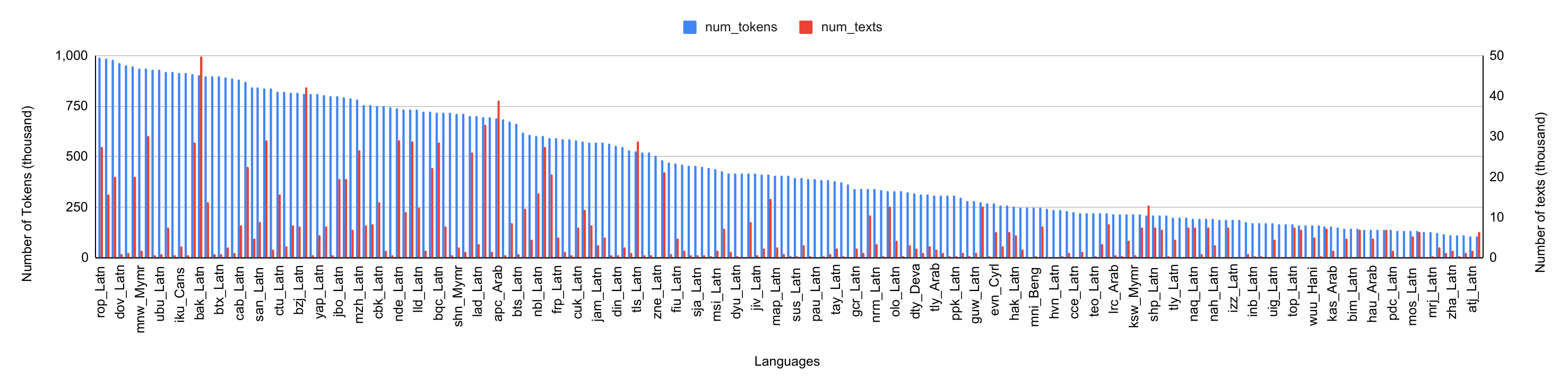}
        \caption{Low}
        \label{fig:mala_counts_low}
    \end{subfigure}
    \caption{The number of texts and tokens of MaLA corpus in different resource groups.}
    \label{fig:mala_counts}
\end{figure*}

\Cref{tab:languages} shows the languages codes of MaLA corpus, where ``unseen'' means the languages are not used for training EMMA-500.
The classification system for token counts categorizes language resources based on their size into five distinct tiers: ``high'' for resources exceeding 1 billion tokens, indicating a vast amount of data; ``medium-high'' for those with more than 100 million tokens, reflecting a substantial dataset; ``medium'' for resources that contain over 10 million tokens, representing a moderate size; ``medium-low'' for datasets with over ``1 million tokens'', indicating a smaller yet significant amount of data; and finally, ``low'' for resources containing less than 1 million tokens, which suggests a minimal data presence. This hierarchy helps in understanding the scale and potential utility of the language resources available.
\Cref{fig:mala_counts} shows the number of texts and tokens in different resource groups.

\begin{table*}[ht]
\caption{Languages by resource groups}
\label{tab:languages}
\centering
\scriptsize
\begin{tabular}{lrp{12cm}}
\toprule
\ch{Category} & \ch{Languages} & \ch{Language Codes} \\
\midrule
high & 27 & fra\_Latn, mon\_Cyrl, kat\_Geor, tgk\_Cyrl, kaz\_Cyrl, glg\_Latn, hbs\_Latn, kan\_Knda, mal\_Mlym, rus\_Cyrl, cat\_Latn, hye\_Armn, guj\_Gujr, slv\_Latn, fil\_Latn, bel\_Cyrl, isl\_Latn, nep\_Deva, mlt\_Latn, pan\_Guru, afr\_Latn, urd\_Arab, mkd\_Cyrl, aze\_Latn, deu\_Latn, eng\_Latn, ind\_Latn \\
low & 210 & prs\_Arab, nqo\_Nkoo, emp\_Latn, pfl\_Latn, teo\_Latn, gpe\_Latn, izz\_Latn, shn\_Mymr, hak\_Latn, pls\_Latn, evn\_Cyrl, djk\_Latn, toj\_Latn, nog\_Cyrl, ctu\_Latn, tca\_Latn, jiv\_Latn, ach\_Latn, mrj\_Latn, ajp\_Arab, apc\_Arab, tab\_Cyrl, hvn\_Latn, tls\_Latn, bak\_Latn, ndc\_Latn, trv\_Latn, top\_Latn, kjh\_Cyrl, guh\_Latn, mni\_Mtei, csy\_Latn, noa\_Latn, dov\_Latn, bho\_Deva, kon\_Latn, hne\_Deva, kcg\_Latn, mni\_Beng, hus\_Latn, pau\_Latn, jbo\_Latn, dtp\_Latn, kmb\_Latn, hau\_Arab, pdc\_Latn, nch\_Latn, acf\_Latn, bim\_Latn, ixl\_Latn, dty\_Deva, kas\_Arab, lrc\_Arab, alz\_Latn, lez\_Cyrl, lld\_Latn, tdt\_Latn, acm\_Arab, bih\_Deva, mzh\_Latn, guw\_Latn, rop\_Latn, rwo\_Latn, ahk\_Latn, qub\_Latn, kri\_Latn, gub\_Latn, laj\_Latn, sxn\_Latn, luo\_Latn, tly\_Latn, pwn\_Latn, mag\_Deva, xav\_Latn, bum\_Latn, ubu\_Latn, roa\_Latn, mah\_Latn, tsg\_Latn, gcr\_Latn, arn\_Latn, csb\_Latn, guc\_Latn, bat\_Latn, knj\_Latn, cre\_Latn, bus\_Latn, anp\_Deva, aln\_Latn, nah\_Latn, zai\_Latn, kpv\_Cyrl, enq\_Latn, gvl\_Latn, wal\_Latn, fiu\_Latn, swh\_Latn, crh\_Latn, nia\_Latn, bqc\_Latn, map\_Latn, atj\_Latn, npi\_Deva, bru\_Latn, din\_Latn, pis\_Latn, gur\_Latn, cuk\_Latn, zne\_Latn, cdo\_Latn, lhu\_Latn, pcd\_Latn, mas\_Latn, bis\_Latn, ncj\_Latn, ibb\_Latn, tay\_Latn, bts\_Latn, tzj\_Latn, bzj\_Latn, cce\_Latn, jvn\_Latn, ndo\_Latn, rug\_Latn, koi\_Cyrl, mco\_Latn, fat\_Latn, olo\_Latn, inb\_Latn, mkn\_Latn, qvi\_Latn, mak\_Latn, ktu\_Latn, nrm\_Latn, kua\_Latn, san\_Latn, nbl\_Latn, kik\_Latn, dyu\_Latn, sgs\_Latn, msm\_Latn, mnw\_Latn, zha\_Latn, sja\_Latn, xal\_Cyrl, rmc\_Latn, ami\_Latn, sda\_Latn, tdx\_Latn, yap\_Latn, tzh\_Latn, sus\_Latn, ikk\_Latn, bas\_Latn, nde\_Latn, dsb\_Latn, seh\_Latn, knv\_Latn, amu\_Latn, dwr\_Latn, iku\_Cans, uig\_Latn, bxr\_Cyrl, tcy\_Knda, mau\_Latn, aoj\_Latn, gor\_Latn, cha\_Latn, fip\_Latn, chr\_Cher, mdf\_Cyrl, arb\_Arab, quw\_Latn, shp\_Latn, spp\_Latn, frp\_Latn, ape\_Latn, cbk\_Latn, mnw\_Mymr, mfe\_Latn, jam\_Latn, lad\_Latn, awa\_Deva, mad\_Latn, ote\_Latn, shi\_Latn, btx\_Latn, maz\_Latn, ppk\_Latn, smn\_Latn, twu\_Latn, blk\_Mymr, msi\_Latn, naq\_Latn, tly\_Arab, wuu\_Hani, mos\_Latn, cab\_Latn, zlm\_Latn, gag\_Latn, suz\_Deva, ksw\_Mymr, gug\_Latn, nij\_Latn, nov\_Latn, srm\_Latn, jac\_Latn, nyu\_Latn, yom\_Latn, gui\_Latn \\
medium & 68 & tha\_Thai, kat\_Latn, lim\_Latn, tgk\_Arab, che\_Cyrl, lav\_Latn, xho\_Latn, war\_Latn, nan\_Latn, grc\_Grek, orm\_Latn, zsm\_Latn, cnh\_Latn, yor\_Latn, arg\_Latn, tgk\_Latn, azj\_Latn, tel\_Latn, slk\_Latn, pap\_Latn, zho\_Hani, sme\_Latn, tgl\_Latn, uzn\_Cyrl, als\_Latn, san\_Deva, azb\_Arab, ory\_Orya, lmo\_Latn, bre\_Latn, mvf\_Mong, fao\_Latn, oci\_Latn, sah\_Cyrl, sco\_Latn, tuk\_Latn, aze\_Arab, hin\_Deva, haw\_Latn, glk\_Arab, oss\_Cyrl, lug\_Latn, tet\_Latn, tsn\_Latn, hrv\_Latn, gsw\_Latn, arz\_Arab, vec\_Latn, mon\_Latn, ilo\_Latn, ctd\_Latn, ben\_Beng, roh\_Latn, kal\_Latn, asm\_Beng, srp\_Latn, bod\_Tibt, hif\_Latn, rus\_Latn, nds\_Latn, lus\_Latn, ido\_Latn, lao\_Laoo, tir\_Ethi, chv\_Cyrl, wln\_Latn, kaa\_Latn, pnb\_Arab \\
medium-high & 79 & div\_Thaa, som\_Latn, jpn\_Japn, hat\_Latn, sna\_Latn, heb\_Hebr, bak\_Cyrl, nld\_Latn, tel\_Telu, kin\_Latn, msa\_Latn, gla\_Latn, bos\_Latn, dan\_Latn, smo\_Latn, ita\_Latn, mar\_Deva, pus\_Arab, srp\_Cyrl, spa\_Latn, lat\_Latn, hmn\_Latn, sin\_Sinh, zul\_Latn, bul\_Cyrl, amh\_Ethi, ron\_Latn, tam\_Taml, khm\_Khmr, nno\_Latn, cos\_Latn, fin\_Latn, ori\_Orya, uig\_Arab, hbs\_Cyrl, gle\_Latn, cym\_Latn, vie\_Latn, kor\_Hang, lit\_Latn, yid\_Hebr, ara\_Arab, sqi\_Latn, pol\_Latn, tur\_Latn, swa\_Latn, hau\_Latn, ceb\_Latn, eus\_Latn, kir\_Cyrl, mlg\_Latn, jav\_Latn, snd\_Arab, sot\_Latn, por\_Latn, uzb\_Cyrl, fas\_Arab, nor\_Latn, est\_Latn, hun\_Latn, ibo\_Latn, ltz\_Latn, swe\_Latn, tat\_Cyrl, ast\_Latn, mya\_Mymr, uzb\_Latn, sun\_Latn, ell\_Grek, ces\_Latn, mri\_Latn, ckb\_Arab, kur\_Latn, kaa\_Cyrl, nob\_Latn, ukr\_Cyrl, fry\_Latn, epo\_Latn, nya\_Latn \\
medium-low & 162 & aym\_Latn, rue\_Cyrl, rom\_Latn, dzo\_Tibt, poh\_Latn, sat\_Olck, ary\_Arab, fur\_Latn, mbt\_Latn, bpy\_Beng, iso\_Latn, pon\_Latn, glv\_Latn, new\_Deva, gym\_Latn, bgp\_Latn, kac\_Latn, abt\_Latn, quc\_Latn, otq\_Latn, sag\_Latn, cak\_Latn, avk\_Latn, pam\_Latn, meo\_Latn, tum\_Latn, bam\_Latn, kha\_Latn, syr\_Syrc, kom\_Cyrl, nhe\_Latn, bal\_Arab, srd\_Latn, krc\_Cyrl, lfn\_Latn, bar\_Latn, rcf\_Latn, nav\_Latn, nnb\_Latn, sdh\_Arab, aka\_Latn, bew\_Cyrl, bbc\_Latn, meu\_Latn, zza\_Latn, ext\_Latn, yue\_Hani, ekk\_Latn, xmf\_Geor, nap\_Latn, mzn\_Arab, pcm\_Latn, lij\_Latn, myv\_Cyrl, scn\_Latn, dag\_Latn, ban\_Latn, twi\_Latn, udm\_Cyrl, som\_Arab, nso\_Latn, pck\_Latn, crs\_Latn, acr\_Latn, tat\_Latn, afb\_Arab, uzs\_Arab, hil\_Latn, mgh\_Latn, tpi\_Latn, ady\_Cyrl, pag\_Latn, kiu\_Latn, ber\_Latn, iba\_Latn, ksh\_Latn, plt\_Latn, lin\_Latn, chk\_Latn, tzo\_Latn, tlh\_Latn, ile\_Latn, lub\_Latn, hui\_Latn, min\_Latn, bjn\_Latn, szl\_Latn, kbp\_Latn, inh\_Cyrl, que\_Latn, ven\_Latn, vls\_Latn, kbd\_Cyrl, run\_Latn, wol\_Latn, ace\_Latn, ada\_Latn, kek\_Latn, yua\_Latn, tbz\_Latn, gom\_Latn, ful\_Latn, mrj\_Cyrl, abk\_Cyrl, tuc\_Latn, stq\_Latn, mwl\_Latn, tvl\_Latn, quh\_Latn, gom\_Deva, mhr\_Cyrl, fij\_Latn, grn\_Latn, zap\_Latn, mam\_Latn, mps\_Latn, tiv\_Latn, ksd\_Latn, ton\_Latn, bik\_Latn, vol\_Latn, ava\_Cyrl, tso\_Latn, szy\_Latn, ngu\_Latn, hyw\_Armn, fon\_Latn, skr\_Arab, kos\_Latn, tyz\_Latn, kur\_Arab, srn\_Latn, tyv\_Cyrl, bci\_Latn, vep\_Latn, crh\_Cyrl, kpg\_Latn, hsb\_Latn, ssw\_Latn, zea\_Latn, ewe\_Latn, ium\_Latn, diq\_Latn, ltg\_Latn, nzi\_Latn, guj\_Deva, ina\_Latn, pms\_Latn, bua\_Cyrl, lvs\_Latn, eml\_Latn, hmo\_Latn, kum\_Cyrl, kab\_Latn, chm\_Cyrl, cor\_Latn, cfm\_Latn, alt\_Cyrl, bcl\_Latn, ang\_Latn, frr\_Latn, mai\_Deva \\
unseen & 393 & rap\_Latn, pmf\_Latn, lsi\_Latn, dje\_Latn, bkx\_Latn, ipk\_Latn, syw\_Deva, ann\_Latn, bag\_Latn, bat\_Cyrl, chu\_Cyrl, gwc\_Arab, adh\_Latn, szy\_Hani, shi\_Arab, njy\_Latn, pdu\_Latn, buo\_Latn, cuv\_Latn, udg\_Mlym, bax\_Latn, tio\_Latn, kjb\_Latn, taj\_Deva, lez\_Latn, olo\_Cyrl, rnl\_Latn, bri\_Latn, inh\_Latn, kas\_Cyrl, wni\_Latn, anp\_Latn, tsc\_Latn, mgg\_Latn, udi\_Cyrl, mdf\_Latn, agr\_Latn, xty\_Latn, llg\_Latn, nge\_Latn, gan\_Latn, tuv\_Latn, stk\_Latn, nut\_Latn, thy\_Thai, lgr\_Latn, hnj\_Latn, dar\_Cyrl, aia\_Latn, lwl\_Thai, tnl\_Latn, tvs\_Latn, jra\_Khmr, tay\_Hani, gal\_Latn, ybi\_Deva, snk\_Arab, gag\_Cyrl, tuk\_Cyrl, trv\_Hani, ydd\_Hebr, kea\_Latn, gbm\_Deva, kwi\_Latn, hro\_Latn, rki\_Latn, quy\_Latn, tdg\_Deva, zha\_Hani, pcg\_Mlym, tom\_Latn, nsn\_Latn, quf\_Latn, jmx\_Latn, kqr\_Latn, mrn\_Latn, bxa\_Latn, abc\_Latn, mve\_Arab, lfa\_Latn, qup\_Latn, yin\_Latn, roo\_Latn, mrw\_Latn, nxa\_Latn, yrk\_Cyrl, bem\_Latn, kvt\_Latn, csw\_Cans, bjr\_Latn, mgm\_Latn, ngn\_Latn, pib\_Latn, quz\_Latn, awb\_Latn, myk\_Latn, otq\_Arab, ino\_Latn, tkd\_Latn, bef\_Latn, bug\_Bugi, aeu\_Latn, nlv\_Latn, dty\_Latn, bkc\_Latn, mmu\_Latn, hak\_Hani, sea\_Latn, mlk\_Latn, cbr\_Latn, lmp\_Latn, tnn\_Latn, qvz\_Latn, pbt\_Arab, cjs\_Cyrl, mlw\_Latn, mnf\_Latn, bfm\_Latn, dig\_Latn, thk\_Latn, zxx\_Latn, lkb\_Latn, chr\_Latn, pnt\_Latn, vif\_Latn, fli\_Latn, got\_Latn, hbb\_Latn, tll\_Latn, bug\_Latn, kxp\_Arab, qaa\_Latn, krr\_Khmr, kjg\_Laoo, isu\_Latn, kmu\_Latn, gof\_Latn, sdk\_Latn, mne\_Latn, baw\_Latn, idt\_Latn, xkg\_Latn, mgo\_Latn, dtr\_Latn, kms\_Latn, ffm\_Latn, hna\_Latn, nxl\_Latn, bfd\_Latn, odk\_Arab, miq\_Latn, mhx\_Latn, kam\_Latn, yao\_Latn, pnt\_Grek, kby\_Latn, kpv\_Latn, kbx\_Latn, cim\_Latn, qvo\_Latn, pih\_Latn, nog\_Latn, nco\_Latn, rmy\_Cyrl, clo\_Latn, dmg\_Latn, aaa\_Latn, rel\_Latn, ben\_Latn, loh\_Latn, thl\_Deva, chd\_Latn, cni\_Latn, cjs\_Latn, lbe\_Latn, ybh\_Deva, zxx\_Zyyy, awa\_Latn, gou\_Latn, xmm\_Latn, nqo\_Latn, rut\_Cyrl, kbq\_Latn, tkr\_Latn, dwr\_Ethi, ckt\_Cyrl, ady\_Latn, yea\_Mlym, nhx\_Latn, niv\_Cyrl, bwt\_Latn, xmg\_Latn, chy\_Latn, mfj\_Latn, hre\_Latn, bbk\_Latn, shn\_Latn, lrc\_Latn, qvc\_Latn, muv\_Mlym, mdr\_Latn, luy\_Latn, lzh\_Hani, fuh\_Latn, mle\_Latn, brx\_Deva, pex\_Latn, kau\_Latn, yrk\_Latn, hin\_Latn, ekm\_Latn, msb\_Latn, unr\_Orya, cac\_Latn, chp\_Cans, ckt\_Latn, bss\_Latn, lts\_Latn, bbj\_Latn, ttt\_Cyrl, kwu\_Latn, smn\_Cyrl, kpy\_Cyrl, tod\_Latn, wbm\_Latn, tcy\_Latn, arc\_Syrc, nst\_Latn, tuz\_Latn, bob\_Latn, bfn\_Latn, pli\_Deva, snl\_Latn, kwd\_Latn, lgg\_Latn, nza\_Latn, wbr\_Deva, lan\_Latn, kmz\_Latn, bzi\_Thai, hao\_Latn, nla\_Latn, qxr\_Latn, ken\_Latn, tbj\_Latn, blk\_Latn, ybb\_Latn, nwe\_Latn, gan\_Hani, snk\_Latn, kak\_Latn, tpl\_Latn, hla\_Latn, tks\_Arab, pea\_Latn, bya\_Latn, enc\_Latn, jgo\_Latn, tnp\_Latn, aph\_Deva, bgf\_Latn, brv\_Laoo, nod\_Thai, niq\_Latn, nwi\_Latn, xmd\_Latn, gbj\_Orya, syr\_Latn, ify\_Latn, xal\_Latn, bra\_Deva, cgc\_Latn, bhs\_Latn, pwg\_Latn, ang\_Runr, oki\_Latn, qve\_Latn, qvm\_Latn, bkm\_Latn, bkh\_Latn, niv\_Latn, zuh\_Latn, mry\_Latn, fiu\_Cyrl, ssn\_Latn, rki\_Mymr, sox\_Latn, yav\_Latn, nyo\_Latn, dag\_Arab, qxh\_Latn, bze\_Latn, myx\_Latn, zaw\_Latn, ddg\_Latn, wnk\_Latn, bwx\_Latn, mqy\_Latn, lad\_Hebr, boz\_Latn, lue\_Latn, ded\_Latn, pli\_Latn, avk\_Cyrl, wms\_Latn, sgd\_Latn, azn\_Latn, ajz\_Latn, psp\_Latn, jra\_Latn, smt\_Latn, ags\_Latn, csw\_Latn, wtk\_Latn, emp\_Cyrl, koi\_Latn, tkr\_Cyrl, amp\_Latn, ymp\_Latn, mfh\_Latn, tdb\_Deva, omw\_Latn, khb\_Talu, doi\_Deva, gld\_Cyrl, ava\_Latn, chu\_Latn, dnw\_Latn, azo\_Latn, dug\_Latn, bce\_Latn, kmr\_Latn, kpy\_Armn, abq\_Cyrl, trp\_Latn, ewo\_Latn, the\_Deva, hig\_Latn, pkb\_Latn, mxu\_Latn, oji\_Latn, tnt\_Latn, mzm\_Latn, mns\_Cyrl, lbe\_Cyrl, qvh\_Latn, kmg\_Latn, sps\_Latn, brb\_Khmr, tah\_Latn, sxb\_Latn, mkz\_Latn, mgq\_Latn, got\_Goth, lns\_Latn, arc\_Latn, akb\_Latn, skr\_Latn, nsk\_Cans, sml\_Latn, pce\_Mymr, eee\_Thai, lhm\_Deva, yux\_Cyrl, bqm\_Latn, bcc\_Arab, nas\_Latn, agq\_Latn, xog\_Latn, tsb\_Latn, fub\_Latn, mqj\_Latn, nsk\_Latn, bxr\_Latn, dln\_Latn, ozm\_Latn, rmy\_Latn, cre\_Cans, kim\_Cyrl, cuh\_Latn, ngl\_Latn, yas\_Latn, bud\_Latn, miy\_Latn, ame\_Latn, pnz\_Latn, raj\_Deva, enb\_Latn, cmo\_Khmr, saq\_Latn, tpu\_Khmr, eve\_Cyrl, cdo\_Hani \\
\bottomrule
\end{tabular}
\end{table*}

\subsection{Data Analysis}

\begin{figure}[ht!]
    \centering
    \includegraphics[width=0.9\linewidth]{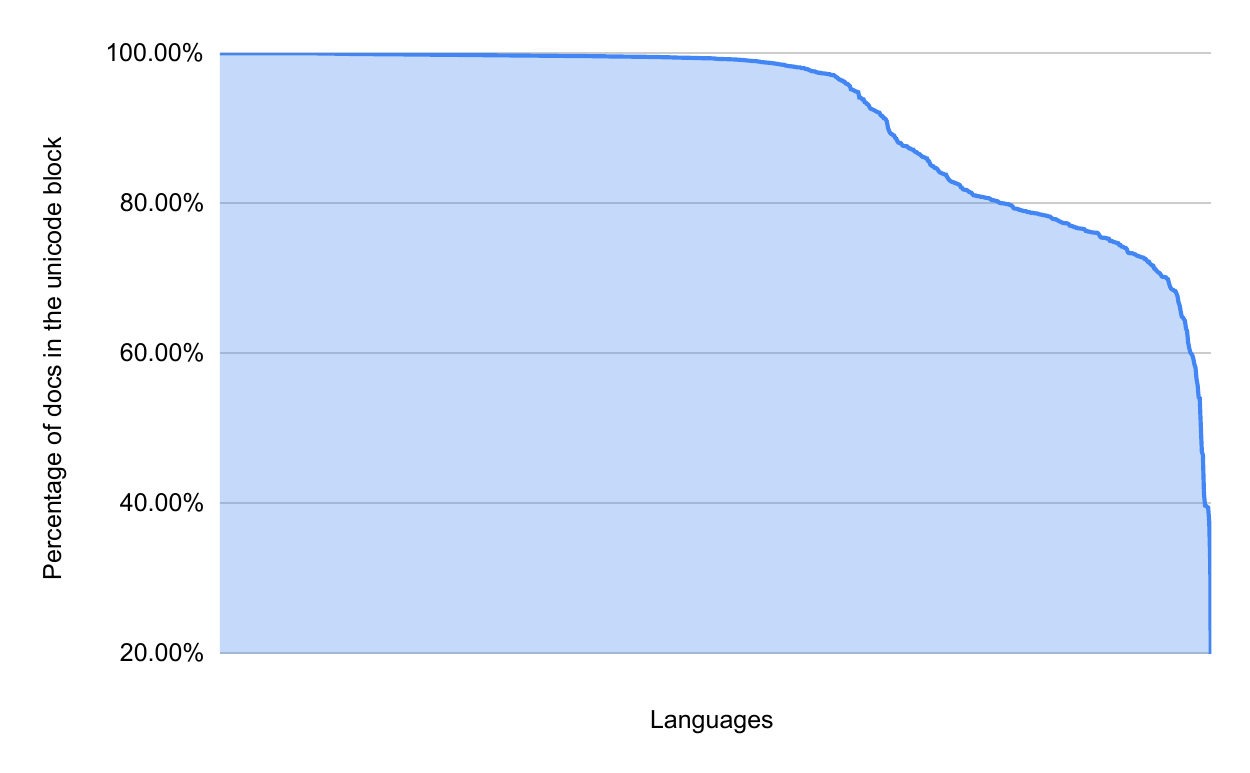}
    \caption{Unicode block distribution that measures the percentage of token counts falling into the Unicode block of each language}
    \label{fig:unicode}

    \centering
    \includegraphics[width=0.9\linewidth]{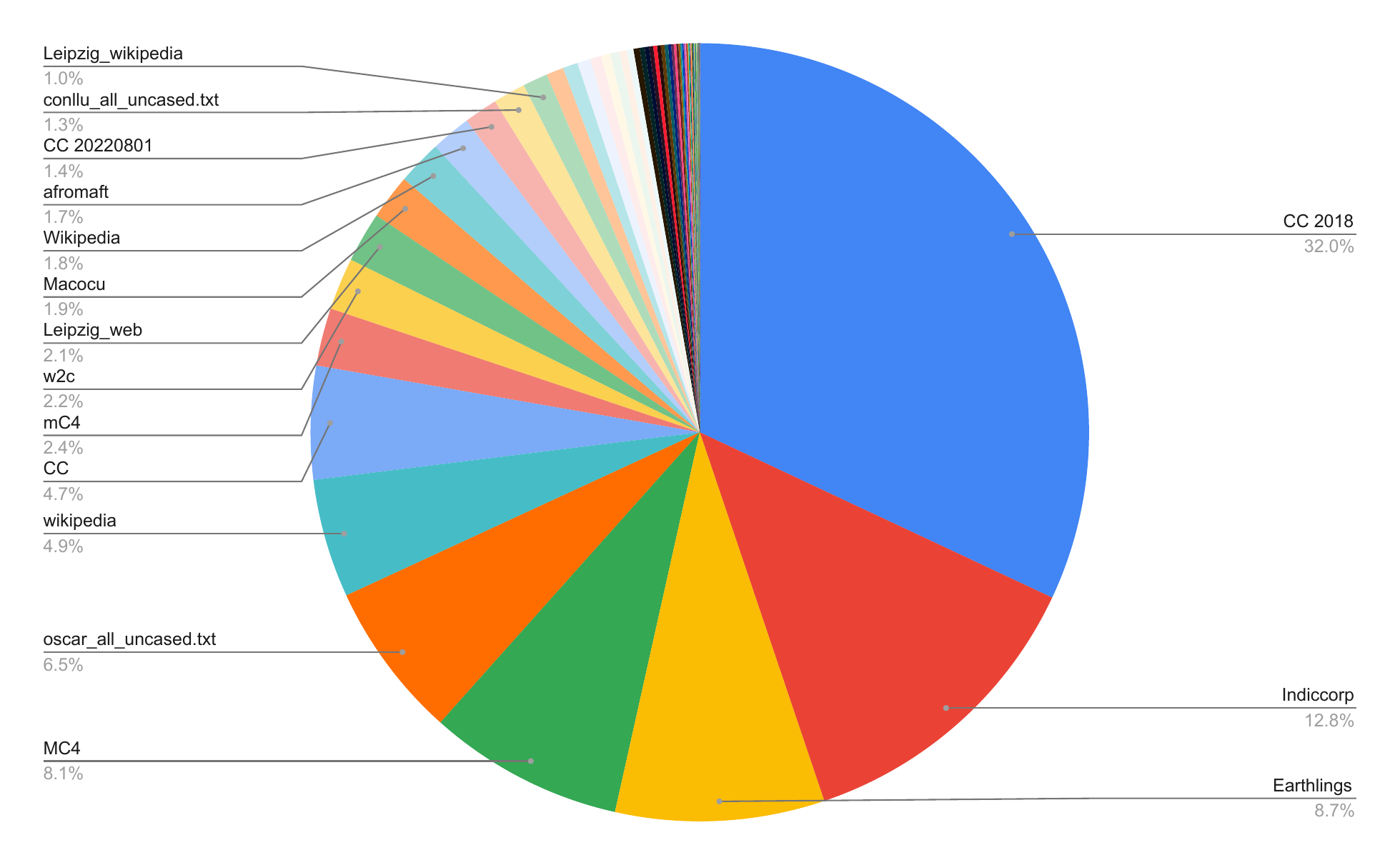}
    \caption{Data source distribution of MaLA corpus calculated by the number of documents}
    \label{fig:data_source}
\end{figure}

We examine the Unicode block distribution of each language, which counts the percentage of tokens falling into the Unicode block of each language.  
This aims to check whether language code conversion and writing system recognition are reasonably good. 
\Cref{fig:unicode} shows the Unicode block distribution. 
The result aligns with our observation of the presence of code-mixing, but in general, the majority of languages have tokens falling in their own Unicode blocks.

We also check the data source distribution as shown in \Cref{fig:data_source} to see where the texts in the MaLA corpus come from. 
The main source is common crawl, e.g., CC 2018, CC, OSCAR, and CC 20220801. 
A large number of documents come from Earthlings which comes from Glot500-c~\citep{imanigooghari2023glot500}. 
For web-crawled data with a URL in the original metadata of corpora like CulturaX \citep{nguyen2023culturax} and HPLT \citep{de2024new}, we extract the domain. 
Thus, the final corpus has many sources with a small portion.

\section{Evaluation Setup}
\label{sec:evaluation_setup_app}

\subsection{Prompts}
\label{sec:prompts}

\subsubsection{Machine Translation}
\label{sec:prompt_mt}

We use the prompt below for machine translation.
\begin{lstlisting}
Translate the following sentence from {src_lang} to {tgt_lang}
[{src_lang}]: {src_sent}
[{tgt_lang}]:
\end{lstlisting}

\subsubsection{Text Classification}
\label{sec:prompt_cls}

The prompt template for SIB-200 is as follows:
\begin{lstlisting}
Topic Classification: science/technology, travel, politics, sports, health, entertainment, geography.
{examples}
The topic of the news "${text}" is
\end{lstlisting}

\noindent For Taxi-1500, the prompt template is as follows:
\begin{lstlisting}
Topic Classification: Recommendation, Faith, Description, Sin, Grace, Violence.
{examples}
The topic of the verse "${text}" is
\end{lstlisting}

\section{Detailed Results}
This section presents detailed results of the evaluation.  
For benchmarks consisting of multiple languages, we host the results on GitHub\footnote{\url{https://github.com/MaLA-LM/emma-500/tree/main/evaluation_results}}.
Those benchmarks include ARC multilingual, BELEBELE, Glot500-c test set, PBC, SIB-200, Taxi-1500, FLORES200, XLSum, Aya evaluation suite, and PolyWrite.
For FLORES200, XLSum, Aya evaluation suite, and PolyWrite, we also release the generated texts of all compared models.

\begin{table*}[ht!]
\caption{0-shot results (ACC \%) on XCOPA in all languages}
\label{tab:xcopa_all}
\setlength{\tabcolsep}{2pt}
\resizebox{\linewidth}{!}{
\centering
\begin{tabular}{lrrrrrrrrrrrrrrrrrrrrrrr}
\toprule
                            \ch{Model} &    \ch{Avg} &  \ch{et-acc} &  \ch{stderr} &  \ch{ht-acc} &  \ch{stderr} &  \ch{id-acc} &  \ch{stderr} &  \ch{it-acc} &  \ch{stderr} &  \ch{qu-acc} &  \ch{stderr} &  \ch{sw-acc} &  \ch{stderr} &  \ch{ta-acc} &  \ch{stderr} &  \ch{th-acc} &  \ch{stderr} &  \ch{tr-acc} &  \ch{stderr} &  \ch{vi-acc} &  \ch{stderr} &  \ch{zh-acc} &  \ch{stderr} \\
\midrule
                       Llama 2 7B & 56.67 &   48.60 &    2.24 &   50.60 &    2.24 &   62.40 &    2.17 &   65.80 &    2.12 &   51.60 &    2.24 &   52.20 &    2.24 &   53.40 &    2.23 &   56.20 &    2.22 &   54.80 &    2.23 &   62.80 &    2.16 &   65.00 &    2.14 \\
                  Llama 2 7B Chat & 55.85 &   47.80 &    2.24 &   50.80 &    2.24 &   62.40 &    2.17 &   67.20 &    2.10 &   50.60 &    2.24 &   52.20 &    2.24 &   50.60 &    2.24 &   55.00 &    2.23 &   55.20 &    2.23 &   61.20 &    2.18 &   61.40 &    2.18 \\
                   CodeLlama 2 7B & 54.69 &   46.80 &    2.23 &   51.80 &    2.24 &   57.40 &    2.21 &   63.00 &    2.16 &   51.60 &    2.24 &   48.80 &    2.24 &   55.00 &    2.23 &   55.40 &    2.23 &   53.80 &    2.23 &   55.80 &    2.22 &   62.20 &    2.17 \\
                LLaMAX Llama 2 7B & 54.38 &   49.20 &    2.24 &   52.60 &    2.24 &   53.80 &    2.23 &   52.60 &    2.24 &   51.40 &    2.24 &   54.00 &    2.23 &   58.00 &    2.21 &   57.20 &    2.21 &   53.00 &    2.23 &   53.00 &    2.23 &   63.40 &    2.16 \\
         LLaMAX Llama 2 7B Alpaca & 56.60 &   51.20 &    2.24 &   54.20 &    2.23 &   57.20 &    2.21 &   61.00 &    2.18 &   52.40 &    2.24 &   55.00 &    2.23 &   57.00 &    2.22 &   56.40 &    2.22 &   55.20 &    2.23 &   55.20 &    2.23 &   67.80 &    2.09 \\
\bottomrule
\end{tabular}
}
\end{table*}

\begin{table*}[ht!]
\setlength{\tabcolsep}{3pt}
\centering
\caption{0-shot results (ACC \%) on XStoryCloze in all languages}
\label{tab:xstorycloze_all}
\setlength{\tabcolsep}{2pt}
\resizebox{\linewidth}{!}{
\begin{tabular}{lrrrrrrrrrrrrrrrrrrrrrrr}
\toprule
\ch{Model} & \ch{Avg} & \ch{ar-acc} & \ch{stderr} & \ch{en-acc} & \ch{stderr} & \ch{es-acc} & \ch{stderr} & \ch{eu-acc} & \ch{stderr} & \ch{hi-acc} & \ch{stderr} & \ch{id-acc} & \ch{stderr} & \ch{my-acc} & \ch{stderr} & \ch{ru-acc} & \ch{stderr} & \ch{sw-acc} & \ch{stderr} & \ch{te-acc} & \ch{stderr} & \ch{zh-acc} & \ch{stderr} \\
\midrule
                       Llama 2 7B & 57.55 &   49.90 &    1.29 &   77.04 &    1.08 &   67.37 &    1.21 &   50.36 &    1.29 &   53.74 &    1.28 &   59.23 &    1.26 &   48.05 &    1.29 &   63.00 &    1.24 &   50.50 &    1.29 &   54.33 &    1.28 &   59.56 &    1.26 \\
                  Llama 2 7B Chat & 58.41 &   50.50 &    1.29 &   78.69 &    1.05 &   67.11 &    1.21 &   50.83 &    1.29 &   54.07 &    1.28 &   59.63 &    1.26 &   48.64 &    1.29 &   65.52 &    1.22 &   52.02 &    1.29 &   53.34 &    1.28 &   62.21 &    1.25 \\
                   CodeLlama 2 7B & 55.68 &   50.10 &    1.29 &   71.48 &    1.16 &   63.40 &    1.24 &   50.43 &    1.29 &   49.70 &    1.29 &   55.86 &    1.28 &   49.37 &    1.29 &   59.23 &    1.26 &   50.03 &    1.29 &   53.74 &    1.28 &   59.17 &    1.26 \\
                LLaMAX Llama 2 7B & 60.36 &   58.90 &    1.27 &   75.51 &    1.11 &   65.25 &    1.23 &   54.47 &    1.28 &   58.17 &    1.27 &   60.62 &    1.26 &   52.48 &    1.29 &   61.22 &    1.25 &   57.18 &    1.27 &   59.30 &    1.26 &   60.82 &    1.26 \\
         LLaMAX Llama 2 7B Alpaca & 63.83 &   60.36 &    1.26 &   81.47 &    1.00 &   70.68 &    1.17 &   54.86 &    1.28 &   62.14 &    1.25 &   66.45 &    1.22 &   53.81 &    1.28 &   67.44 &    1.21 &   60.16 &    1.26 &   59.30 &    1.26 &   65.45 &    1.22 \\
          MaLA-500 Llama 2 10B v1 & 53.07 &   48.18 &    1.29 &   73.53 &    1.14 &   62.41 &    1.25 &   49.90 &    1.29 &   47.65 &    1.29 &   47.92 &    1.29 &   46.26 &    1.28 &   54.93 &    1.28 &   48.71 &    1.29 &   52.61 &    1.28 &   51.69 &    1.29 \\
\bottomrule
\end{tabular}
}
\end{table*}

\begin{table*}[ht!]
\setlength{\tabcolsep}{3pt}
\scriptsize
    \centering
    \caption{0-shot results (ACC \%) on XWinograd in all languages}
    \label{tab:xwinograd_all}
\begin{tabular}{lrrrrrrrrrrrrr}
\toprule
\ch{Model} & \ch{Avg} & \ch{en-acc} & \ch{stderr} & \ch{fr-acc} & \ch{stderr} & \ch{jp-acc} & \ch{stderr} & \ch{pt-acc} & \ch{stderr} & \ch{ru-acc} & \ch{stderr} & \ch{zh-acc} & \ch{stderr} \\
\midrule
                       Llama 2 7B & 72.47 &   87.91 &    0.68 &   66.27 &    5.22 &   70.28 &    1.48 &   72.24 &    2.77 &   68.25 &    2.63 &   69.84 &    2.05 \\
                  Llama 2 7B Chat & 69.45 &   85.55 &    0.73 &   71.08 &    5.01 &   68.40 &    1.50 &   64.64 &    2.95 &   65.71 &    2.68 &   61.31 &    2.17 \\
                   CodeLlama 2 7B & 70.92 &   84.52 &    0.75 &   67.47 &    5.17 &   63.61 &    1.55 &   72.24 &    2.77 &   66.67 &    2.66 &   71.03 &    2.02 \\
                LLaMAX Llama 2 7B & 67.49 &   77.89 &    0.86 &   61.45 &    5.37 &   71.01 &    1.47 &   65.40 &    2.94 &   55.56 &    2.80 &   73.61 &    1.97 \\
         LLaMAX Llama 2 7B Alpaca & 69.86 &   82.75 &    0.78 &   66.27 &    5.22 &   72.37 &    1.44 &   66.16 &    2.92 &   59.37 &    2.77 &   72.22 &    2.00 \\
          MaLA-500 Llama 2 10B v1 & 65.89 &   83.66 &    0.77 &   63.86 &    5.31 &   60.17 &    1.58 &   69.20 &    2.85 &   61.90 &    2.74 &   56.55 &    2.21 \\
          MaLA-500 Llama 2 10B v2 & 65.89 &   83.66 &    0.77 &   63.86 &    5.31 &   60.17 &    1.58 &   69.20 &    2.85 &   61.90 &    2.74 &   56.55 &    2.21 \\
\bottomrule
\end{tabular}
\end{table*}

\begin{table*}[ht!]
\centering
\caption{Pass@1 on Multipl-E.}
\label{tab:me_p1}
\scriptsize
\begin{tabular}{lrrrrrrr}
\toprule
\ch{Model} & \ch{C++} & \ch{C\#} & \ch{Java} & \ch{JavaScript} & \ch{Python} & \ch{Rust} & \ch{TypeScript} \\
\midrule
    Llama 2 7B & 6.74\% & 5.65\% & 8.54\% & 11.34\% & 11.98\% & 6.12\% & 12.04\%\\
    CodeLlama 2 7B & 26.70\% & 20.44\% & 30.56\% & 32.89\% & 28.76\% & 26.23\% & 33.45\%\\
    LLaMAX 2 7B & 0.0\% & 0.0\% & 0.02\% & 0.68\% & 0.0\% & 0.0\% & 1.78\%\\
    MaLA-500 Llama 2 10B V2 & 0.0\% & 0.0\% & 0.0\% & 0.0\% & 0.0\% & 0.0\% & 0.0\%\\
    TowerBase Llama 2 7B & 1.01\% & 6.85\% & 7.86\% & 0.87\% & 8.68\% & 0.0\% & 0.0\%\\
\midrule
    Occiglot Mistral 7B v0.1 & 23.16\% & 16.14\% & 19.83\% & 26.92\% & 21.45\% & 16.34\% & 24.97\%\\
    Bloom 7B & 5.25\% & 2.97\% & 6.37\% & 6.76\% & 7.65\% & 1.01\% & 7.34\%\\
    BloomZ 2 7B & 6.87\% & 3.59\% & 6.02\% & 6.96\% & 7.33\% & 2.13\% & 8.04\%\\
    Aya23 8B & 16.03\% & 7.91\% & 14.29\% & 6.23\% & 3.56\% & 11.43\% & 4.90\%\\
\midrule
    Mistral 7B v0.3 & 26.12\% & 22.87\% & 25.54\% & 35.24\% & 24.91\% & 19.24\% & 28.76\%\\
    Llama 3 8B & 34.12\% & 21.06\% & 26.56\% & 36.12\% & 30.22\% & 25.19\% & 37.34\%\\
    LLaMAX 3 8B & 0.0\% & 0.0\% & 0.0\% & 9.73\% & 0.91\% & 0.21\% & 10.12\%\\
    Gemma 7B & 29.66\% & 21.40\% & 27.35\% & 35.29\% & 30.11\% & 25.48\% & 30.57\%\\
    CodeGemma 7B & 33.91\% & 21.05\% & 29.43\% & 37.78\% & 32.56\% & 28.70\% & 37.14\%\\
    Qwen 1.5 7B & 22.04\% & 12.42\% & 17.68\% & 27.58\% & 32.11\% & 8.32\% & 27.21\%\\
    Qwen 2 7B & 43.51\% & 21.47\% & 38.95\% & 46.31\% & 37.49\% & 34.61\% & 48.41\%\\
\midrule
    EMMA-500 Llama 2 7B & 11.34\% & 8.94\% & 11.93\% & 11.67\% & 18.97\% & 6.86\% & 9.94\%\\

\bottomrule
\end{tabular}

\centering
\caption{Pass@10 on Multipl-E.}
\label{tab:me_p10}
\scriptsize
\begin{tabular}{lrrrrrrr}
\toprule
\ch{Model} & \ch{C++} & \ch{C\#} & \ch{Java} & \ch{JavaScript} & \ch{Python} & \ch{Rust} & \ch{TypeScript} \\
\midrule
    Llama 2 7B & 17.92\% & 14.22\% & 20.78\% & 23.12\% & 24.86\% & 13.08\% & 22.19\%\\
    CodeLlama 2 7B & 51.39\% & 37.13\% & 50.38\% & 59.08\% & 56.58\% & 47.68\% & 53.60\%\\
    LLaMAX 2 7B & 0.96\% & 0.0\% & 0.53\% & 3.99\% & 0.0\% & 0.12\% & 5.64\%\\
    MaLA-500 Llama 2 10B V2 & 0.0\% & 0.0\% & 0.0\% & 0.0\% & 0.0\% & 0.0\% & 0.0\%\\
    TowerBase Llama 2 7B & 1.01\% & 6.85\% & 7.86\% & 0.87\% & 8.68\% & 0.0\% & 0.0\%\\
\midrule
    Occiglot Mistral 7B v0.1 & 36.08\% & 27.49\% & 35.87\% & 47.45\% & 41.76\% & 29.80\% & 43.13\%\\
    Bloom 7B & 11.35\% & 6.86\% & 12.79\% & 12.44\% & 14.24\% & 3.68\% & 12.08\%\\
    BloomZ 2 7B & 12.55\% & 8.21\% & 12.94\% & 14.10\% & 14.84\% & 3.91\% & 13.27\%\\
    Aya23 8B & 28.49\% & 17.34\% & 27.12\% & 23.19\% & 26.72\% & 26.14\% & 15.67\%\\
\midrule
    Mistral 7B v0.3 & 50.23\% & 35.18\% & 46.83\% & 57.23\% & 54.29\% & 42.77\% & 54.22\%\\
    Llama 3 8B & 55.13\% & 36.67\% & 54.34\% & 62.65\% & 59.24\% & 45.68\% & 63.02\%\\
    LLaMAX 3 8B & 0.72\% & 0.0\% & 0.97\% & 22.91\% & 1.58\% & 1.38\% & 23.04\%\\
    Gemma 7B & 55.21\% & 39.09\% & 52.02\% & 61.88\% & 60.09\% & 50.34\% & 61.23\%\\
    CodeGemma 7B & 62.29\% & 45.11\% & 59.74\% & 70.96\% & 69.76\% &
 61.27\% & 72.76\%\\
    Qwen 1.5 7B & 40.11\% & 28.95\% & 37.56\% & 48.35\% & 40.34\% & 20.19\% & 44.85\%\\
    Qwen 2 7B & 64.58\% & 43.17\% & 63.28\% & 73.21\% & 54.34\% & 65.48\% & 74.32\%\\
\midrule
    EMMA-500 Llama 2 7B & 22.84\% & 14.29\% & 21.28\% & 18.22\% & 24.94\% & 15.68\% & 15.91\%\\

\bottomrule
\end{tabular}

\centering
\caption{Pass@25 on Multipl-E.}
\label{tab:me_p25}
\scriptsize
\begin{tabular}{lrrrrrrr}
\toprule
\ch{Model} & \ch{C++} & \ch{C\#} & \ch{Java} & \ch{JavaScript} & \ch{Python} & \ch{Rust} & \ch{TypeScript} \\
\midrule
    Llama 2 7B & 24.89\% & 18.77\% & 26.45\% & 30.31\% & 31.09\% & 18.70\% & 29.55\%\\
    CodeLlama 2 7B & 64.12\% & 45.96\% & 61.98\% & 72.76\% & 71.77\% & 59.98\% & 70.89\%\\
    LLaMAX 2 7B & 1.47\% & 0.0\% & 1.04\% & 6.78\% & 0.0\% & 0.34\% & 9.04\%\\
    MaLA-500 Llama 2 10B V2 & 0.0\% & 0.0\% & 0.0\% & 0.0\% & 0.0\% & 0.0\% & 0.0\%\\
    TowerBase Llama 2 7B & 1.86\% & 15.74\% & 22.45\% & 2.53\% & 20.18\% & 0.0\% & 0.0\%\\
\midrule
    Occiglot Mistral 7B v0.1 & 44.17\% & 33.89\% & 42.77\% & 56.63\% & 49.87\% & 37.94\% & 55.76\%\\
    Bloom 7B & 17.27\% & 10.07\% & 17.23\% & 17.55\% & 17.93\% & 5.70\% & 16.81\%\\
    BloomZ 2 7B & 17.41\% & 11.98\% & 18.66\% & 18.37\% & 18.78\% & 6.83\% & 18.31\%\\
    Aya23 8B & 37.56\% & 21.87\% & 36.45\% & 33.57\% & 36.09\% & 35.34\% & 23.75\%\\
\midrule
    Mistral 7B v0.3 & 60.93\% & 44.80\% & 56.12\% & 66.34\% & 65.33\% & 55.31\% & 64.51\%\\
    Llama 3 8B & 64.42\% & 43.34\% & 62.14\% & 73.51\% & 71.82\% & 57.78\% & 75.09\%\\
    LLaMAX 3 8B & 1.71\% & 0.0\% & 2.32\% & 35.61\% & 2.77\% & 2.08\% & 30.19\%\\
    Gemma 7B & 66.14\% & 47.08\% & 63.24\% & 70.22\% & 70.47\% & 63.52\% & 72.56\%\\
    CodeGemma 7B & 73.66\% & 50.79\% & 69.96\% & 79.48\% & 78.68\% & 74.87\% & 81.12\%\\
    Qwen 1.5 7B & 51.87\% & 37.69\% & 48.71\% & 59.71\% & 49.69\% & 29.78\% & 55.98\%\\
    Qwen 2 7B & 74.33\% & 51.87\% & 72.03\% & 81.67\% & 65.38\% & 74.97\% & 80.59\%\\
\midrule
    EMMA-500 Llama 2 7B & 31.93\% & 18.69\% & 29.85\% & 26.45\% & 32.55\% & 21.32\% & 22.34\%\\

\bottomrule
\end{tabular}
\end{table*}

\begin{table*}[ht!]
\caption{0-shot results (ACC \%) on XNLI in all languages.}
\label{tab:xnli_all}
\setlength{\tabcolsep}{2pt}
\resizebox{\linewidth}{!}{
\begin{tabular}{lrrrrrrrrrrrrrrrrrrrrrrrrrrrrrrr}
\toprule
\ch{Model} & \ch{Avg} & \ch{ar-acc} & \ch{stderr} & \ch{bg-acc} & \ch{stderr} & \ch{de-acc} & \ch{stderr} & \ch{el-acc} & \ch{stderr} & \ch{en-acc} & \ch{stderr} & \ch{es-acc} & \ch{stderr} & \ch{fr-acc} & \ch{stderr} & \ch{hi-acc} & \ch{stderr} & \ch{ru-acc} & \ch{stderr} & \ch{sw-acc} & \ch{stderr} & \ch{th-acc} & \ch{stderr} & \ch{tr-acc} & \ch{stderr} & \ch{ur-acc} & \ch{stderr} & \ch{vi-acc} & \ch{stderr} & \ch{zh-acc} & \ch{stderr} \\
\midrule
                       Llama 2 7B & 40.19 &   35.42 &    0.96 &   42.65 &    0.99 &   47.11 &    1.00 &   36.67 &    0.97 &   55.30 &    1.00 &   40.52 &    0.98 &   50.08 &    1.00 &   37.71 &    0.97 &   42.37 &    0.99 &   34.94 &    0.96 &   36.35 &    0.96 &   37.27 &    0.97 &   33.61 &    0.95 &   36.63 &    0.97 &   36.18 &    0.96 \\
                  Llama 2 7B Chat & 38.58 &   34.42 &    0.95 &   37.07 &    0.97 &   43.09 &    0.99 &   38.15 &    0.97 &   50.24 &    1.00 &   39.44 &    0.98 &   44.82 &    1.00 &   35.78 &    0.96 &   42.09 &    0.99 &   34.22 &    0.95 &   33.49 &    0.95 &   36.95 &    0.97 &   33.90 &    0.95 &   38.11 &    0.97 &   36.95 &    0.97 \\
                   CodeLlama 2 7B & 40.19 &   33.41 &    0.95 &   37.75 &    0.97 &   47.23 &    1.00 &   37.63 &    0.97 &   54.78 &    1.00 &   44.38 &    1.00 &   49.20 &    1.00 &   35.94 &    0.96 &   46.06 &    1.00 &   33.29 &    0.94 &   35.02 &    0.96 &   38.59 &    0.98 &   33.25 &    0.94 &   40.40 &    0.98 &   35.94 &    0.96 \\
                LLaMAX Llama 2 7B & 44.27 &   33.78 &    0.95 &   46.83 &    1.00 &   48.96 &    1.00 &   42.57 &    0.99 &   54.90 &    1.00 &   47.59 &    1.00 &   47.79 &    1.00 &   45.50 &    1.00 &   45.54 &    1.00 &   43.05 &    0.99 &   41.85 &    0.99 &   43.29 &    0.99 &   44.18 &    1.00 &   43.86 &    0.99 &   34.38 &    0.95 \\
\bottomrule
\end{tabular}
}
\end{table*}

\Cref{tab:mgsm_direct_all,tab:mgsm_cot_all} show 3-shot results on MGSM by direct and Chain-of-Thought prompting respectively. 
All the scores are obtained by flexible matching.

\begin{table*}[ht!]
\caption{3-shot results (ACC \%) on MGSM in all languages by direct prompting and flexible matching.}
\label{tab:mgsm_direct_all}
\setlength{\tabcolsep}{2pt}
\resizebox{\linewidth}{!}{
\begin{tabular}{lrrrrrrrrrrrrrrrrrrrrrrr}
\toprule
\ch{Model} & \ch{Avg} & \ch{bn} & \ch{bn-stderr} & \ch{de} & \ch{de-stderr} & \ch{en} & \ch{en-stderr} & \ch{es} & \ch{es-stderr} & \ch{fr} & \ch{fr-stderr} & \ch{ja} & \ch{ja-stderr} & \ch{ru} & \ch{ru-stderr} & \ch{sw} & \ch{sw-stderr} & \ch{te} & \ch{te-stderr} & \ch{th} & \ch{th-stderr} & \ch{zh} & \ch{zh-stderr} \\
\midrule
                       Llama 2 7B &          6.69 &  2.80 &       1.05 &  8.00 &       1.72 & 17.60 &       2.41 & 11.20 &       2.00 & 12.00 &       2.06 &  2.40 &       0.97 &  8.00 &       1.72 &  2.80 &       1.05 &  1.20 &       0.69 &  0.80 &       0.56 &  6.80 &       1.60 \\
                  Llama 2 7B Chat &         10.22 &  2.80 &       1.05 & 16.80 &       2.37 & 22.80 &       2.66 & 19.60 &       2.52 & 19.20 &       2.50 &  2.40 &       0.97 & 14.40 &       2.22 &  0.80 &       0.56 &  0.80 &       0.56 &  2.80 &       1.05 & 10.00 &       1.90 \\
                   CodeLlama 2 7B &          5.93 &  1.60 &       0.80 &  8.80 &       1.80 & 12.80 &       2.12 &  8.80 &       1.80 & 10.40 &       1.93 &  5.60 &       1.46 &  6.00 &       1.51 &  2.00 &       0.89 &  1.20 &       0.69 &  5.20 &       1.41 &  2.80 &       1.05 \\
                LLaMAX Llama 2 7B &          3.35 &  2.80 &       1.05 &  3.60 &       1.18 &  6.00 &       1.51 &  2.00 &       0.89 &  7.20 &       1.64 &  3.20 &       1.12 &  2.40 &       0.97 &  1.60 &       0.80 &  0.80 &       0.56 &  1.60 &       0.80 &  5.60 &       1.46 \\
         LLaMAX Llama 2 7B Alpaca &          5.05 &  4.00 &       1.24 &  3.60 &       1.18 & 10.80 &       1.97 &  6.00 &       1.51 &  6.40 &       1.55 &  3.20 &       1.12 &  4.40 &       1.30 &  4.80 &       1.35 &  1.60 &       0.80 &  3.20 &       1.12 &  7.60 &       1.68 \\
\bottomrule
\end{tabular}
}
\end{table*}

\begin{table*}[ht!]
\caption{3-shot results (ACC \%) on MGSM in all languages by CoT prompting and flexible matching.}
\label{tab:mgsm_cot_all}
\setlength{\tabcolsep}{2pt}
\resizebox{\linewidth}{!}{
\begin{tabular}{lrrrrrrrrrrrrrrrrrrrrrrr}
\toprule
\ch{Model} & \ch{Avg} & \ch{bn} & \ch{bn-stderr} & \ch{de} & \ch{de-stderr} & \ch{en} & \ch{en-stderr} & \ch{es} & \ch{es-stderr} & \ch{fr} & \ch{fr-stderr} & \ch{ja} & \ch{ja-stderr} & \ch{ru} & \ch{ru-stderr} & \ch{sw} & \ch{sw-stderr} & \ch{te} & \ch{te-stderr} & \ch{th} & \ch{th-stderr} & \ch{zh} & \ch{zh-stderr} \\
\midrule
                       Llama 2 7B &                6.36 &  2.00 &       0.89 &  7.60 &       1.68 & 16.00 &       2.32 & 12.40 &       2.09 &  9.20 &       1.83 &  3.60 &       1.18 &  8.00 &       1.72 &  2.00 &       0.89 &  0.80 &       0.56 &  1.60 &       0.80 &  8.00 &       1.72 \\
                  Llama 2 7B Chat &               10.91 &  2.40 &       0.97 & 17.60 &       2.41 & 27.20 &       2.82 & 19.20 &       2.50 & 18.80 &       2.48 &  4.00 &       1.24 & 15.20 &       2.28 &  2.00 &       0.89 &  0.80 &       0.56 &  2.80 &       1.05 & 11.60 &       2.03 \\
                   CodeLlama 2 7B &                6.64 &  1.60 &       0.80 &  9.20 &       1.83 & 13.20 &       2.15 & 10.80 &       1.97 & 10.80 &       1.97 &  4.40 &       1.30 &  7.60 &       1.68 &  1.20 &       0.69 &  1.60 &       0.80 &  6.00 &       1.51 &  4.00 &       1.24 \\
                LLaMAX Llama 2 7B &                3.62 &  3.60 &       1.18 &  3.60 &       1.18 &  7.20 &       1.64 &  3.60 &       1.18 &  5.20 &       1.41 &  2.80 &       1.05 &  2.80 &       1.05 &  0.80 &       0.56 &  0.80 &       0.56 &  2.00 &       0.89 &  4.80 &       1.35 \\
         LLaMAX Llama 2 7B Alpaca &                6.35 &  3.20 &       1.12 &  4.80 &       1.35 & 15.20 &       2.28 &  9.60 &       1.87 &  5.20 &       1.41 &  4.40 &       1.30 &  3.60 &       1.18 &  4.80 &       1.35 &  0.40 &       0.40 &  4.80 &       1.35 &  6.80 &       1.60 \\
\bottomrule
\end{tabular}
}
\end{table*}

 \section{Open-Ended Generation}

We have moved the open-ended generation task from the main text to the appendix due to the challenges in establishing a reliable evaluation method. 
Automated metrics, while useful, often fail to capture the full quality of open-ended responses, particularly in a multilingual and low-resource setting. 
Similarly, LLM-as-judge approaches can inherit biases from the models they rely on, leading to inconsistent or skewed assessments. Reward modeling, though promising, remains an active research area and is not yet robust enough for evaluating diverse multilingual outputs. 
Given these limitations, we opted to present the results of automatic metrics from the first version of this paper in the appendix for transparency while keeping the focus of the main text on tasks with more reliable evaluation methods

 \subsection{Evaluation Benchmarks}
 \label{sec:ploywrite}

\begin{figure}[ht!]
     \centering
     \includegraphics[width=\linewidth]{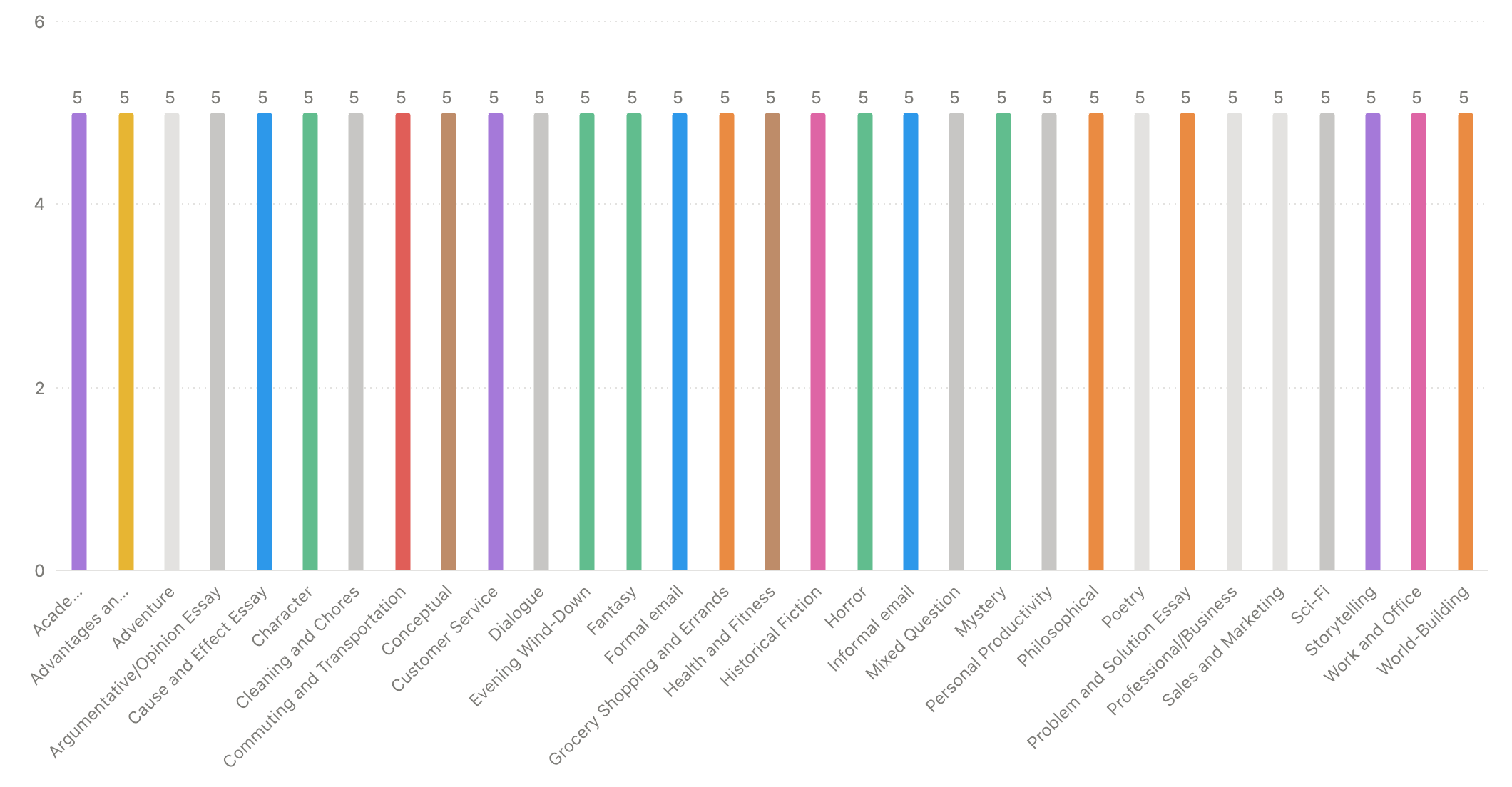}
     \caption{Writing tasks in the PolyWrite dataset.}
     \label{fig:polywrite-tasks}
     \vspace{3ex}
     \centering
     \includegraphics[width=\linewidth]{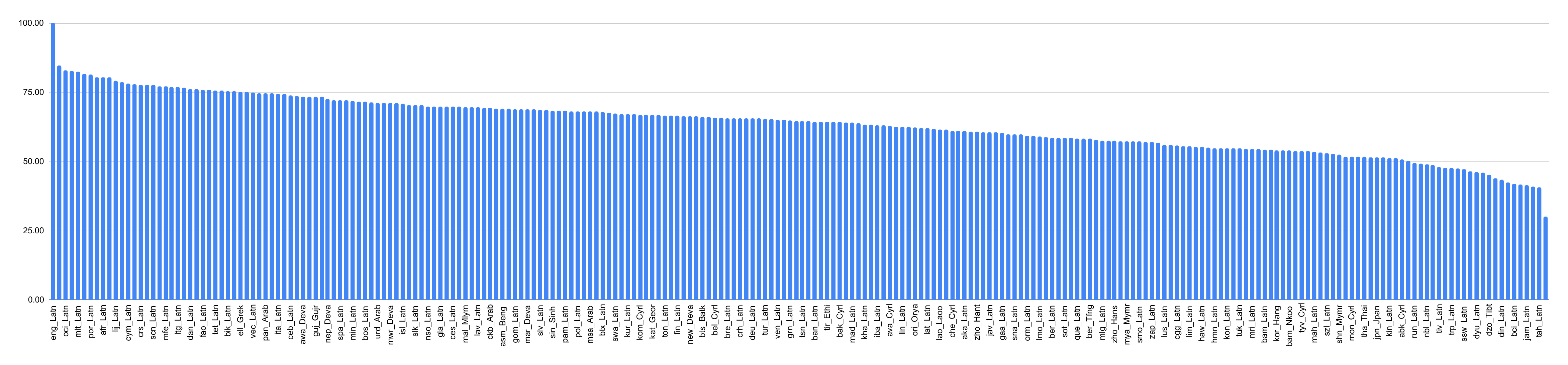}
     \caption{Mean BLEU scores per language in the PolyWrite dataset.}
     \label{fig:polywrite-bleu}
 \end{figure}

 To facilitate the evaluation on open-ended generation. We build a benchmark, called PolyWrite, for writing tasks.
 The PolyWrite dataset has 51 writing tasks with the number of prompts per task shown in \Cref{fig:polywrite-tasks}.
 We use Google Translate to translate the English prompts into 240 languages and back-translate for translation quality assessment. 

 The metadata of PolyWrite includes several key fields. 
 The \texttt{category} specifies the task type. 
 The \texttt{name} field typically holds the specific identifier for each prompt, while \texttt{prompt\_en} contains the English version of the prompt. 
 \texttt{lang\_script} identifies the language and script used, ensuring correct language processing.
 The \texttt{prompt\_translated} field holds the translated prompt in the target language, and \texttt{prompt\_backtranslated} contains the back-translated version to assess translation quality. Both \texttt{bleu} and \texttt{chrf++} fields provide numeric evaluation metrics, with BLEU and chrF++ scores measuring the quality of the generated text.
 Finally, the \texttt{uuid} ensures a unique identifier for each dataset entry, allowing for precise reference and tracking of individual prompts. To mitigate errors introduced by the machine translation process, we filter out prompts with a BLEU score of less than 20. \Cref{fig:polywrite-bleu} shows the average BLEU score of each language in the final dataset. 

\subsection{Evaluation Results}
 \paragraph{Aya Evaluation}
 We choose the two subsets \texttt{aya-human-annotated} and \texttt{dolly-machine-translated} from the Aya evaluation suite~\citep{singh-etal-2024-aya}, which have both inputs and targets for subsequent evaluation. 
 To quantitatively assess the quality of the generated text by the models, we employ two metrics: BLEU~\citep{papineni-etal-2002-bleu} and Self-BLEU~\citep{zhu2018texygen}. 
 BLEU is crucial for assessing the linguistic accuracy and relevance of the generated text in comparison to the expected human-like text present in the dataset. 
 Self-BLEU is used to evaluate the diversity of the text generated by a model. 
 It measures how similar different texts from the same model are to each other by treating one generated text as the ``candidate'' and others as the ``reference'' texts. 
 This metric is useful in scenarios where high degrees of variation are desirable, as it helps identify models that might be overfitting to particular styles or patterns of text.

 The results are presented in \Cref{tab:aya_polywrite}. 
 The BLEU metric is an indicator that measures how generated texts are close to the references. However, in open-ended generation, LLMs cannot generate identical texts to references, leading to low BLEU scores.
 The EMMA-500 model obtains remarkably high BLEU scores in both high-resource and low-resource settings when compared with baselines. 
 Its performance in low-resource settings is particularly noteworthy, as it not only sustains high BLEU scores but also exhibits a Self-BLEU score of 5.09, the highest among all models evaluated. This high Self-BLEU score indicates less diversity in the generated text, suggesting that while EMMA-500 maintains consistency, it may produce less varied outputs.
 When compared to other high-performing models like Qwen 2 7B and Llama 3.1 8B, the EMMA-500 model exhibits a superior balance between accuracy and linguistic creativity. 
 Unlike Qwen 2 7B, which shows a spike in performance primarily in medium-low resource settings, EMMA-500 maintains a consistently high performance across varying levels of resource availability.

 \paragraph{PolyWrite} 
 This is a novel multilingual benchmark composed in this work for evaluating open-ended generation in 240 languages. 
 We use ChatGPT to generate different prompts in English and use Google Translate to translate them into different languages for models to generate creative content.
 This benchmark consists of 31 writing tasks, such as storytelling and email writing, and 155 prompts in total.
 We back-translate the multilingual prompts to English, calculate the BLEU scores between original English prompts and back-translation, and filter out translated prompts with BLEU scores below 20, and the entire dataset contains a total of 35,751 prompts.
 We release the PolyWrite dataset on Huggingface\footnote{\url{https://huggingface.co/datasets/MaLA-LM/PolyWrite}}. 
 The details of PolyWrite are described in \Cref{sec:ploywrite}.

 We use Self-BLEU~\citep{zhu2018texygen} to evaluate the diversity of generated texts in the PolyWrite benchmark, as presented in \Cref{tab:aya_polywrite}. 
 A lower Self-BLEU score indicates more diverse generation, but does not means a better generation quality.
 Our EMMA-500 model demonstrates comparable performance across various languages. 
 Compared to other models like Llama 3/3.1 and Qwen 1.5/2, particularly in medium-low and low-resource languages, EMMA-500 has higher Self-BLEU scores, indicating lower diversity in its generated content. 

 Evaluating open-ended generation poses significant challenges, as it goes beyond simply measuring accuracy or correctness. Metrics like BLEU or Self-BLEU, while useful for assessing similarity to reference texts or the diversity of given texts, often fail to capture more nuanced aspects.
 Subjective factors like cultural relevance and the appropriateness of responses in low-resource languages are difficult to quantify. This makes it challenging to create evaluation benchmarks and metrics that fully capture the strengths and weaknesses of models like EMMA-500 in diverse, real-world scenarios.

 \begin{table*}[ht!]
 \caption{Results on Aya (BLEU/Self-BLEU) and PolyWrite (Self-BLEU). EMMA-500 Llama 2 7B has higher average BLEU scores than all baselines on Aya.}
 \label{tab:aya_polywrite}
 \setlength{\tabcolsep}{3pt}
 \scriptsize
 \centering
 \resizebox{\linewidth}{!}{
 \begin{tabular}{lrrrrrrrrrrrr}
 \toprule
 \multicolumn{1}{c}{\multirow{2}{*}{\textbf{Model}}} & \multicolumn{6}{c}{\textbf{Aya}}                                                                                                                                      & \multicolumn{6}{c}{\textbf{PolyWrite}}                                                                                                                                    \\ 
 \cmidrule(lr){2-7}\cmidrule(lr){8-13}
 \multicolumn{1}{c}{}                                         & \ch{Avg} & \ch{High} & \ch{Med-High} & \ch{Medium} & \ch{Med-Low} & \ch{Low} & \ch{Avg} & \ch{High} & \ch{Med-High} & \ch{Medium} & \ch{Med-Low} & \ch{Low} \\
 \midrule
 Llama 2 7B                        & 1.24/0.74             & 1.27/0.57              & 1.47/0.60                     & 0.86/0.37                & 1.17/0.45                    & 0.77/1.87             & 0.5358                & 0.4282                 & 0.6545                        & 0.3769                   & 0.4766                       & 0.6587                \\
 Llama 2 7B Chat                   & 1.17/1.29             & 1.46/1.15              & 1.36/1.15                     & 0.69/1.03                & 1.14/1.14                    & 0.54/2.23             & 1.1550                & 0.8640                 & 0.8902                        & 1.1877                   & 1.4435                       & 1.4167                \\
 CodeLlama 2 7B                    & 1.22/1.21             & 1.31/1.19              & 1.40/1.00                     & 0.85/0.84                & 1.23/0.56                    & 0.78/2.57             & 1.0313                & 1.2052                 & 1.2883                        & 0.9191                   & 0.9092                       & 0.6798                \\
 LLaMAX Llama 2 7B                 & 1.72/1.70             & 1.80/1.37              & 2.03/1.48                     & 1.30/1.24                & 1.65/0.89                    & 0.98/3.74             & 0.9564                & 1.0066                 & 1.1655                        & 0.8786                   & 0.8363                       & 0.7709                \\
 LLaMAX Llama 2 7B Alpaca          & 1.68/1.67             & 1.82/1.22              & 1.96/1.42                     & 1.28/1.14                & 1.55/1.04                    & 1.00/3.94             & 0.8086                & 0.7321                 & 0.9517                        & 0.8322                   & 0.8351                       & 0.4981                \\
 MaLA-500 Llama 2 10B v1           & 0.40/2.29             & 0.42/2.53              & 0.49/2.79                     & 0.32/1.22                & 0.31/2.42                    & 0.18/1.16             & 3.7079                & 3.7902                 & 3.5066                        & 2.4541                   & 4.8052                       & 3.5163                \\
 MaLA-500 Llama 2 10B v2           & 0.41/2.31             & 0.42/2.01              & 0.50/2.65                     & 0.32/1.42                & 0.33/1.02                    & 0.18/3.09             & 4.0059                & 3.1737                 & 4.0148                        & 3.0476                   & 5.0652                       & 3.8916                \\
 Yayi Llama 2 7B                   & 1.65/0.61             & 1.84/0.82              & 1.88/0.62                     & 1.26/0.62                & 1.53/0.45                    & 1.02/0.41             & 0.6274                & 0.6207                 & 0.6921                        & 0.4169                   & 0.6813                       & 0.6352                \\
 TowerBase Llama 2 7B              & 1.44/0.83             & 1.45/0.64              & 1.67/0.56                     & 1.19/0.49                & 1.46/0.38                    & 0.88/2.54             & 0.4938                & 0.7268                 & 0.4736                        & 0.3945                   & 0.4417                       & 0.5396                \\
 TowerInstruct Llama 2 7B          & 1.55/0.93             & 1.80/0.85              & 1.82/0.78                     & 1.15/0.65                & 1.21/0.54                    & 0.85/1.97             & 0.7124                & 0.9565                 & 0.7651                        & 0.6492                   & 0.5998                       & 0.6615                \\ 
 \midrule
 Occiglot Mistral 7B v0.1          & 1.53/2.43             & 1.57/0.91              & 1.78/2.89                     & 1.17/1.73                & 1.61/1.31                    & 0.95/4.27             & 0.9647                & 0.8818                 & 0.7975                        & 0.9543                   & 0.9563                       & 1.4157                \\
 Occiglot Mistral 7B v0.1 Instruct & 0.75/2.81             & 0.82/2.38              & 0.89/3.10                     & 0.43/2.72                & 0.79/1.17                    & 0.45/3.50             & 3.9033                & 5.3884                 & 4.7140                        & 3.7555                   & 2.8444                       & 2.9568                \\
 BLOOM 7B                          & 0.85/1.17             & 0.92/1.20              & 0.96/1.32                     & 0.73/1.05                & 0.78/1.27                    & 0.55/0.72             & 1.3892                & 1.2845                 & 1.6705                        & 1.9685                   & 1.1513                       & 0.6600                \\
 BLOOMZ 7B                         & 0.12/0.61             & 0.07/0.33              & 0.17/0.62                     & 0.08/1.00                & 0.06/0.89                    & 0.08/0.51             & 0.0024                & 0.0000                 & 0.0005                        & 0.0093                   & 0.0000                       & 0.0049                \\
 mGPT                              & 1.24/0.55             & 1.22/0.64              & 1.47/0.59                     & 0.91/0.48                & 1.21/0.60                    & 0.84/0.29             & 0.7560                & 0.9222                 & 0.6291                        & 0.4534                   & 0.8156                       & 1.1134                \\
 mGPT-13B                          & 1.42/0.57             & 1.42/0.80              & 1.63/0.58                     & 1.00/0.48                & 1.53/0.44                    & 1.03/0.36             & 0.7479                & 0.8483                 & 0.7091                        & 0.4230                   & 0.7962                       & 1.0201                \\
 Yayi 7B                           & 1.05/0.39             & 1.22/0.38              & 1.18/0.41                     & 0.76/0.42                & 1.01/0.54                    & 0.67/0.22             & 0.5151                & 0.4266                 & 0.3574                        & 0.3283                   & 0.6706                       & 0.8574                \\ 
 \midrule
 Llama 3 8B                        & 1.59/0.94             & 1.03/0.60              & 1.31/0.53                     & 2.96/0.54                & 1.11/0.28                    & 2.43/3.47             & 0.5796                & 0.5753                 & 0.5921                        & 0.7141                   & 0.5586                       & 0.4440                \\
 Llama 3.1 8B                      & 1.85/1.08             & 1.41/0.95              & 1.60/0.64                     & 3.11/0.52                & 1.33/0.39                    & 2.52/3.52             & 0.7995                & 0.6805                 & 0.8253                        & 0.5044                   & 0.6965                       & 1.3585                \\
 Gemma 2 9B                        & 1.55/0.82             & 1.59/0.94              & 1.73/0.93                     & 1.38/0.70                & 1.33/0.47                    & 1.21/0.65             & 1.1736                & 1.2347                 & 1.2913                        & 1.1616                   & 0.9261                       & 1.3240                \\
 Gemma 7B                          & 1.29/0.57             & 1.41/0.62              & 1.39/0.66                     & 1.16/0.40                & 1.26/0.50                    & 0.93/0.40             & 1.0541                & 0.9629                 & 1.1222                        & 0.9275                   & 1.1112                       & 1.0284                \\
 Qwen 1.5 7B                       & 1.93/1.13             & 1.66/0.85              & 1.64/0.65                     & 3.14/0.79                & 1.51/0.58                    & 2.45/3.69             & 0.6441                & 0.9398                 & 0.7457                        & 0.3797                   & 0.4164                       & 0.8738                \\
 Qwen 2 7B                         & 1.99/1.13             & 1.84/0.94              & 1.69/0.70                     & 3.29/0.72                & 1.36/0.44                    & 2.47/3.53             & 0.5709                & 0.7763                 & 0.6135                        & 0.3695                   & 0.4186                       & 0.7989                \\ 
 \midrule
 EMMA-500 Llama 2 7B               & 2.93/1.54             & 2.90/1.29              & 2.75/0.83                     & 3.79/0.82                & 2.86/0.95                    & 2.87/5.09             & 0.9879                & 0.9833                 & 0.7058                        & 0.8465                   & 1.3130                       & 1.1702                \\ 
 \bottomrule
 \end{tabular}
 }
 \end{table*}

\end{appendices}

\end{document}